\documentclass[Afour,sageh,times]{sagej}
\usepackage{moreverb,url}
\usepackage[colorlinks,bookmarksopen,bookmarksnumbered,citecolor=blue,urlcolor=red]{hyperref}
\usepackage{dsfont}

\usepackage{longtable}
\usepackage{supertabular,booktabs}
\usepackage{makecell}
\usepackage{comment}
\usepackage{amsmath}
\usepackage{multirow}
\usepackage{tabularx}
\newcolumntype{Y}{>{\centering\arraybackslash}X} 
\newcommand{\specialcell}[2][c]{%
  \begin{tabular}[#1]{@{}c@{}}#2\end{tabular}}
\usepackage{diagbox}

\usepackage{comment}
\usepackage{siunitx}
\usepackage{relsize}
\usepackage{ifthen}
\usepackage[colorinlistoftodos]{todonotes}

\usepackage[caption=false]{subfig}

\usepackage[vlined,ruled,linesnumbered]{algorithm2e}
\usepackage{graphics} %
\usepackage{rotating}
\usepackage{color}
\usepackage{enumerate}
\usepackage[T1]{fontenc}
\usepackage{psfrag}
\usepackage{epsfig} %
\usepackage{booktabs}
\usepackage{graphicx,url}
\usepackage{multirow}
\usepackage{array}
\usepackage{latexsym}
\usepackage{amsfonts}
\usepackage{amsmath}
\usepackage{amssymb}
\usepackage{xstring}
\usepackage[noend]{algorithmic}
\usepackage{multirow}
\usepackage{xcolor}
\usepackage{prettyref}
\usepackage{flexisym}
\usepackage{bigdelim}
\usepackage{breqn} %
\usepackage{listings}

\usepackage{enumitem}
\usepackage{xspace}
\usepackage{bm}
\graphicspath{{./figures/}}
\usepackage{tikz}
\usetikzlibrary{matrix,calc}

\usepackage{mdwlist}

\makecompactlist{itemize}{stditemize}

\newrefformat{prob}{Problem\,\ref{#1}}
\newrefformat{def}{Definition\,\ref{#1}}
\newrefformat{sec}{Section\,\ref{#1}}
\newrefformat{sub}{Section\,\ref{#1}}
\newrefformat{prop}{Proposition\,\ref{#1}}
\newrefformat{app}{Appendix\,\ref{#1}}
\newrefformat{alg}{Algorithm\,\ref{#1}}
\newrefformat{cor}{Corollary\,\ref{#1}}
\newrefformat{thm}{Theorem\,\ref{#1}}
\newrefformat{lem}{Lemma\,\ref{#1}}
\newrefformat{fig}{Fig.\,\ref{#1}}
\newrefformat{tab}{Table\,\ref{#1}}

\newtheorem{theorem}{Theorem}

\newtheorem{remark}[theorem]{Remark}

\newcommand{\bdmath}{\begin{dmath}}
\newcommand{\edmath}{\end{dmath}}
\newcommand{\beq}{\begin{equation}}
\newcommand{\eeq}{\end{equation}}
\newcommand{\bdm}{\begin{displaymath}}
\newcommand{\edm}{\end{displaymath}}
\newcommand{\bea}{\begin{eqnarray}}
\newcommand{\eea}{\end{eqnarray}}
\newcommand{\beal}{\beq \begin{array}{ll}}
\newcommand{\eeal}{\end{array} \eeq}
\newcommand{\beas}{\begin{eqnarray*}}
\newcommand{\eeas}{\end{eqnarray*}}
\newcommand{\ba}{\begin{array}}
\newcommand{\ea}{\end{array}}
\newcommand{\bit}{\begin{itemize}}
\newcommand{\eit}{\end{itemize}}
\newcommand{\ben}{\begin{enumerate}}
\newcommand{\een}{\end{enumerate}}

\newcommand{\calG}{{\cal G}}

\newcommand{\calZ}{{\cal Z}}

\newcommand{\eg}{\emph{e.g.,}\xspace}
\newcommand{\ie}{\emph{i.e.,}\xspace}
\newcommand{\myParagraph}[1]{{\bf #1.}\xspace}

\newcommand{\M}[1]{{\bm #1}} %
\renewcommand{\boldsymbol}[1]{{\bm #1}}

\newcommand{\LC}[1]{{\color{red} \textbf{LC}: #1}}

\newcommand{\hide}[1]{}

\newcommand{\hiddenText}{{\color{gray} hidden text.}}
\newcommand{\hideWithText}[1]{\hiddenText}

\DeclareMathOperator*{\argmin}{arg\,min}

\newcommand{\trace}[1]{\mathrm{tr}\left(#1\right)}

\newcommand{\eye}{{\mathbf I}}

\newcommand{\Real}[1]{ { {\mathbb R}^{#1} } }

\newcommand{\SEthree}{\ensuremath{\mathrm{SE}(3)}\xspace}

\newcommand{\MA}{\M{A}}

\newcommand{\ME}{\M{E}}

\newcommand{\MG}{\M{G}}
\newcommand{\MM}{\M{M}}

\newcommand{\MR}{\M{R}}

\newcommand{\MI}{\M{I}}

\newcommand{\MT}{\M{T}}
\newcommand{\MX}{\M{X}}

\newcommand{\MZ}{\M{Z}}
 
\newcommand{\MOmega}{\M{\Omega}}

\newcommand{\vg}{\boldsymbol{g}}

\newcommand{\vv}{\boldsymbol{v}}
\newcommand{\vt}{\boldsymbol{t}}

\newcommand{\scenario}[1]{{\smaller \sf#1}\xspace}

\newcommand{\blue}[1]{{\color{blue}#1}}

\newcommand{\linkToPdf}[1]{\href{#1}{\blue{(pdf)}}}
\newcommand{\linkToPpt}[1]{\href{#1}{\blue{(ppt)}}}
\newcommand{\linkToCode}[1]{\href{#1}{\blue{(code)}}}
\newcommand{\linkToWeb}[1]{\href{#1}{\blue{(web)}}}
\newcommand{\linkToVideo}[1]{\href{#1}{\blue{(video)}}}
\newcommand{\linkToMedia}[1]{\href{#1}{\blue{(media)}}}
\newcommand{\award}[1]{\xspace} %

\lstdefinelanguage{mine} 
{morekeywords={while,True,if,break,=,return,function,for,until,in,input,output,assumptions,assumption,invariant,loop,variant,end,invariant,precondition,variables}, 
sensitive=false, 
morecomment=[l]{\#}, 
morecomment=[il]\%{.}, 
morecomment=[s]{/*}{*/}, 
morestring=[b]", 
} 

\usepackage{calc}
\newlength\listingnumberwidth
\usepackage[font=small,labelfont=bf]{caption}
\usepackage{pifont}%

\newcommand{\Kimera}{{Kimera}\xspace}
\newcommand{\KimeraCore}{{Kimera-Core}\xspace}
\newcommand{\KimeraDSG}{{Kimera-DSG}\xspace}
\newcommand{\KimeraVIO}{{Kimera-VIO}\xspace}
\newcommand{\KimeraRPGO}{{Kimera-RPGO}\xspace}
\newcommand{\KimeraPGMO}{{Kimera-PGMO}\xspace}
\newcommand{\KimeraMesher}{{Kimera-Mesher}\xspace}
\newcommand{\KimeraSemantics}{{Kimera-Semantics}\xspace}
\newcommand{\KimeraObjects}{{Kimera-Objects}\xspace}
\newcommand{\KimeraHumans}{{Kimera-Humans}\xspace}
\newcommand{\KimeraBuildingParser}{{Kimera-BuildingParser}\xspace}

\newcommand{\KimeraDVIOsuff}{DVIO}
\newcommand{\KimeraDVIO}{{\KimeraDVIOsuff}\xspace}

\newcommand{\perFrameMeshLatency}{20\text{ms}}
\newcommand{\globalMeshLatency}{0.1\text{s}}

\newcommand{\KimeraUrl}{{\small\href{https://github.com/MIT-SPARK/Kimera}{https://github.com/MIT-SPARK/Kimera}}}
\newcommand{\VideoUrl}{{\small\href{https://youtu.be/-5XxXRABXJs}{https://youtu.be/-5XxXRABXJs}}}
\newcommand{\DSGVideoUrl}{{\small\href{https://youtu.be/SWbofjhyPzI}{https://youtu.be/SWbofjhyPzI}}}

\newcommand{\toAdd}[1]{}

\newcommand{\vinVersion}[2]{#2} %

\newcommand{\Euroc}{EuRoC\xspace}

\newcommand{\DSG}{{DSG}\xspace}
\newcommand{\DSGs}{{DSGs}\xspace}
\newcommand{\MeshName}{{\tt GraphCMR}\xspace}

\newcommand{\TEASERpp}{\scenario{TEASER++}}

\newcommand{\SPES}{{Spatial Perception Engine}\xspace}

\newcommand{\SPESlong}{Spatial Perception Engine\xspace}

\newcommand{\FigFrontCover}{Figure~\ref{fig:DSG}}%

\newcommand{\layerOneName}{Metric-Semantic Mesh\xspace}
\newcommand{\layerTwoName}{Objects and Agents\xspace}
\newcommand{\layerThreeName}{Places and Structures\xspace}
\newcommand{\layerFourName}{Rooms\xspace}
\newcommand{\layerSixName}{Building\xspace}

\newcommand{\maxJointDist}{3\rm{m}}

\newcommand{\UnityHumans}{\scenario{uHumans}}
\newcommand{\UnityHumansTwo}{\scenario{uHumans2}}

\newcommand{\ESDF}{ESDF\xspace}

\newcommand{\optional}[1]{}%
\newcommand{\veryOptional}[1]{}%
\newcommand{\hideout}[1]{}

\newcommand{\RealSense}{RealSense\xspace}
\newcommand{\RealSenseLong}{Intel RealSense D435i\xspace}
\newcommand{\AzureKinect}{Azure Kinect\xspace}
\newcommand{\WhiteOwl}{`White Owl'\xspace}
\newcommand{\BldgThirtyOne}{`AeroAstro'\xspace}
\newcommand{\School}{`School'\xspace}

\def\journalname{The International Journal of Robotics Research}
\usepackage{soul}
\usepackage{xr}
\usepackage{cleveref}
\newcommand{\omitted}[1]{}

\newcommand\rebuttal[1]{{#1}}

\graphicspath{{./figures/}}

\usepackage{enumitem}
\setlist{leftmargin=5mm}

\usepackage{tikz}

\PassOptionsToPackage{end}{algorithmic} 

\begin{document}

\runninghead{Rosinol et al.}

\title{\LARGE{\Kimera: from SLAM to Spatial Perception \\ with  3D Dynamic Scene Graphs}}

\author{Antoni Rosinol, Andrew Violette, {Marcus Abate}, {Nathan Hughes},\\ 
{Yun Chang}, Jingnan Shi, {Arjun Gupta,} Luca Carlone}

\affiliation{Laboratory for Information \& Decision Systems, Massachusetts Institute of Technology, USA. \\ \\
This work was partially funded by ARL DCIST CRA W911NF-17-2-0181, 
ONR RAIDER N00014-18-1-2828, MIT Lincoln Laboratory, an Amazon Research Award,
and ``la Caixa'' Foundation (ID 100010434), LCF/BQ/AA18/11680088 (A. Rosinol).
}

\corrauth{Antoni Rosinol,
Laboratory for Information \& Decision Systems,
Massachusetts Institute of Technology,
Cambridge, MA,
USA
\email{arosinol@mit.edu}
}

\begin{abstract}
Humans are able to form a complex mental model of the environment they move in.
This mental model captures geometric and semantic aspects of the scene, 
describes the environment at multiple levels of abstractions (\eg objects, rooms, buildings), 
includes static and dynamic entities and their relations 
(\eg a person is in a room at a given time).
 In contrast, current robots' internal representations still provide a 
 partial and fragmented understanding of the environment, either in the form of a 
 sparse or dense set of geometric primitives (\eg points, lines, planes, voxels), 
 or as a collection of objects.
 This paper attempts to %
 reduce the gap between robot and human perception 
 by introducing a 
 novel representation, a \emph{3D Dynamic Scene Graph} (\DSG), 
 that seamlessly captures metric and semantic aspects of a dynamic environment. %
 A \DSG is a layered graph where nodes represent 
 spatial concepts at different levels of abstraction, and edges represent spatio-temporal relations among nodes.
Our second contribution is \emph{\Kimera}, the first fully automatic method to build 
a \DSG from visual-inertial data. %
\Kimera includes \rebuttal{accurate algorithms} for visual-inertial SLAM, 
metric-semantic 3D reconstruction, object localization, human pose and shape estimation, 
and scene parsing.  %
Our third contribution is a comprehensive evaluation of \Kimera in real-life datasets and photo-realistic simulations, 
including a newly released dataset, \UnityHumansTwo, which simulates a collection of  crowded indoor and outdoor scenes.
Our evaluation shows that \Kimera achieves \rebuttal{competitive} performance in visual-inertial SLAM, 
estimates an accurate 3D metric-semantic mesh model in real-time, and builds a \DSG of a complex indoor environment 
 with tens of objects and humans in minutes.
Our final contribution is to showcase how to use a \DSG for real-time hierarchical semantic path-planning. 
The core modules in \Kimera have been released open source.
\end{abstract}

\maketitle

\begin{tikzpicture}[overlay, remember picture]
\path (current page.north east) ++(-6.2,-0.0) node[below left] {
Accepted for publication at IJRR 2021, please cite as follows:
};
\end{tikzpicture}
\begin{tikzpicture}[overlay, remember picture]
\path (current page.north east) ++(-5.8,-0.4) node[below left] {
Antoni Rosinol, Andrew Violette, {Marcus Abate}, {Nathan Hughes},
};
\end{tikzpicture}
\begin{tikzpicture}[overlay, remember picture]
\path (current page.north east) ++(-6.8,-0.8) node[below left] {
{Yun Chang}, Jingnan Shi, {Arjun Gupta,} Luca Carlone
};
\end{tikzpicture}
\begin{tikzpicture}[overlay, remember picture]
\path (current page.north east) ++(-4.8,-1.2) node[below left] {
``Kimera: from SLAM to Spatial Perception with  3D Dynamic Scene Graphs''
};
\end{tikzpicture}
\begin{tikzpicture}[overlay, remember picture]
\path (current page.north east) ++(-6.6,-1.6) node[below left] {
The International Journal of Robotics Research, 2021.
};
\end{tikzpicture}

\thispagestyle{empty} 
\pagestyle{empty}

\vspace{-1mm}
\section*{Supplementary Material} 
\vspace{-1mm}
{Code: \KimeraUrl}\\
{Video 1: \VideoUrl}\\
{Video 2: \DSGVideoUrl}


\section{Introduction}
\label{sec:intro}

High-level scene understanding is a prerequisite for safe and long-term
autonomous operation of robots and autonomous vehicles, and for effective human-robot interaction. %
 The next generations of robots must be able to %
 understand and execute high-level instructions, such as 
``search for survivors on the second floor'' or ``go and pick up the grocery bag in the kitchen''.
 They must be able to plan and act over long distances and extended time horizons to support lifelong operation.
Moreover, they need a holistic understanding of the scene that allows reasoning about inconsistencies, 
causal relations, and occluded objects. %

As humans, we perform all these operations effortlessly: 
we understand high-level instructions, plan over long distances 
(\eg plan a trip from Boston to Rome), and perform advanced reasoning about the environment.  
For instance, as humans, we can easily infer that if the car preceding us is suddenly stopping 
in the middle of the road and in proximity 
to a pedestrian crossing, a pedestrian is likely to be crossing the street even if occluded by the car
in front of us. %
 This is in stark contrast with today's robot capabilities: 
 robots are often issued geometric commands (\eg ``reach coordinates XYZ''), 
 do not have %
 suitable representations (nor inference algorithms) to support decision making at multiple levels of abstraction, 
 and have no notion of causality or high-level reasoning.

  High-level understanding of 3D dynamic scenes involves three key ingredients: 
  (i) understanding the geometry, semantics, and physics of the scene, 
  (ii) representing the scene at multiple levels of abstraction, and 
  (iii) capturing spatio-temporal relations among entities (objects, structure, humans). 
  We discuss the importance of each aspect and highlight shortcomings of current methods below.

The first ingredient, \emph{metric-semantic understanding}, is the capability of grounding semantic concepts 
 (\eg survivor, grocery bag, kitchen) into a spatial representation (\ie a metric map). 
  Geometric information is critical for robots to navigate safely and to manipulate objects, while semantic information provides 
  an ideal level of abstraction for the robot to understand and execute human instructions
  (\eg``bring me a cup of coffee'') and to provide humans with models of the environment that are easy to understand.
  Despite the unprecedented progress in \textit{geometric reconstruction} (\eg SLAM~\citep{Cadena16tro-SLAMsurvey}, Structure from Motion~\citep{Enqvist11iccv}, and Multi-View Stereo~\citep{Schoeps17cvpr}) and deep-learning-based \textit{semantic segmentation} (\eg~\citep{GarciaGarcia17arxiv,Krizhevsky12cvpr-deepNets,Redmon17cvpr-yolo9000,Ren15nips-RCNN,He17iccv-maskRCNN,Hu18cvpr-maskXRCNN,Badrinarayanan15pami-segnet}),
  research in these two fields has traditionally proceeded in isolation, and 
  there has been a recent and growing  
research at the intersection of these areas~\citep{Bao11cvpr,Cadena16tro-SLAMsurvey,Bowman17icra,Hackel17arxiv-semantic3d,Grinvald19ral-voxbloxpp,Zheng19arxiv-metricSemantic,Davison18-futuremapping}.

The second ingredient is the capability of providing an \emph{actionable understanding
of the scene at multiple levels of abstraction}.
The need for abstractions is mostly dictated by computation and communication 
constraints. As humans, when planning a long trip, we reason in terms of cities or airports, since that is more (computationally) 
convenient than reasoning over Cartesian coordinates.  
 Similarly, when asked for directions in a building, we find more convenient to list corridors, rooms, and floors, rather than 
 drawing a metrically accurate path to follow. 
 Similarly, 
 robots break down the complexity of decision making, by planning at multiple levels of abstractions, 
 from high-level task planning, to motion planning and trajectory optimization, to low-level control and obstacle avoidance, 
 where each abstraction trades-off model fidelity for computational efficiency. 
 Supporting hierarchical decision making and planning demands robot perception to be capable of building a 
\emph{hierarchy of consistent abstractions} to feed task planning, motion planning, and reactive control. %
Early work on map representation in robotics, \eg~\citep{Kuipers00ai,Kuipers78cs,Chatila85,Vasudevan06iros,Galindo05iros-multiHierarchicalMaps,Zender08ras-spatialRepresentations}, 
 investigated hierarchical representations but mostly in 2D and assuming static environments; moreover, 
 these works were proposed before the ``deep learning revolution'', hence they could not afford advanced semantic 
 understanding. 
 On the other hand,
 the growing literature on metric-semantic mapping~\citep{Salas-Moreno13cvpr,Bowman17icra,Behley19iccv-semanticKitti,Tateno15iros-metricSemantic,Rosinol20icra-Kimera,Grinvald19ral-voxbloxpp,McCormac17icra-semanticFusion},
  focuses on ``flat'' representations (object constellations, metric-semantic meshes or volumetric models) that are not
  hierarchical in nature. 

The third ingredient of high-level understanding is the capability of describing
 both \emph{static and dynamic entities in the scene and reason on their relations}.
  Reasoning at the level of objects and their (geometric and physical) relations %
  is again instrumental to parse high-level instructions (\eg ``pick up the glass on the table'').
  It is also crucial to guarantee safe operation: in many application from 
 self-driving cars to collaborative robots on factory floors,  
identifying obstacles is not sufficient for safe and effective navigation/action, 
and  it becomes 
 crucial to capture the \emph{dynamic} entities in the scene (in particular, \emph{humans}), 
 and predict their behavior or intentions~\citep{Everett18iros-motionAmongDynamicAgents}.
  Very recent work~\citep{Armeni19iccv-3DsceneGraphs,Kim19tc-3DsceneGraphs} attempts to capture object relations thought 
  a rich representation, namely
  \emph{3D Scene Graphs}.
 A scene graph is a data structure commonly used in computer graphics and gaming applications that consists of a 
 graph where nodes represent entities in the scene and edges represent spatial or logical relationships among nodes.
 While the works~\citep{Armeni19iccv-3DsceneGraphs,Kim19tc-3DsceneGraphs} pioneered 
 the use of 3D scene graphs in robotics and vision (prior work in vision focused on 2D scene graphs defined in the image space~\citep{Choi13cvpr-sceneParsing,Zhao13cvpr-sceneParsing,Huang18eccv-sceneParsing, Jiang18ijcv-sceneParsing}), 
 they have important drawbacks. \cite{Kim19tc-3DsceneGraphs} only capture objects and miss multiple levels of abstraction. \cite{Armeni19iccv-3DsceneGraphs} provide a hierarchical model that is useful for visualization and knowledge organization, but does not capture \emph{actionable} information, such as traversability, which is key to robot navigation. 
 Finally, %
 neither~\cite{Kim19tc-3DsceneGraphs} nor~\cite{Armeni19iccv-3DsceneGraphs}  account for or model dynamic entities in the environment, which is crucial for robots moving in human-populated environments.

\myParagraph{Contributions} 
While the design and implementation of a robot perception system that 
effectively includes all these ingredients %
 can only be the goal of a long-term research agenda,
  this paper provides the first step towards this goal, by 
  proposing a novel and general representation of the environment, and practical algorithms to infer it from data.
In particular, this paper provides four contributions.

The first contribution (Section~\ref{sec:DSG}) is a 
unified representation for actionable spatial perception:
a {\bf 3D Dynamic Scene Graph} (\DSG).
  A \DSG is a \emph{layered} directed graph where nodes represent
  \emph{spatial concepts} (\eg objects, rooms, agents) and edges represent pairwise spatio-temporal relations. 
   The graph is \emph{layered}, in that nodes are grouped into layers that correspond to different levels of abstraction of the scene 
   (\ie~a \DSG is a hierarchical representation).
   Our choice of nodes and edges in the \DSG also captures \emph{places} and their connectivity, hence providing a 
   strict generalization of the notion of topological maps~\citep{Ranganathan04iros,Remolina04} and making \DSGs an \emph{actionable} representation for navigation and planning. 
   Finally, edges in the \DSG capture spatio-temporal relations and explicitly model dynamic entities in the scene, and in particular humans, for which we estimate both 3D poses over time (using a \emph{pose graph} model) and a dense mesh model. 

Our second contribution (\Cref{sec:Kimera}) is {\bf \Kimera, the first \SPESlong}
that builds a \DSG from  visual-inertial data collected by a robot. 
\Kimera has two sets of modules: \KimeraCore and \KimeraDSG.
\KimeraCore~\citep{Rosinol20icra-Kimera} is in charge of the real-time metric-semantic reconstruction of the scene, and comprises the following modules:
\begin{itemize}%
  \item \emph{\KimeraVIO} (\Cref{sec:vio}) is a visual-inertial odometry (VIO) module for fast and locally accurate 3D pose estimation (localization).
  \item \emph{\KimeraMesher} (\Cref{sec:mesher}) reconstructs a fast local 3D mesh for collision avoidance.
  \item \emph{\KimeraSemantics} (\Cref{sec:semantic})  builds a global 
     3D mesh using a volumetric approach~\citep{Oleynikova2017iros-voxblox}, and
     semantically annotates the 3D mesh using 2D pixel-wise segmentation and 3D Bayesian updates. 
     \KimeraSemantics uses the pose estimates  from \KimeraVIO.
  \item \emph{\KimeraPGMO} (Pose Graph and Mesh Optimization, Section~\ref{sec:meshLoopClosure}) enforces visual loop closures by simultaneously optimizing the 
    pose graph describing the robot trajectory and \KimeraSemantics's global metric-semantic mesh.
    This new module generalizes \emph{\KimeraRPGO} (Robust Pose Graph Optimization,~\cite{Rosinol20icra-Kimera}), which only optimizes the pose graph describing the robot trajectory. 
    Like \KimeraRPGO, \KimeraPGMO includes a mechanism to reject outlying loop closures. 
\end{itemize}

\KimeraDSG is in charge of building the \DSG of the scene and works on top of \KimeraCore.
\KimeraDSG comprises the following modules:
\begin{itemize}%
  \item \emph{\KimeraHumans} (\Cref{ssec:humans}) reconstructs dense
    meshes of humans in the scene, and estimates their trajectories using a pose graph model.
    The dense meshes are parametrized using the \emph{Skinned Multi-Person Linear Model} (SMPL) by~\cite{Loper15tg-smpl}. 
  \item \emph{\KimeraObjects} (Section~\ref{sec:objects}) estimates a bounding box for objects of unknown shape and 
    fits a CAD model to objects of known shape in the metric-semantic mesh using \TEASERpp by~\cite{Yang20tro-teaser}.
   \item \emph{\KimeraBuildingParser} (Section~\ref{sec:buildingParser}) parses the metric-semantic mesh into a 
    topological graph of places (\ie obstacle-free locations), segments rooms, and identifies structures (\ie walls, ceiling) enclosing the rooms.  
 \end{itemize}

The notion of a \textbf{\SPES generalizes SLAM}, which becomes a module in \Kimera, and augments it to capture  
a hierarchy of spatial concepts and their relations.
Besides the novelty of many modules in \Kimera (\eg~\KimeraPGMO, \KimeraHumans, \KimeraBuildingParser), 
our \SPES (i) is the first to construct a scene graph from sensor data 
(in contrast to \cite{Armeni19iccv-3DsceneGraphs} that assume an annotated mesh model to be given),
 (ii) provides a \emph{lightweight and scalable} CPU-based
  solution, %
  and (iii) is robust to dynamic environments and incorrect place recognition.
The core modules of \Kimera have been released at \KimeraUrl.

Our third contribution (\Cref{sec:experiments}) is an extensive {\bf experimental evaluation and the release of a new 
photo-realistic dataset}.
We test \Kimera in both real and simulated datasets, including 
the \Euroc dataset~\citep{Burri16ijrr-eurocDataset}, and the \UnityHumans dataset we released with~\citep{Rosinol20rss-dynamicSceneGraphs}.
In  addition to these datasets, we release the \UnityHumansTwo dataset, which encompasses crowded indoor and outdoor scenes, including 
an apartment, an office building, a subway, and a residential neighborhood.
{Finally, we qualitatively evaluate \Kimera on real datasets collected in an apartment and an office space.}
The evaluations shows that \Kimera's modules (i) achieve \rebuttal{competitive} performance in visual-inertial SLAM,
 (ii) can reconstruct a metric-semantic mesh in real-time on an embedded CPU,
  (iii) can correctly deform a dense mesh to enforce loop closures,
   (iv) can accurately localize and track objects and humans,
    and (v) can correctly partition an indoor building into rooms, places, and structures.

Our final contribution (\Cref{sec:examples}) is to {\bf demonstrate potential queries that can 
be implemented on a \DSG, %
including an example of {hierarchical semantic path planning}}. %
In particular, we show how a robot can use a \DSG to understand and execute high-level instructions, such as ``reach the person near the sofa'' (\ie semantic path planning). 
We also demonstrate that a \DSG allows to compute path planning queries in a fraction of the time taken 
by a planner using volumetric approaches, by taking advantage of the hierarchical nature of the \DSG.

We conclude the paper with an extensive literature review (Section~\ref{sec:relatedWork}) 
and a discussion of future work (Section~\ref{sec:conclusion}).

\myParagraph{Novelty with respect to previous work~\citep{Rosinol20icra-Kimera,Rosinol20rss-dynamicSceneGraphs}}
This paper brings to maturity our previous work on \Kimera~\citep{Rosinol20icra-Kimera} 
 (whose modules are now extended and included in \KimeraCore),
and 3D Dynamic Scene Graphs~\citep{Rosinol20rss-dynamicSceneGraphs} and provides several novel contributions.
First, we introduce a new loop closure mechanism that deforms the metric-semantic mesh (\KimeraPGMO), while 
the mesh in~\citep{Rosinol20icra-Kimera} did not incorporate corrections resulting from loop closures.
Second, we implement and test a semantic hierarchical path-planning algorithm on \DSGs, which 
was only discussed in~\citep{Rosinol20rss-dynamicSceneGraphs}.
Third, we provide a more comprehensive evaluation, including our own real datasets and new simulated datasets (\UnityHumansTwo).
Moreover, we release this new simulated dataset (with 12 new scenes, including outdoor environments, going beyond the 
indoor evaluation of \citep{Rosinol20rss-dynamicSceneGraphs}).
Finally, we test \Kimera on an NVIDIA TX2 computer and 
show it executes in real-time on embedded hardware.


\begin{figure*}[h!]
  \centering
  
  \begin{minipage}{\textwidth}
    \begin{center}
        \begin{tabular}{cc}
        \hspace{-2mm}\includegraphics[trim={0cm 0cm 0cm 0cm}, clip, width=0.99\columnwidth]{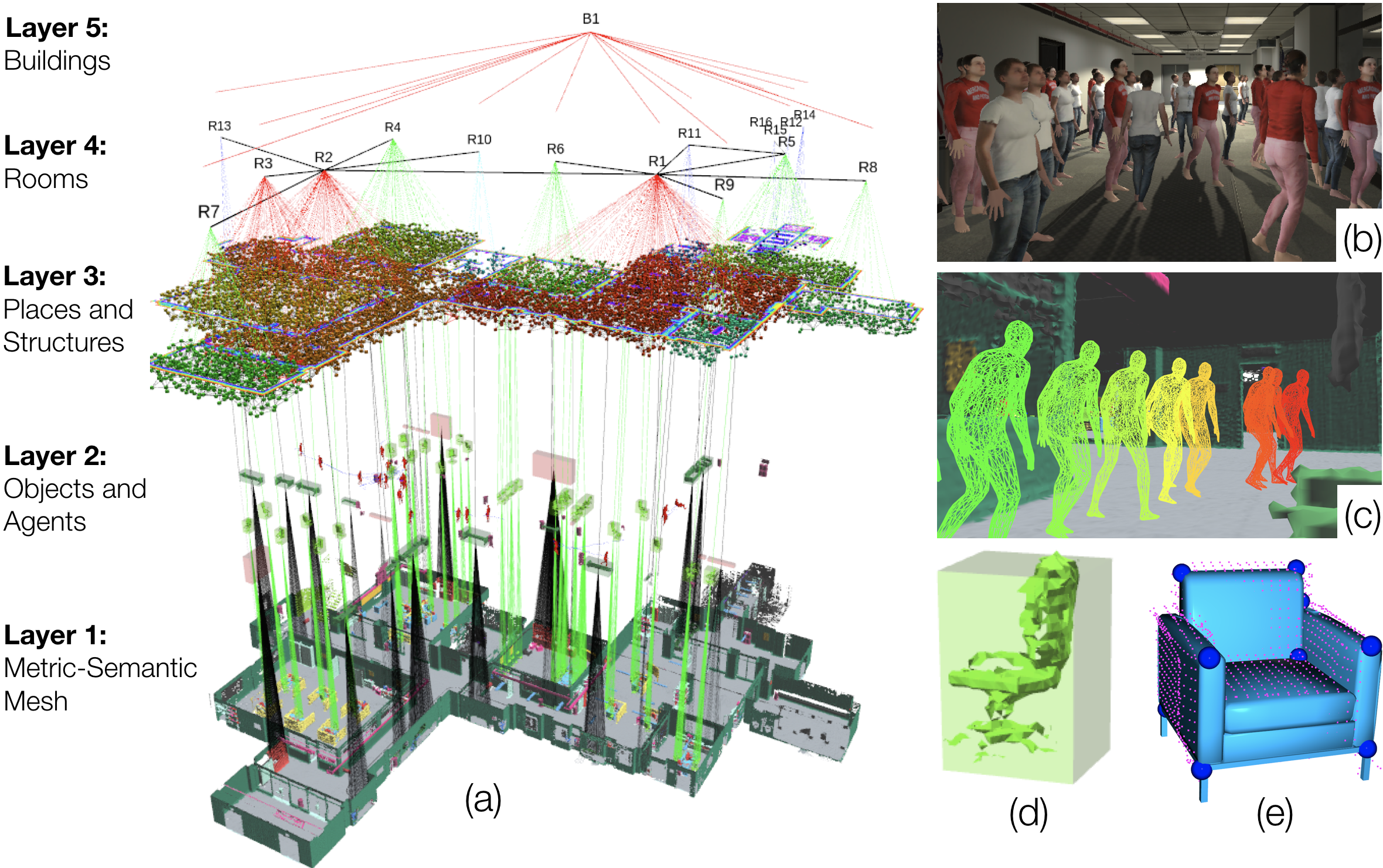}
        \end{tabular}
    \end{center}
    \vspace{-2mm}
  \end{minipage}

  \caption{
(a) A 3D {\it Dynamic Scene Graph} (\DSG) is a layered and hierarchical representation that abstracts a dense 3D model 
(\eg a metric-semantic mesh) into higher-level \emph{spatial concepts} (\eg objects, agents, places, rooms) 
and models their spatio-temporal relations (e.g., ``agent A is in room B at time $t$'').
\Kimera is the first \SPES that reconstructs a \DSG from visual-inertial data, and  
(a) segments places, structures (\eg walls), and rooms,
(b) is robust to extremely crowded environments,
(c) tracks dense mesh models of human agents in real time, 
(d) estimates centroids and bounding boxes of objects of unknown shape, 
(e) estimates the 3D pose of objects for which a CAD model is given. 
}
\label{fig:DSG}
\end{figure*}

\section{3D Dynamic Scene Graphs}
\label{sec:DSG}

A 3D \emph{Dynamic Scene Graph} (\DSG, \Cref{fig:DSG}) is an actionable spatial representation that captures the 3D geometry and semantics of a scene at different levels of abstraction,
 and models objects, places, structures, and agents and their relations.
 More formally, a \DSG is a \emph{layered directed graph} where nodes represent
  \emph{spatial concepts} (\eg objects, rooms, agents) and edges represent pairwise spatio-temporal relations (\eg ``agent A is in room B at time $t$'').

  Contrarily to \emph{knowledge bases}~\citep{Krishna92book-knowledgeBase}, spatial concepts are semantic concepts that are \emph{spatially grounded} (in other words, each node in our \DSG includes spatial coordinates and shape or bounding-box information as attributes).
  A \DSG is a \emph{layered} graph, \ie nodes are grouped into layers that correspond to different levels of abstraction.  Every node has a unique ID.

The \DSG of a single-story indoor environment includes 5 layers (from low to high abstraction level):
(i) \layerOneName,
(ii) \layerTwoName,
(iii) \layerThreeName,
(iv) \layerFourName,
and
(v) \layerSixName.
We discuss each layer and the corresponding nodes and edges below.

\subsection{Layer 1: \layerOneName} %
\label{sec:layer-mesh}

The lower layer of a \DSG is a semantically annotated 3D mesh (bottom of \Cref{fig:DSG}(a)).
The nodes in this layer are 3D points (vertices of the mesh) and each node has the following attributes:
(i) 3D position, (ii) normal, (iii) RGB color, and (iv) a panoptic semantic label.\footnote{Panoptic segmentation~\citep{Kirillov19cvpr-panopticSegmentation,Li18arxiv-thingsAndStuff} segments both objects (\eg chairs, tables, drawers) and structures (\eg walls, ground, ceiling).}
Edges connecting triplets of points (\ie a clique with 3 nodes) describe faces in the mesh and define the
topology of the environment. %
Our metric-semantic mesh includes everything in the environment that is \emph{static}, while for storage convenience we
store meshes of dynamic objects in a separate structure (see ``Agents'' below).

\newcommand{\myhspace}{\hspace{-6mm}}
\newcommand{\mpw}{4.5cm}
\newcommand{\myRate}{1}

\begin{figure}[htbp]
	\begin{center}
		\begin{minipage}{\columnwidth}
		\includegraphics[trim=0mm 1mm 0mm 0mm, clip, width=\myRate\columnwidth]{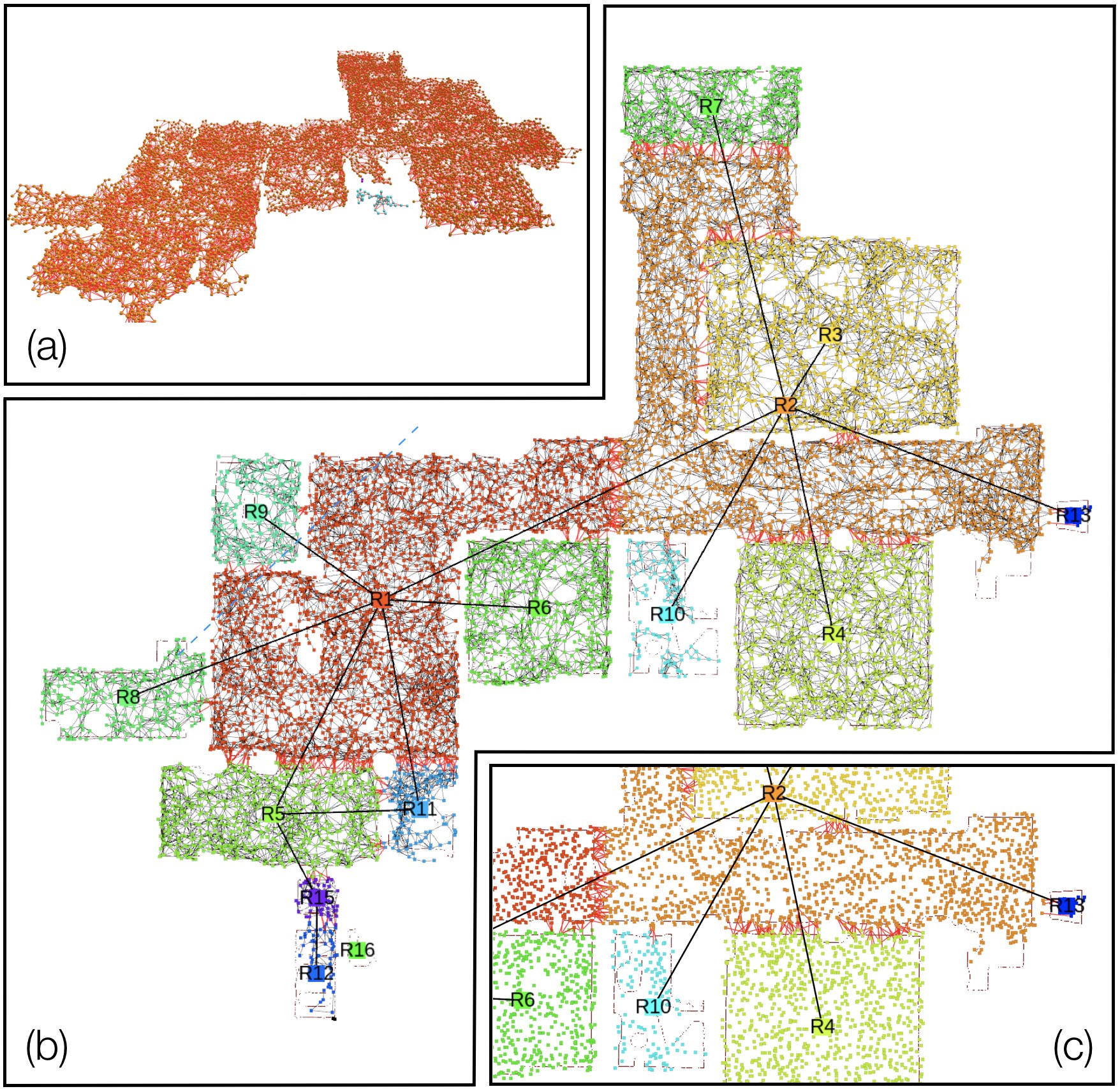} \\
		\end{minipage}
		\caption{
		{\bf Places} and their connectivity shown as a graph.
		(a) Skeleton (places and topology) produced by~\citep{Oleynikova18iros-topoMap} (side view);
		(b) {\bf Room} parsing produced by our approach 
		(top-down view);
		(c) Zoomed-in view; red edges connect different rooms.
		}
		\label{fig:rooms}
	\end{center}
\end{figure}

\subsection{Layer 2: \layerTwoName}
\label{sec:layer-objectsAgents}

This layer contains two types of nodes: objects and agents (\Cref{fig:DSG}(c-e)), whose main distinction is the fact that agents are time-varying entities, while
objects are static.\footnote{The distinction between objects and agents is only made for the sake of the presentation.
The \DSG provides a unified approach to model static and dynamic entities,
since the latter only requires storing pose information over time.}

{\bf Objects} represent static elements in the environment that are not considered \emph{structural} (\ie walls, floor, ceiling, pillars are considered \emph{structure} and are not modeled in this layer).
Each object is a node and node attributes include (i) a 3D object pose,
\veryOptional{\footnote{3D poses of static objects are sometimes called ``Static Coordinate Systems'' in  graphics~\cite{Rohlf94accgit-iris}.}} (ii) a bounding box, and (ii) its semantic class (\eg chair, desk).
While not investigated in this paper, we refer the reader to~\citep{Armeni19iccv-3DsceneGraphs} for a more comprehensive list of attributes, including materials and affordances.
Edges between objects describe relations, such as  co-visibility, relative size, distance, or contact (``the cup is on the desk'').
Each object node is connected to the corresponding set of points belonging to the object in the \layerOneName.
Moreover, each object is connected to the nearest reachable \emph{place} node (see \Cref{sec:layer-places}).

{\bf Agents} represent dynamic entities in the environment, including humans.
In general, there might be many types of dynamic entities 
(\eg animals, vehicles, or bicycles in outdoor environments).
\rebuttal{In this paper,}
we focus on two classes: \emph{humans} and \emph{robots}.
\footnote{These classes can be considered instantiations of more general concepts:
``rigid'' agents (such as robots, for which we only need to keep track a 3D pose), and ``deformable'' agents
(such as humans, for which we also need to keep track of a
time-varying shape).} 
\rebuttal{Our approach to track dynamic agents 
relies solely on defining which labels
are considered to be dynamic 
in the semantic segmentations of the 2D images.
}

 Both human and robot nodes have three attributes:
 (i) a 3D pose graph describing their trajectory over time,
 \veryOptional{\footnote{3D poses of dynamic objects are sometimes called ``Dynamic Coordinate Systems'' in  graphics~\cite{Rohlf94accgit-iris}.}}
 (ii) a mesh model describing their (non-rigid) shape, and
 (iii) a semantic class (\ie human, robot).
 A pose graph~\citep{Cadena16tro-SLAMsurvey} is a collection of time-stamped 3D poses where edges
 model pairwise relative measurements.
The robot collecting the data is also modeled as an agent in this layer.

\subsection{Layer 3: \layerThreeName}
\label{sec:layer-places}

This layer contains two types of nodes: places and structures.
Intuitively, places are a model for the free space, while structures capture separators between different spaces.

{\bf Places} (\Cref{fig:rooms}) correspond to positions in the free-space and edges between places represent traversability (in particular: presence of a straight-line path between places).
Places and their connectivity form a \emph{topological map}~\citep{Ranganathan04iros,Remolina04} that can be used for path planning.
\optional{Moreover, place nodes provide more granularity in the scene description (\eg the \DSG can differentiate the back of the room from its front).} %
Place attributes only include a 3D position, but can also include a semantic class (\eg back or front of the room) and an obstacle-free bounding box around the place position.
Each object and agent in Layer 2 is connected with the nearest place (for agents, the connection is for each time-stamped pose, since
agents move from place to place).
Places belonging to the same room are also connected to the same room node in Layer~4.
\Cref{fig:rooms}(b-c) shows a visualization with places color-coded by rooms. %

{\bf Structures} (\Cref{fig:structure}) include nodes describing structural elements in the environment, \eg walls, floor, ceiling, pillars.
The notion of structure captures elements often called  ``stuff'' in related work~\citep{Li18arxiv-thingsAndStuff}.
Structure nodes' attributes are: (i) 3D pose, (ii) bounding box, and (iii) semantic class (\eg walls, floor).
Structures may have edges to the rooms they enclose.
Structures may also have edges to an object in Layer 3, \eg a ``frame'' (object) ``is hung'' (relation) on
a ``wall'' (structure), or a ``ceiling light is mounted on the ceiling''.

\renewcommand{\myhspace}{\hspace{-6mm}}
\renewcommand{\mpw}{4.5cm}
\renewcommand{\myRate}{1}

\begin{figure}[h]
	\begin{center}
	\begin{minipage}{\columnwidth}
	\includegraphics[trim=0mm 5mm 0mm 5mm, clip,width=0.9\myRate\columnwidth]{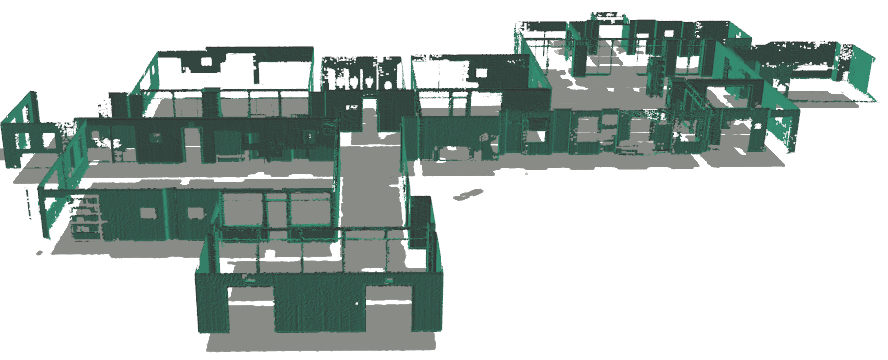} \\
	\includegraphics[trim=0mm 5mm 0mm 5mm, clip,width=0.9\myRate\columnwidth]{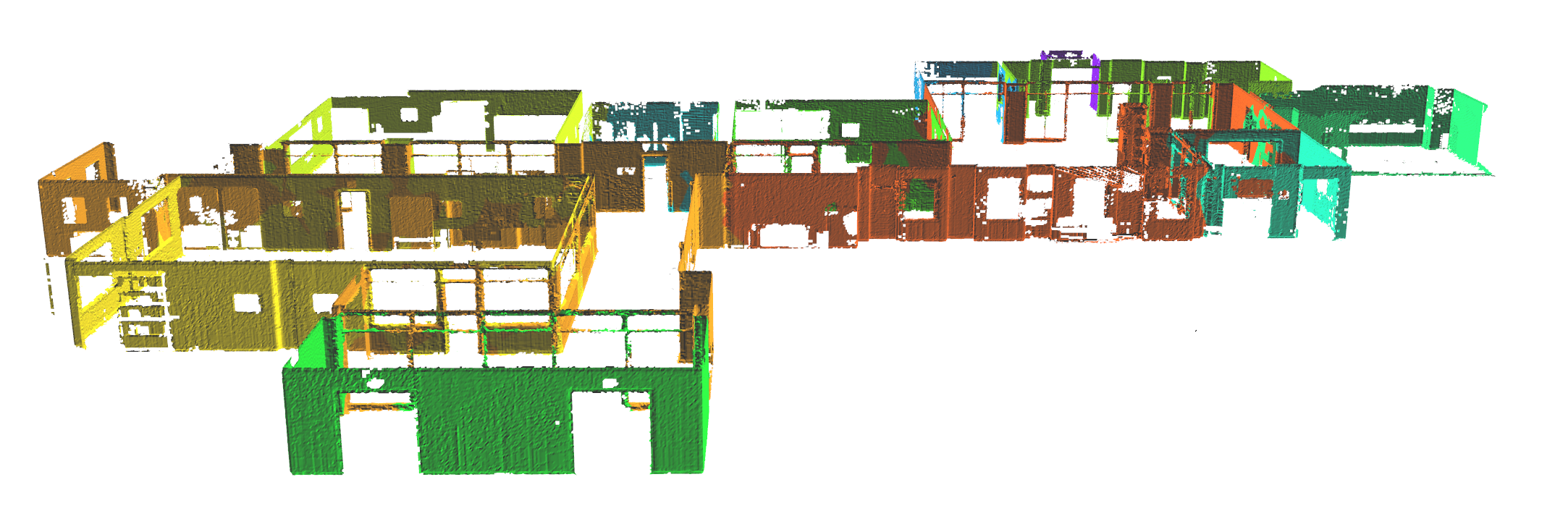} \\
	\end{minipage}
	\vspace{-5mm} 
	\caption{{\bf Structures}: exploded view of walls and floor (top). Segmented walls according to the room id (bottom). 
	 \label{fig:structure}}
	\vspace{-8mm} 
	\end{center}
\end{figure}

\subsection{Layer 4: \layerFourName} %
\label{sec:layer-rooms}

This layer includes nodes describing rooms, corridors, and halls.
Room nodes (\Cref{fig:rooms}) have the following attributes: (i) 3D pose, (ii) bounding box, and (iii) semantic class (\eg kitchen, dining room, corridor).
Two rooms are connected by an edge if they are adjacent (\ie there is a door connecting them).
A room node has edges to the %
places (Layer 3) it contains (since each place is connected to nearby objects, the \DSG also captures which object/agent is contained in each room).
All rooms are connected to the building they belong to (Layer 5).

\subsection{Layer 5: \layerSixName} %
\label{sec:layer-building}

Since we are considering a representation over a single building, there is a single \emph{building node}
with the following attributes: (i) 3D pose, (ii) bounding box, and (iii) semantic class (\eg office building,
residential house). The building node has edges towards all rooms in the building.

\subsection{Composition and Queries} %

\emph{Why should we choose this set of nodes or edges rather than a different one?}
 Clearly, the choice of nodes in the \DSG is not unique and is task-dependent.
  Here we first motivate our choice of nodes  in terms of \emph{planning queries} the \DSG is designed for
  (see Remark~\ref{rmk:queries} and the broader discussion in \Cref{sec:examples}), and we then show that the representation is compositional, in the sense that
  it can be easily expanded to encompass more layers, nodes, and edges (Remark~\ref{rmk:composition}).

\begin{remark}[Planning Queries]
\label{rmk:queries}

The proposed \DSG is designed with  task and motion planning queries in mind.
The semantic node attributes (\eg semantic class) support planning from high-level specification
(``pick up the red cup from the table in the dining room'').
The geometric node attributes (\eg meshes, positions, bounding boxes)
and the edges are used for motion planning. For instance,
the places can be used as a topological graph for path planning, and the bounding boxes can be used for fast
collision checking. \optional{A more extensive discussion about \DSG queries is postponed to \Cref{sec:discussion}.}

\end{remark}

\begin{remark}[Composition of \DSGs]
\label{rmk:composition}

A second re-ensuring property of a \DSG is its compositionality:
  one can easily concatenate more layers at the top and the bottom of the \DSG in \Cref{fig:DSG}(a), and
  even add intermediate layers.
  For instance, in a multi-story building, we can include a ``Level'' layer between the ``Building'' and ``Rooms'' layers in
  \Cref{fig:DSG}(a). Moreover, we can add further abstractions or layers at the top, for instance going from
  buildings to neighborhoods, and then to cities.
\end{remark}


\begin{figure*}[htbp]
  \centering
  \begin{minipage}{\textwidth}
    \begin{center}
      \begin{tabular}{cc}
        \hspace{-2mm}\includegraphics[trim={0cm 0cm 0cm 0cm}, clip, width=0.98\columnwidth]{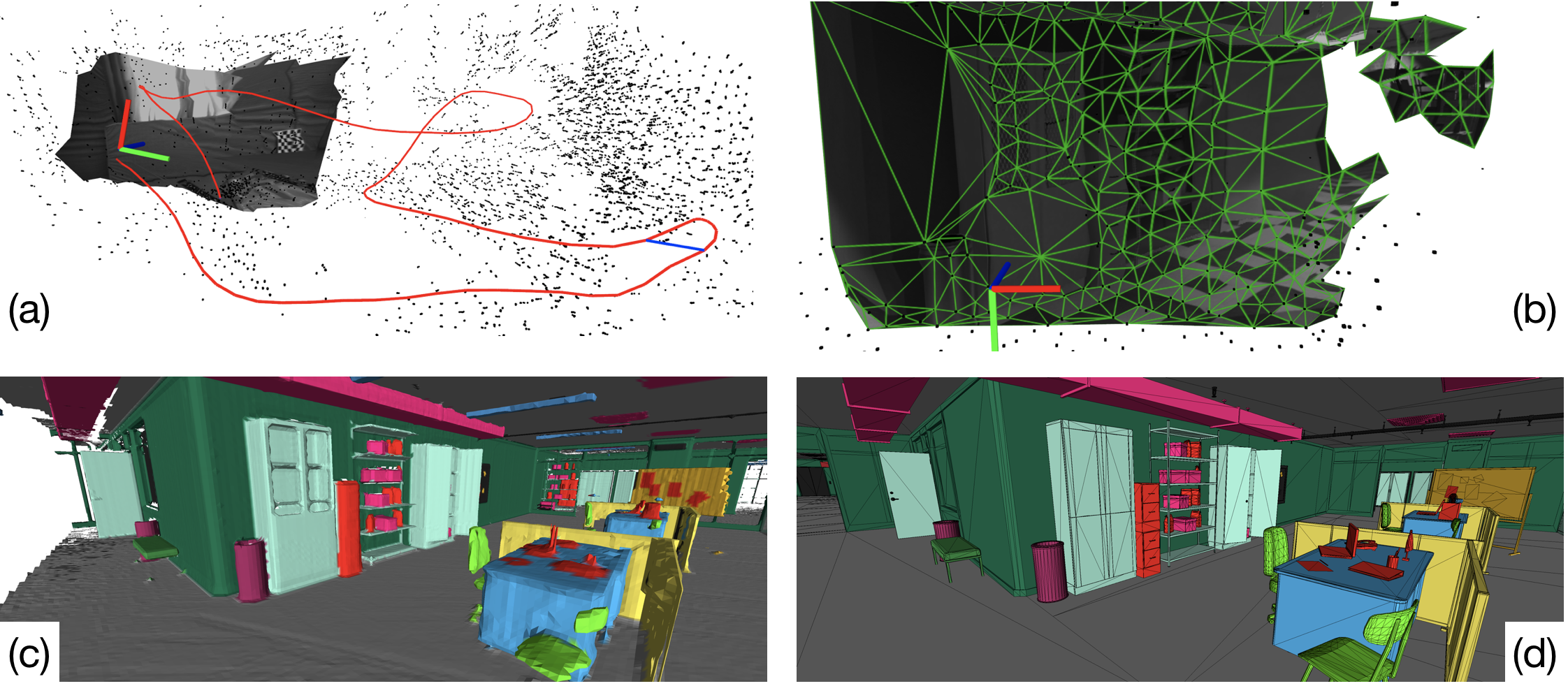}
      \end{tabular}
    \end{center}
    \vspace{-2mm}
  \end{minipage}
  \caption{
  \KimeraCore is an open-source library for real-time metric-semantic SLAM.
  It provides (a) visual-inertial state estimates at IMU rate (\KimeraVIO),
  and a globally consistent and outlier-robust trajectory estimate (\KimeraRPGO), 
  computes (b) a low-latency local mesh of the scene (\KimeraMesher), 
  and builds (c) a semantically annotated 3D mesh (\KimeraSemantics),
  which can be optimized for global consistency (\KimeraPGMO) and accurately reflects the ground truth model (d).
  }
  \label{fig:semantic_mesh}
\end{figure*}

\begin{figure*}[t!]
    \centering

    \includegraphics[trim=0.2cm 1.8cm 1mm 1.5cm, clip, width=\textwidth]{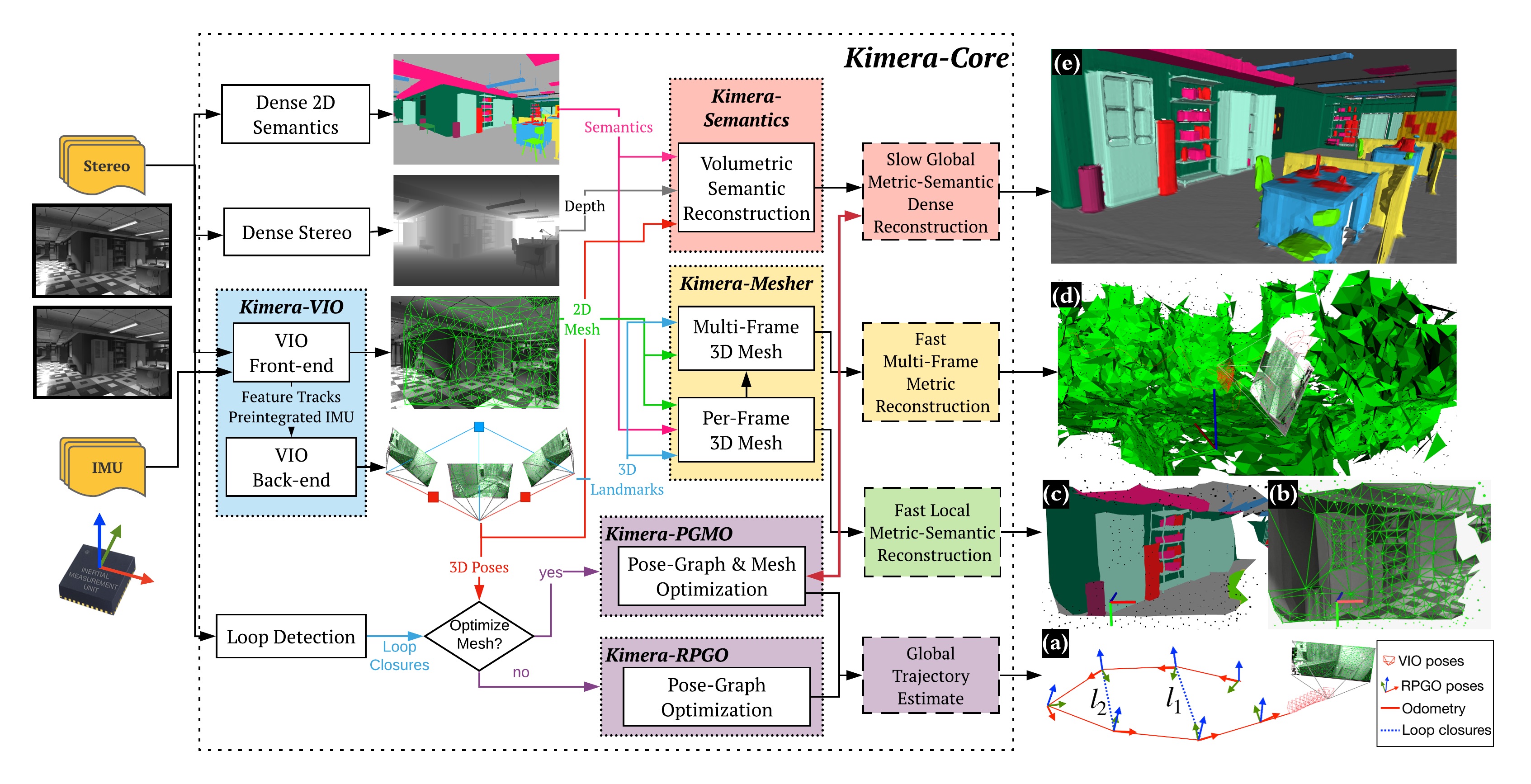}
    \caption{\KimeraCore's architecture. \KimeraCore uses stereo images (or RGB-D) and IMU data as input (shown on the left) 
    	and outputs (a) pose estimates and (b-e) multiple metric-semantic reconstructions. 
        \KimeraCore has four key modules: \KimeraVIO, \KimeraPGMO (alternatively, \KimeraRPGO), \KimeraMesher, and \KimeraSemantics.
    \label{fig:kimera_diagram}  \vspace{-5mm}}    
\end{figure*}

\begin{figure*}[htbp]
  \centering
  \includegraphics[trim={1.00cm 1.30cm 1.00cm 1.00cm}, clip, width=\textwidth]{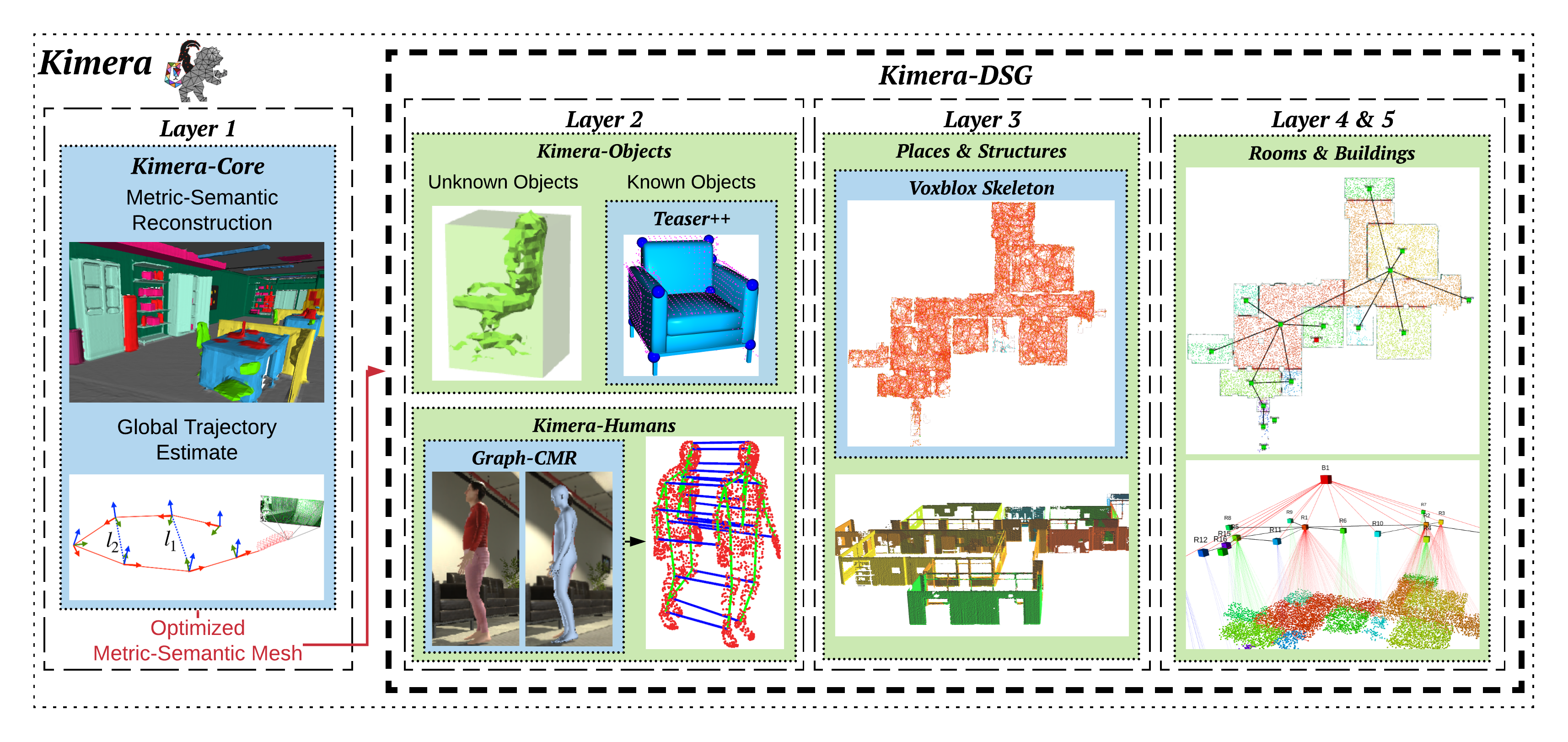}
  \caption{\Kimera's architecture, with \KimeraCore and \KimeraDSG as sub-modules.
  \KimeraCore generates a globally consistent 3D metric-semantic mesh (\Cref{fig:kimera_diagram}) that represents the first layer of the \DSG 
  and is further used by \KimeraDSG to build the subsequent layers.
  \KimeraDSG further comprises three key modules: \KimeraObjects, \KimeraHumans, 
  and \KimeraBuildingParser (which generates layers 3 to 5).}
  \label{fig:kimera_architecture}
\end{figure*}

\section{\Kimera: \SPESlong}
\label{sec:Kimera}

This section describes \Kimera, our \emph{\SPESlong}, that populates the \DSG nodes and edges using sensor data.
 The input to \Kimera is streaming data from a stereo or RGB-D camera, and an Inertial Measurement Unit (IMU).
The output is a 3D \DSG. 
In our current implementation, the metric-semantic mesh and the agent nodes are incrementally built from sensor data in real-time,
while the remaining nodes (objects, places, structure, rooms) are automatically built at the end of the run.

\myParagraph{\KimeraCore}
We use \KimeraCore~\citep{Rosinol20icra-Kimera} to reconstruct a semantically annotated 3D mesh from visual-inertial data in real-time (\Cref{fig:semantic_mesh}).
\KimeraCore is open source and includes four main modules: 
(i) \KimeraVIO: a visual-inertial odometry module implementing IMU preintegration and fixed-lag smoothing~\citep{Forster17tro}, 
(ii) \KimeraPGMO: a robust pose graph and mesh optimizer, that generalizes \KimeraRPGO, which only optimized the pose graph.
(iii) \KimeraMesher: a per-frame and multi-frame mesher~\citep{Rosinol19icra-mesh}, 
and 
(iv) \KimeraSemantics: a volumetric approach to produce a semantically annotated mesh and an Euclidean Signed Distance Function (ESDF)
 based on Voxblox~\citep{Oleynikova2017iros-voxblox}.
 \KimeraSemantics uses a 2D semantic segmentation of the camera images to label the 3D mesh using Bayesian updates. 
We take the metric-semantic mesh produced by \KimeraSemantics and optimized by \KimeraPGMO as Layer 1 in the \DSG in \FigFrontCover(a).

\Cref{fig:kimera_diagram} shows \KimeraCore's architecture. 
\Kimera takes stereo frames and high-rate inertial measurements as input and returns 
(i) a highly accurate state estimate at IMU rate, 
(ii) a globally-consistent trajectory estimate, and
(iii) multiple meshes of the environment, including a fast local mesh and a global 
semantically annotated mesh. 
\KimeraCore is heavily parallelized and uses five threads to accommodate inputs and outputs at different rates 
(\eg IMU, frames, keyframes). Here we describe the architecture \emph{by threads}, while the description of each module is given in the following sections.

The first thread includes the \KimeraVIO front-end (\Cref{sec:vio}) that takes stereo images and IMU data, 
and outputs feature tracks and preintegrated IMU measurements.
 The front-end also publishes IMU-rate state estimates.
The second thread runs \KimeraVIO's back-end, and outputs optimized state estimates (most importantly, the robot's 3D pose).
The third thread runs \KimeraMesher (\Cref{sec:mesher}), that computes low-latency ($<\!\perFrameMeshLatency$) per-frame and multi-frame 3D meshes. 
These three threads allow creating the per-frame mesh in~\Cref{fig:kimera_diagram}(b) (which can also come with semantic labels as in~\Cref{fig:kimera_diagram}(c)), 
as well as the multi-frame mesh in~\Cref{fig:kimera_diagram}(d).
The next two threads operate at slower rates and are designed to support low-frequency functionalities, such as path planning.
The fourth thread includes \KimeraSemantics (\Cref{sec:semantic}), that uses a depth map, from RGB-D or dense-stereo, 
and 2D semantic labels to obtain a metric-semantic mesh, using the pose estimates from \KimeraVIO.
The last thread includes \KimeraPGMO (\Cref{sec:meshLoopClosure}), that uses the detected loop closures, together with \KimeraVIO's pose estimates and \KimeraSemantics' 3D metric-semantic mesh,
to estimate a globally consistent trajectory (\Cref{fig:kimera_diagram}(a)) and 3D metric-semantic mesh (\Cref{fig:kimera_diagram}(e)).
As shown in \Cref{fig:kimera_diagram}, \KimeraRPGO can be used instead of \KimeraPGMO if the optimized 3D metric-semantic mesh is not required.

\myParagraph{\KimeraDSG} 
Here we describe \KimeraDSG's architecture layer by layer,
 while the description of each module is given in the following sections.
We use \KimeraDSG to build the \DSG from the globally consistent 3D metric-semantic mesh generated by \KimeraCore,
 which represents the \DSG's first layer, as shown in \Cref{fig:kimera_architecture}.
Then, \KimeraDSG builds the second layer containing objects and agents.
For the objects, \KimeraObjects (Section~\ref{sec:objects}) 
either estimates a bounding box for the objects of unknown shape
or fits a CAD model for the objects of known shape using \TEASERpp~\citep{Yang20tro-teaser}.
\KimeraHumans (\Cref{ssec:humans}) reconstructs dense meshes of humans in the scene using \MeshName~\citep{Kolotouros19cvpr-shapeRec},
 and estimates their trajectories using a pose graph model.
Then, \KimeraBuildingParser (Section~\ref{sec:buildingParser}) generates the remaining three layers.
It first generates layer 3 by parsing the metric-semantic mesh to identify structures (\ie walls, ceiling),
and further extracts a topological graph of places using \citep{Oleynikova18iros-topoMap}.
Then, \KimeraBuildingParser generates layer 4 by segmenting layer 3 into rooms,
 and generates layer 5 by further segmenting layer 4 into buildings.
    
\subsection{\KimeraVIO: Visual-Inertial Odometry}
\label{sec:vio}

\KimeraVIO implements the keyframe-based maximum-a-posteriori visual-inertial estimator presented in~\citep{Forster17tro}.
 In our implementation, the estimator can perform both \emph{full} smoothing or \emph{fixed-lag} smoothing, depending on the 
 specified time horizon; we typically use the latter to bound the estimation time. 
 \KimeraVIO includes a (visual and inertial) front-end which is in charge of processing the raw sensor data, and a 
 back-end, that fuses the processed measurements to obtain an estimate of the state of the sensors (\ie pose, velocity, and sensor biases). %

\myParagraph{VIO Front-end} 
Our IMU front-end performs on-manifold preintegration~\citep{Forster17tro} to obtain compact preintegrated measurements of the 
relative state between two consecutive keyframes from raw IMU data. 
The vision front-end detects Shi-Tomasi corners~\citep{Shi94}, 
tracks them across frames using the Lukas-Kanade tracker~\citep{Bouguet00-lktracking},
 finds left-right stereo matches, and performs geometric verification. 
We perform both mono(cular) verification using 5-point RANSAC~\citep{Nister04pami} and stereo verification using 
 3-point RANSAC~\citep{Horn87josa}; the code also offers the option to use the IMU rotation 
 and perform mono and stereo verification 
 using  2-point~\citep{Kneip11bmvc} and 1-point RANSAC, respectively.  
Since our robot moves in crowded (dynamic) environments,
we seed the Lukas-Kanade tracker with an initial guess (of the location 
of the corner being tracked) given by the rotational optical flow estimated from the IMU,
similar to~\citep{Hwangbo09iros-IMUKLT}.
Moreover, we default to using 2-point (stereo) and 1-point (mono) RANSAC, which uses the IMU rotation to prune outlier
correspondences in the feature tracks.
Feature detection, stereo matching, and geometric verification are executed at each \emph{keyframe}, while 
we track features at intermediate \emph{frames}.
  
\myParagraph{VIO Back-end} 
At each keyframe, preintegrated IMU and visual measurements are added to a fixed-lag smoother (a factor graph) which constitutes our VIO back-end. 
We use the preintegrated IMU model and  the structureless vision model of~\citep{Forster17tro}.
The factor graph is solved using iSAM2~\citep{Kaess12ijrr} in GTSAM~\citep{Dellaert12tr}.
At each iSAM2 iteration, 
the structureless vision model estimates the 3D position of the observed features using DLT~\citep{Hartley04book}
 and analytically eliminates the corresponding 3D points from the VIO state~\citep{Carlone14icra-smartFactors}. 
Before elimination, degenerate points (\ie points behind the camera or without enough parallax for triangulation) and outliers (\ie points with large reprojection error) are removed, providing an extra robustness layer. 
Finally, states that fall out of the 
smoothing horizon are marginalized out using GTSAM. %

\subsection{\KimeraMesher: 3D Mesh Reconstruction}
\label{sec:mesher}

\KimeraMesher can quickly generate two types of 3D meshes: (i) a per-frame 3D mesh, and (ii) a multi-frame 3D mesh spanning
 the keyframes in the VIO fixed-lag smoother.

\myParagraph{Per-frame mesh} 
 As in \citep{Rosinol19icra-mesh}, we first perform a 2D Delaunay triangulation over the successfully tracked 2D features (generated by the VIO front-end) in the current keyframe. %
Then, we back-project the 2D Delaunay triangulation to generate a 3D mesh (\Cref{fig:kimera_diagram}(b)),
 using the 3D point estimates from the VIO back-end. 
 While the per-frame mesh is designed to provide low-latency obstacle detection, we 
 also provide the option to semantically label the resulting mesh, by texturing the mesh with 2D labels (\Cref{fig:kimera_diagram}(c)).

\myParagraph{Multi-frame mesh} 
 The multi-frame mesh fuses the per-frame meshes collected over the VIO receding horizon into a single mesh (\Cref{fig:kimera_diagram}(d)). 
Both per-frame and multi-frame 3D meshes are encoded as a list of vertex positions, together with a list of triplets of vertex IDs to describe the triangular faces.
Assuming we already have a multi-frame mesh at time $t-1$, for each new per-frame 3D mesh that we generate (at time $t$), we loop over its vertices and triplets and add vertices and triplets that are in the per-frame mesh but are missing in the multi-frame one.
Then we loop over the multi-frame mesh vertices and update their 3D position according to the latest VIO back-end estimates.
Finally, we remove vertices and triplets corresponding to old features observed outside the VIO time horizon.
The result is an up-to-date 3D mesh spanning the keyframes in the current VIO time horizon. 
If planar surfaces are detected in the mesh, \emph{regularity factors}~\citep{Rosinol19icra-mesh} are added to the VIO back-end,
 which results in a tight coupling between VIO and mesh regularization, see~\citep{Rosinol19icra-mesh} for further details.

\subsection{\KimeraSemantics: 3D Metric-Semantic Reconstruction}
\label{sec:semantic}

We adapt the \textit{bundled raycasting} technique introduced by \cite{Oleynikova2017iros-voxblox} to 
(i) build an accurate global 3D mesh (covering the entire trajectory), 
and (ii) semantically annotate the mesh. 

\myParagraph{Global mesh}
\mbox{Our implementation builds on Voxblox} \citep{Oleynikova2017iros-voxblox} and 
uses a voxel-based (TSDF) model to filter out noise and extract the global mesh.
At each keyframe, we obtain depth maps using dense stereo (semi-global matching~\citep{Hirschmuller08pami}) to obtain a 3D point cloud, or from RGB-D if available.
Then, we run bundled raycasting using Voxblox~\citep{Oleynikova2017iros-voxblox}.
This process is repeated at each keyframe and produces a TSDF, from which a mesh is extracted using marching cubes~\citep{Lorensen87siggraph-marchingCubes}. 

\myParagraph{Semantic annotation}
\KimeraSemantics uses 2D semantically labeled images (produced at each keyframe) to semantically annotate the global mesh;
the 2D semantic labels can be obtained using off-the-shelf tools for pixel-level 2D semantic segmentation,
 \eg deep neural networks~\citep{Huang19arxiv,Zhang19bmvc,Chen17pami,Zhao17cvpr,Yang18eccv,Paszke16arxiv,Ren15nips-RCNN,He17iccv-maskRCNN,Hu18cvpr-maskXRCNN}.
In our real-life experiments, we use Mask-RCNN~\citep{He17iccv-maskRCNN}.
Then, during the bundled raycasting, we also propagate the semantic labels. 
Using the 2D semantic segmentation, we attach a label to each 3D point produced by dense stereo.
Then, for each bundle of rays in the {bundled raycasting}, 
we build a vector of label probabilities from the frequency of the observed labels in the bundle.
We then propagate this information along the ray only within the TSDF truncation distance (\ie near the surface) to spare computation.
In other words, we spare the computational effort of updating probabilities for the ``empty'' label.
While traversing the voxels along the ray, we use a Bayesian update to estimate the posterior label probabilities at each voxel, similar to~\citep{McCormac17icra-semanticFusion}.
After bundled semantic raycasting, each voxel has a vector of label probabilities, from which we extract the most likely label.
The metric-semantic mesh is finally also extracted using marching cubes~\citep{Lorensen87siggraph-marchingCubes}.
The resulting mesh is significantly more accurate than the multi-frame mesh of \Cref{sec:mesher},
 but it is slower to compute ($\approx \globalMeshLatency$, see \Cref{ssec:timing_performance}).

\subsection{\KimeraPGMO: Pose Graph and Mesh Optimization with Loop Closures}
\label{sec:meshLoopClosure}

The mesh from \KimeraSemantics is built from the poses from \KimeraVIO and drifts over time.
The loop closure module detects loop closures to correct the global trajectory and the mesh.
The mesh is corrected via a deformation, since this is more scalable compared to rebuilding
the mesh from scratch or using `de-integration'~\citep{Dai17tog-bundlefusion}.
This is achieved via a novel simultaneous pose graph and mesh deformation approach which utilizes an
embedded deformation graph that optimizes in a single run the environment and the robot trajectory.
The optimization is formulated as a factor graph in GTSAM.
In the following, we review the individual components.

\myParagraph{Loop Closure Detection}
The loop closure detection relies on the DBoW2 library~\citep{Galvez12tro-dbow} and uses a bag-of-word
representation with ORB descriptors to quickly detect putative loop closures.
For each putative loop closure, we reject outlier loop closures using mono 5-point RANSAC~\citep{Nister04pami}
 and stereo 3-point RANSAC~\citep{Horn87josa} geometric verification, and pass the remaining loop closures to the outlier rejection and pose solver.
Note that the resulting loop closures can still contain outliers due to perceptual aliasing (\eg two identical rooms on
different floors of a building).
\rebuttal{While most open-source SLAM algorithms,
such as ORB-SLAM3~\citep{Campos21-TRO}, VINS-Mono~\citep{Qin18tro-vinsmono}, Basalt~\citep{Usenko19ral-basalt},
are overly cautious when accepting loop-closures, by fine-tuning DBoW2 for example,
we instead make our backend robust to outliers, as we explain below.}

\myParagraph{Outlier Rejection}
We filter out bad loop closures with a modern outlier rejection method, \emph{Pairwise
Consistent Measurement Set Maximization} (PCM)~\citep{Mangelson18icra}, that we
tailor to a single-robot and online setup. We store separately the odometry edges (produced by \KimeraVIO)
and the loop closures (produced by the loop closure detection); each time a loop closure is detected, we
select inliers by finding the largest set of consistent loop closures using a modified version of PCM.

\begin{figure}[htbp]
  \centering
  \includegraphics[width=\columnwidth]{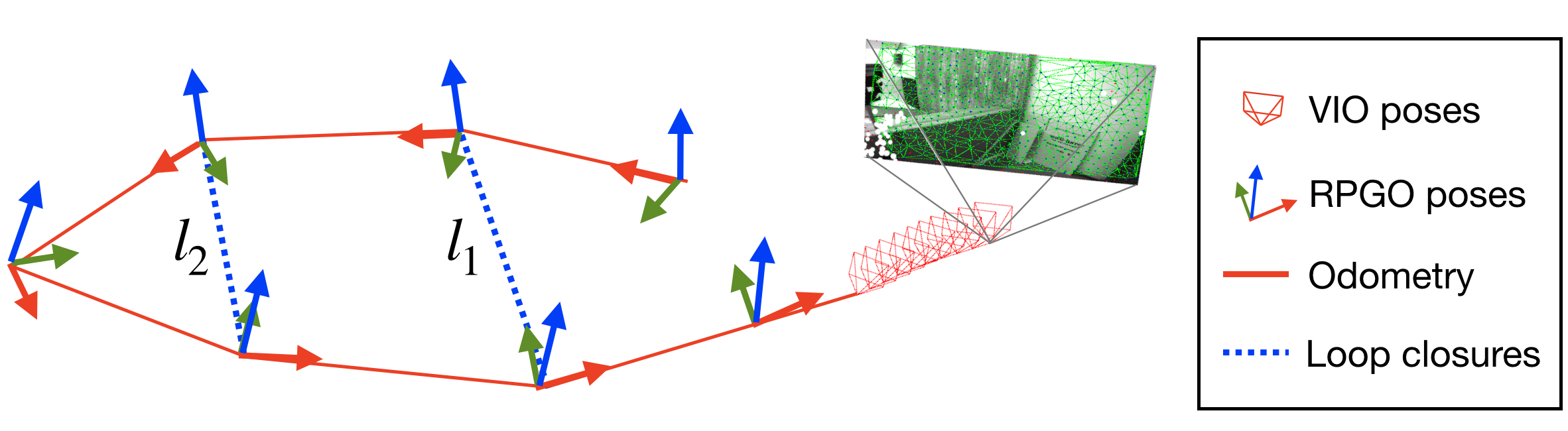}
  \caption{\KimeraRPGO detects visual loop closures, rejects spurious loop closures, and estimates a globally consistent trajectory. In contrast to \KimeraPGMO, \KimeraRPGO does not optimize the 3D mesh.}
  \label{fig:RPGO}
\end{figure}

The original PCM is designed for the multi-robot case and only checks that inter-robot loop closures are consistent.
We developed an implementation of PCM that (i) adds an \emph{odometry consistency check} on the loop closures and
(ii) \emph{incrementally} updates the set of consistent measurements to enable online operation.
The odometry check verifies that each loop closure (\eg $l_1$ in \Cref{fig:RPGO}) is consistent with the odometry (in red in the figure):
in the absence of noise, the poses along the cycle formed by the odometry and the loop $l_1$ must compose to the identity.
As in PCM, we flag as outliers loops for which the error accumulated along the cycle is not consistent with the measurement noise using a Chi-squared test.
If a loop detected at the current time $t$ passes the odometry check, we test if it is pairwise consistent with previous loop closures as in~\citep{Mangelson18icra}
(\eg check if loops $l_1$ and $l_2$ in \Cref{fig:RPGO} are consistent with each other).
While PCM~\citep{Mangelson18icra} builds an adjacency matrix $\MA \in \Real{L \times L}$ from scratch to keep track of pairwise-consistent loops
(where $L$ is the number of detected loop closures), we enable online operation by building the matrix $\MA$ incrementally.
Each time a new loop is detected, we add a row and column to the matrix $\MA$ and only test the new loop against the previous ones.
Finally, we use the fast maximum clique implementation of~\citep{Pattabiraman15im-maxClique} to compute the largest set of consistent loop closures.
The set of consistent measurements are added to the pose graph (together with the odometry).

\begin{figure}[h]
\centering
\includegraphics[width=\columnwidth,trim=0mm 0mm 0mm 0mm,clip]{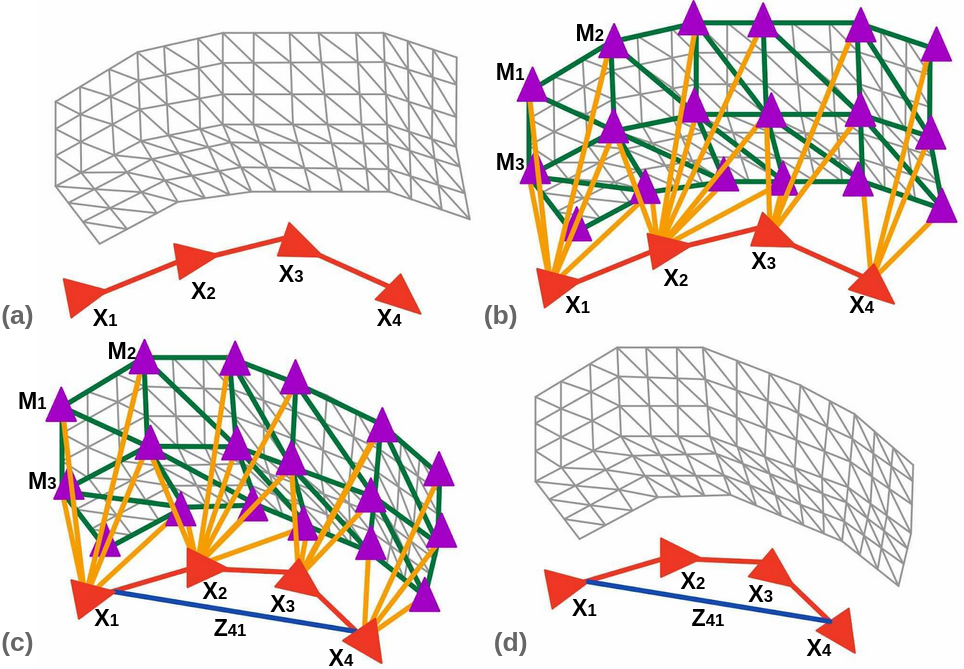} \\
\caption{\KimeraPGMO's mesh deformation and pose graph optimization.
(a) shows the received mesh and the pose graph with no loop closures detected yet.
(b) shows the creation of the deformation graph where the red vertices are the pose vertices with associated transform $\MX_i$ and purple vertices are the mesh vertices with associated transform $\MM_i$.
The green edges are the edges describing the connectivity of the simplified mesh, which are also the edges connecting the mesh vertices to each other in the deformation graph.
The yellow edges are the edges connecting the pose vertices to the mesh vertices based on visibility from the camera.
(c) shows the deformation that happens when a loop closure (the blue edge) between pose graph node 4 and node 1 ($\MZ_{41}$) is added.
 $\MX_i$ and $\MM_i$ have been updated based on the optimization results.
(d) shows the optimized mesh and pose graph.
    \label{fig:deformation-graph}}
\end{figure}

\myParagraph{Pose Graph and Mesh Optimization}
When a loop closure passes the outlier rejection, either \KimeraRPGO optimizes the pose graph of the robot trajectory (\Cref{fig:RPGO}) or
\KimeraPGMO simultaneously optimizes the mesh and the trajectory (\Cref{fig:deformation-graph}).
The user can select the solver depending on the computational considerations and the need for a consistent map;
see Remark~\ref{rmk:rpgo-pgmo}.
Note that \KimeraPGMO is a strict generalization of \KimeraRPGO as described in~\citep{Rosinol20icra-Kimera}.
For \KimeraPGMO, the deformation of the mesh induced by the loop closure is based on deformation graphs~\citep{Summer07siggraph-embeddedDeformation}.
In our approach, we create a unified deformation graph including a simplified mesh and a pose graph of robot poses.
We simplify the mesh with an online vertex clustering method by storing the vertices of the mesh in an octree data structure;
as the mesh grows, the vertices in the same voxel of the octree are merged and degenerate faces and edges are removed.
The voxel size is tuned according to the environment or the dataset ($1$ to $4$ meters in our tests).

We add two types of vertices to the deformation graph: mesh vertices and pose vertices.
The mesh vertices correspond to the vertices of the simplified mesh
and have an associated transformation $\MM_k = \begin{bmatrix}\MR^M_k & \vt^M_k \\ \mathbf{0}^\top & 1 \end{bmatrix}$ for some mesh vertex $k$.
When the mesh is not yet deformed, $\MR^M_k = \eye_3$ and $\vt^M_k = \vg_k$,
where $\vg_k$ is the original world frame position of vertex $k$.
Intuitively, these transformations describe the local deformations on the mesh:
$\MR^M_k$ is the local rotation centered at vertex $k$ while $\vt^M_k - \vg_k$ is the local translation.
Mesh vertices are connected to each other using the edges of the simplified mesh (the green edges in \Cref{fig:deformation-graph}).

We then add the nodes of the robot pose graph to the deformation graph as pose vertices
with associated transformation $\MX_i = \begin{bmatrix} \MR^X_i & \vt^X_i \\ \mathbf{0}^\top & 1 \end{bmatrix}$.
When the mesh is not yet deformed, $\MX_i$ is just the odometric pose for node $i$.
The pose vertices are connected according to the original connectivity of the pose graph (the red edges in \Cref{fig:deformation-graph}).
A pose vertex $i$ is connected to the mesh vertex $k$ if the mesh vertex $k$ is visible by the camera associated to pose $i$.

\Cref{fig:deformation-graph} showcases the components and creation of the deformation graph,
noting the pose and mesh vertices and the edges, and the deformation that happens when a loop closure is detected.
Based on the loop closure and odometry measurements $\MZ_{ij}$ and given $n$ pose vertices and $m$ mesh vertices, 
the deformation graph optimization is as follows:

\begin{align} \label{eq:pgmo}
\argmin_{\substack{\MX_1,\ldots,\MX_n \in \SEthree \\
    \MM_1, \ldots,\MM_m \in \SEthree}} &\sum_{\MZ_{ij}}||\MX_i^{-1}\MX_j - \MZ_{ij}||^2_{\M\Omega_{ij}} \notag\\
 + &\sum_{k=0}^m\sum_{l \in \mathcal{N}^M(k)}||\MR^M_k(\vg_l - \vg_k) + \vt^M_k - \vt^M_l||^2_{\M\Omega_{kl}} \notag\\
 + &\sum_{i=0}^n\sum_{l \in \mathcal{N}^M(i)}||\MR^X_i\tilde{\vg}_{il} + \vt^X_i - \vt^M_l||^2_{\M\Omega_{il}}
\end{align}

\noindent where $\mathcal{N}^M(i)$ indicates the neighboring mesh vertices to a vertex $i$ in the deformation graph
and $\vg_i$ denotes the non-deformed (initial) \rebuttal{world frame} position of mesh or pose vertex $i$ in the deformation graph,
and $\widetilde{\vg}_{il}$ denotes the non-deformed position of vertex $l$ in the coordinate frame of the odometric pose of node $i$
\rebuttal{(note that $\widetilde{\vg}_{il} \neq \vg_l - \vg_i$ except when the undeformed orientation of node $i$ is identity.)}.

The first term in the optimization enforces the odometric and loop closure measurements on the poses in the pose graph, 
these are the same as in standard pose graph optimization~\citep{Cadena16tro-SLAMsurvey};
the second term is adapted from~\citep{Summer07siggraph-embeddedDeformation} and enforces local rigidity between mesh vertices
by minimizing the change in relative translation between connected mesh vertices (\ie preserving the edge connecting two mesh vertices);
the third term enforces the local rigidity between a pose vertex $i$ and a mesh vertex $l$,
again by minimizing the change in relative translation between the two vertices.
Note that $||\cdot||_{\M\Omega}$ indicates the weighted Frobenius norm
\begin{equation}
  ||\MA||^2_\MOmega = \trace{\MA \MOmega \MA^\top}
\end{equation}
where $\MOmega$ takes the form $\begin{bmatrix} \omega_R\MI_{3} & \mathbf{0} \\ \mathbf{0}^\top & \omega_t \end{bmatrix}$ 
with $\omega_R$ and $\omega_t$ respectively corresponding to the rotation and translation weights, 
as defined in~\citep{Briales17ral}.

In the following , we show that ~\cref{eq:pgmo} can be formulated as an augmented pose graph optimization problem.
Towards this goal, we define $\tilde{\MR_i}$ as the initial odometric rotation of pose vertex $i$ in the deformation graph; 
we then define,

\begin{equation}
  \MG_{ij} = \begin{bmatrix} \MI_{3} & \vg_j - \vg_i \\ 0 & 1 \end{bmatrix}
\end{equation}
\begin{equation}
  \bar{\MG}_{ij} = \begin{bmatrix} \rebuttal{\MI_{3}} & \tilde{\MR}^{-1}_i(\vg_j - \vg_i) \\ 0 & 1 \end{bmatrix}
\end{equation}
and rewrite the optimization as,
\begin{align}
\argmin_{\substack{\MX_1,\ldots,\MX_n \in \SEthree \\
    \MM_1, \ldots,\MM_m \in \SEthree}} &\sum_{\MZ_{ij} \in \calZ}||\MX_i^{-1}\MX_j - \MZ_{ij}||^2_{\M\Omega_{\MZ_{ij}}} + \notag\\
  &\sum_{\MG_{ij} \in \calG}||\MM_i^{-1}\MM_j - \MG_{ij}||^2_{\M\Omega_{\MG_{ij}}} + \notag\\
  &\sum_{\bar{\MG}_{ij} \in \bar{\calG}}||\MX_i^{-1}\MM_j - \bar{\MG}_{ij}||^2_{\M\Omega_{\bar{\MG}_{ij}}}
\end{align}
where $\calZ$ is the set of all odometry and loop closures edges in the deformation graph
(the red and blue edges in \Cref{fig:deformation-graph}),
$\calG$ is the set edges from the simplified mesh
(the green edges in \Cref{fig:deformation-graph}),
and $\bar{\calG}$ is the set of all the edges connecting a pose vertex to a mesh vertex
(the yellow edges in \Cref{fig:deformation-graph}).
We only optimize over translation in the second and third term hence 
the rotation weights $\omega_R$ for $\M\Omega_{\MG_{ij}}$ and $\M\Omega_{\bar{\MG}_{ij}}$ are set to zero.
Taking it one step further and observing that the terms
are all based on the edges in the deformation graph,
we can define $\MT_i$ as the transformation of pose or mesh vertex $i$
and $\ME_{ij}$ is the transformation that corresponds to an edge in the deformation graph
that is of the form $\MZ_{ij}$, $\MG_{ij}$, $\bar{\MG}_{ij}$ depending on the type of edge.
With this reparametrization, we are left with a pose graph optimization problem
akin to the ones found in the literature~\citep{Rosen18ijrr-sesync,Cadena16tro-SLAMsurvey}.

\begin{equation}
  \argmin_{\substack{\MT_1,\ldots,\MT_{n + m} \in \SEthree}} \sum_{\ME_{ij}} ||\MT_i^{-1}\MT_j - \ME_{ij}||^2_{\M\Omega_{ij}}
\end{equation}

\rebuttal{Note that the deformation graph approach, as originally presented in~\citep{Summer07siggraph-embeddedDeformation},
is equivalent to pose graph optimization only when rotations are used in place of affine transformations.}
The pose graph is then optimized using GTSAM.

After the optimization, the positions of the vertices of the complete mesh are updated
as affine transformations of the nodes in the deformation graph:
\begin{equation}
  \widetilde{\vv}_i = \sum_{j=1}^m w_j(\vv_i)[\MR^M_j(\vv_i - \vg_j) + \vt^M_j]
\end{equation}
where $\vv_i$ indicates the original vertex positions and $\widetilde{\vv}_i$ are the new deformed positions.
The weights $w_j$ are defined as
\begin{equation}
    w_j(\vv_i) = \left(1 - ||\vv_i - \vg_j||/{d_{\text{max}}}\right)^2
\end{equation}
and then normalized to sum to one. Here $d_{\text{max}}$ is the distance to the $k + 1$ nearest node
as described in~\citep{Summer07siggraph-embeddedDeformation} (we set $k=4$).

\begin{remark}[\KimeraRPGO and \KimeraPGMO]
\label{rmk:rpgo-pgmo}

\KimeraRPGO is the robust pose graph optimizer we introduced in~\citep{Rosinol20icra-Kimera} which uses
a modified Pairwise Consistent Measurement Set Maximization (PCM)~\citep{Mangelson18icra} approach
to filter out incorrect loop closures caused by perceptual aliasing
then optimizes the poses of the robot.
\KimeraPGMO is a strict generalization of \KimeraRPGO.
Both \KimeraRPGO and \KimeraPGMO perform loop closure detection and outlier rejection,
the difference is that \KimeraPGMO additionally optimizes the mesh with extra computational cost to solve a larger pose graph.
As we will see in the experimental section, \KimeraPGMO takes almost three times the time of \KimeraRPGO
since it optimizes a larger graph. For instance, \KimeraPGMO optimizes 728 pose nodes and 1031 mesh nodes for the \Euroc V1\_01 dataset,
while \KimeraRPGO only optimizes over the 728 pose nodes.
\end{remark}

\subsection{\KimeraHumans: Human Shape Estimation and Robust Tracking}
\label{ssec:humans}

\myParagraph{Robot Node}
In our setup, the only robotic agent is the one collecting the data.
Hence, \KimeraPGMO directly produces a time-stamped
pose graph describing the poses of the robot at discrete time steps.
 To complete the robot node, we assume a CAD model of the robot to be given
(only used for visualization).

\myParagraph{Human Nodes}
Contrary to related work that models dynamic targets as a point or a 3D
pose~\citep{Chojnacki18ijmav-motionAndObjectTracking,Azim12ivs-datmo,Aldoma13icra-objectTracking,Peiliang18eccv-motionAndObjectTracking,Qiu19tro-vioObjectTracking},
\KimeraHumans tracks a dense time-varying mesh model describing the shape of the human over time.
Therefore, to create a human node \KimeraHumans needs to detect and estimate the shape of a human in the camera images,
and then track the human over time.

Besides using them for tracking,
we feed the human detections back to \KimeraSemantics, such that dynamic elements are not
reconstructed in the 3D mesh.
We achieve this by only using the free-space information
when ray casting the depth for pixels labeled as humans, an approach we dubbed \emph{dynamic masking}
(see results in \Cref{fig:dynamicMesh}).

\renewcommand{\myhspace}{\hspace{-6mm}}
\renewcommand{\mpw}{3.2cm}
\renewcommand{\myRate}{3.3cm}

\begin{figure}[h]
\vspace{-2mm}
	\begin{center}
	\begin{minipage}{\textwidth}
	\begin{tabular}{ccc}%
	\myhspace \hspace{3mm}
			\begin{minipage}{\mpw}%
			\centering%
			\includegraphics[height=\myRate]{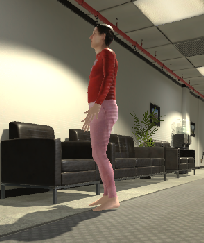} \\
			(a) Image
			\end{minipage}
		& \myhspace 
			\begin{minipage}{\mpw}%
			\centering%
			\includegraphics[trim=8mm 0mm 8mm 0mm, clip, height=\myRate]{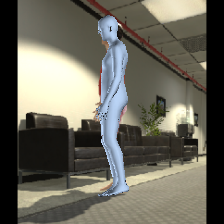} \\
			(b) Detection
			\end{minipage}
		& \myhspace 
			\begin{minipage}{\mpw}%
			\centering%
			\includegraphics[trim=120mm 40mm 20mm 110mm, clip,height=\myRate]{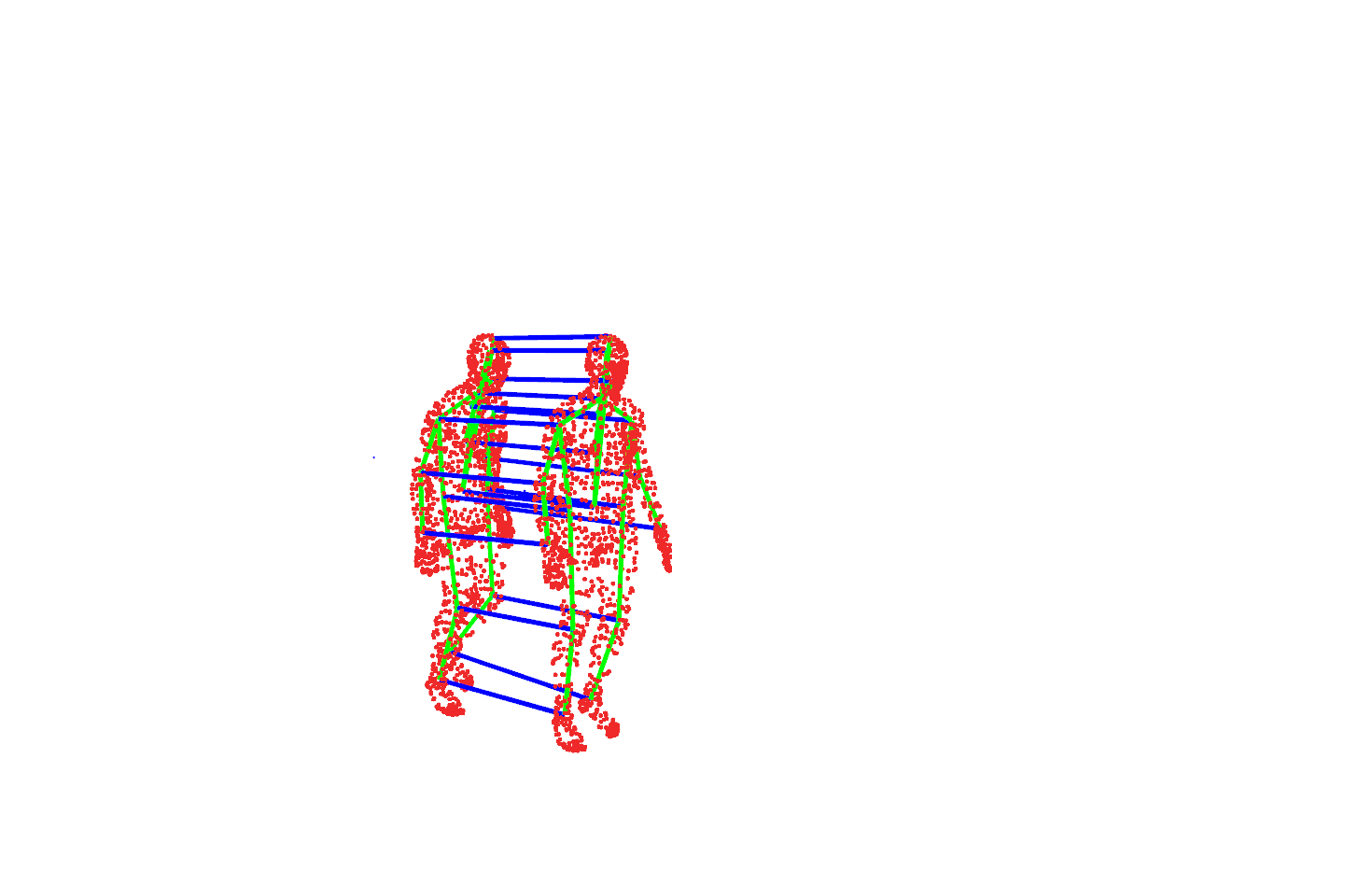} \\
			(c) Tracking
			\end{minipage}
		\end{tabular}
	\end{minipage}
	\begin{minipage}{\textwidth}
	\end{minipage}
	\vspace{-1mm} 
	\caption{{\bf Human nodes}: (a) Input camera image from Unity, (b) SMPL mesh detection and pose/shape estimation using~\citep{Kolotouros19cvpr-shapeRec}, (c) Temporal tracking and consistency checking on the maximum joint displacement between detections.
	 \label{fig:humans}}
	\vspace{-8mm} 
	\end{center}
\end{figure}

For human shape and pose estimation, we use the Graph-CNN approach of \cite{Kolotouros19cvpr-shapeRec}~(\MeshName),
which directly regresses the 3D location of the
vertices of an SMPL~\citep{Loper15tg-smpl} mesh model from a single image.
An example mesh is shown in \Cref{fig:humans}(a-b).

Given a pixel-wise 2D segmentation of the image, we crop the left camera image to
a bounding box around each detected human, which then becomes an input to \MeshName. \MeshName outputs a 3D SMPL
mesh for the corresponding human,
as well as camera parameters ($x$ and $y$ image position and a scale factor corresponding to a weak perspective camera model).
We then use the camera model to project the human
mesh vertices into the image frame. After obtaing the projection, we then compute the location and orientation
of the full-mesh with respect to the camera using PnP~\citep{Zheng2013ICCV-revisitPnP} to optimize the camera
pose based on the reprojection error of the
mesh into the camera frame. The translation is recovered from the depth-image, which is used to get
the approximate 3d position of the pelvis joint of the human in the image.
Finally, we transform the mesh location to the global frame based on the world transformation output by the \KimeraVIO.

\myParagraph{Human Tracking and Monitoring}
The above approach relies heavily on the accuracy of \MeshName and
discards useful temporal information about the human. In fact, \MeshName outputs are unreliable in
several scenarios, especially when the human is partially occluded. In this section, we describe our
method for (i) maintaining persistent information about human trajectories,
(ii) monitoring \MeshName location and pose estimates to determine which estimates are inaccurate, and (iii) mitigating
human location errors through pose-graph optimization using motion priors.
We achieve these results by maintaining a pose graph for each human the robot encounters and updating
the pose graphs using simple but robust data association.

\myParagraph{Pose Graph}
To maintain persistent information about human location, we build a pose graph for each human where each node in the graph
corresponds to the location of the pelvis of the human at a discrete time. Consecutive poses are connected by a
factor (\cite{Dellaert17fnt-factorGraph}) modeling a zero velocity prior on the human motion with a permissive noise model to allow
for small motions.
The location information from \MeshName is modelled as a prior factor, providing the estimated global
coordinates at each timestep. In addition to the pelvis locations, we maintain a persistent history of the SMPL parameters of
the human as well as joint locations for pose analysis.

\begin{figure}[htbp]
	\begin{center}
    \begin{minipage}{\columnwidth}
      \includegraphics[trim=4.0cm 1.0cm 25.0cm 13.0cm, clip, width=\columnwidth]{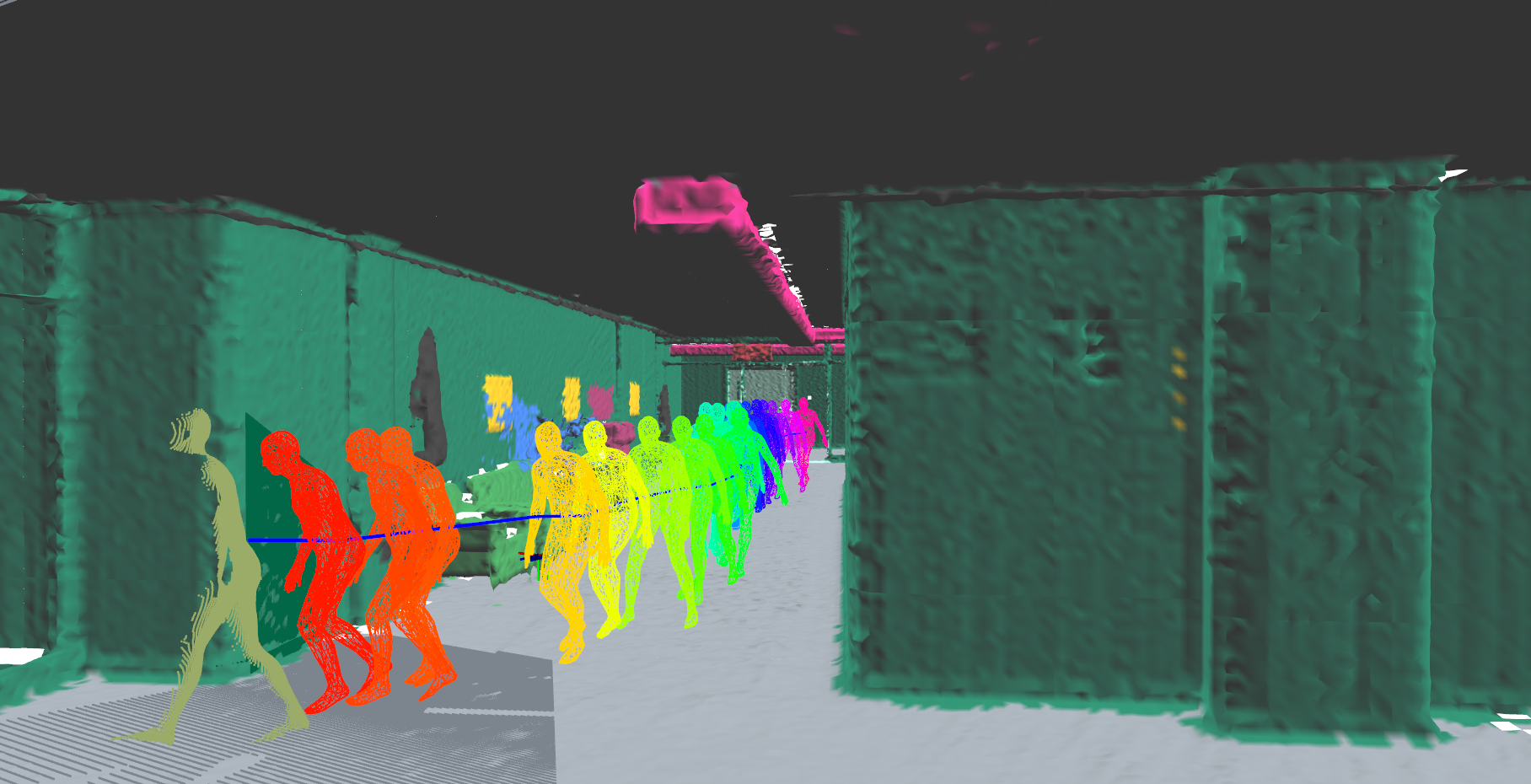} \\
    \end{minipage}
    \vspace{-1em}
    \caption{{\bf Human Pose-Graph}: Optimized pose-graph (blue line) for a single human.
     The detected human shape is shown as a 3D mesh, color-coded from the most recent detection in red to the oldest one in pink.
    \label{fig:humans_posegraphs}}
	\end{center}
\end{figure}

The advantage of the pose-graph system is two-fold. First, using a pose-graph for each human's trajectory allows for the application
of pose-graph-optimization techniques to get a trajectory estimate that is smooth and robust to misdetections.
Many of the detections from \MeshName propagate to the pose-graph even if they are not immediately rejected by
the consistency checks described in the next section. However, by using Kimera-RPGO and PCM outlier rejection,
the pose-graphs of the humans can be regularly optimized to smooth the trajectory and remove bad detections.
PCM outlier rejection is particularly good at removing detections that would require the human to move/rotate arbitrarily fast.
Second, using pose graphs to model both the humans and the robot's global trajectory allows for unified visualization tools
between the two use-cases. \Cref{fig:humans_posegraphs} shows the pose graph (blue line) of a human in the office environment,
as well as the detection associated with each pose in the graph (rainbow-like color-coded human mesh).

\renewcommand{\myhspace}{\hspace{-6mm}}
\renewcommand{\mpw}{3.2cm}
\renewcommand{\myRate}{3.3cm}

\myParagraph{Data Association}
A key issue in the process of building the pose graph is associating which nodes belong to the same human over time and
then linking them appropriately. We use a simplified data association model which associates a new node with the node that
has the closest euclidean distance to it. This form of data association works well under the mild assumption that the distance
a human moves between timesteps is smaller than the distance between humans.

We do not have information for when a human enters the frame and when they leave (although we do know the number of people in a given frame).
To avoid associating new humans with the pose graphs of previous humans, we add a spatio-temporal consistency check before adding the
pose to the human's pose-graph, as discussed below.

To check consistency, we extract the human skeleton at time $t-1$ (from the pose graph) and $t$ (from the current detection) and check that the
motion of each joint (\Cref{fig:humans}(c)) is physically plausible in that time interval (\ie we leverage the fact that
the joint and torso motion cannot be arbitrarily fast). This check is visualized in \Cref{fig:humans}(c). We first ensure that the
rate of centroid movement is plausible between the two sets of skeletons.
 Median human walking speed being about $1.25$ m/s~\citep{Schimpl2011plos-walkingSpeed}, we
use a conservative $3$m/s bound on the movement rate to threshold the feasibility for data association. In addition, we use a conservative
bound of \maxJointDist on the maximum allowable joint displacement to bound irregular joint movements.

The data association check is made more robust by using the beta-parameters of the SMPL model~\citep{Loper15tg-smpl}, which encode the various
shape attributes of the mesh in 8 floating-point parameters. These shape parameters include, for example, the width and height of different
features of the human model. We check the current detection's beta parameters against those of the skeleton at time~$t-1$ and ensure that the
average of the difference between each pair of beta parameters does not exceed a certain threshold ($0.1$ in our experiments). This helps to differentiate humans from each
other based on their appearance.
In \KimeraHumans, the beta parameters are estimated by \MeshName~\citep{Kolotouros19cvpr-shapeRec}. 

If the centroid movement and joint movement between the timesteps are within the bounds and the beta-parameter check passes, we
add the new node to the pose graph that has the closest final node as described earlier. If no pose graph meets the
consistency criteria, we initialize a new pose graph with a single node corresponding to the current detection.

\myParagraph{Node Error Monitoring and Mitigation}
As mentioned earlier, \MeshName outputs are very sensitive to occluded humans, and prediction quality is poor in those circumstances.
To gain robustness to simple occlusions, we mark detections when the bounding box of the human approaches the boundary of the image
or is too small ($\leq 30$ pixels in our tests) as incorrect. In addition, we use the size of the pose graph as a proxy to monitor the error of the nodes.
When a pose graph has few nodes, it is highly likely that those nodes are erroneous. We determined through experimental results that
pose graphs with fewer than $10$ nodes tend to have extreme errors in human location. We mark those graphs as erroneous and remove them from the
\DSG, a process we refer to as pose-graph pruning.
 This is similar to removing short feature tracks in visual tracking.
  Finally, we mitigate node errors by running optimization over the pose graphs using the
stationary motion priors and we see that we can achieve a great reduction in existing errors.

\subsection{\KimeraObjects: Object Pose Estimation}
\label{sec:objects}

Within \KimeraDSG, \KimeraObjects is the module that extracts static objects from the optimized metric-semantic mesh 
produced by \KimeraPGMO.
 We give the user the flexibility to provide a catalog of CAD models 
for some of the object classes. If a shape is available, 
\KimeraObjects will try to fit it to the mesh (paragraph ``Objects with Known Shape'' below),
otherwise it will only attempt to estimate 
a centroid and a bounding box (paragraph ``Objects with Unknown Shape'').

\myParagraph{Objects with Unknown Shape}
The optimized metric-semantic mesh from \KimeraPGMO already contains semantic labels.  
Therefore, \KimeraObjects first extracts the portion of the mesh belonging to a given object class
(\eg chairs in~\FigFrontCover(d)); this mesh contains multiple objects belonging to the same class.
To break down the mesh into multiple object instances, \KimeraObjects performs Euclidean clustering using PCL~\citep{Rasu11icra} (with a distance threshold of {twice the voxel size, $0.05m$, used in \KimeraSemantics, that 
is $0.1$ m)}.
From the segmented clusters, \KimeraObjects obtains a centroid of the object (from the vertices of the corresponding mesh), 
and assigns a canonical orientation with axes aligned with the world frame.
Finally, it computes a bounding box with axes aligned with the canonical orientation. 

\myParagraph{Objects with Known Shape}
if a CAD model for a class of objects is given, 
\KimeraObjects will attempt to fit the known shape to the object mesh.
This is done in three steps. First, 
\KimeraObjects extracts 3D keypoints from the CAD model of the object, and the corresponding object mesh.
The 3D keypoints are extracted by transforming each mesh to a point cloud (by picking the vertices of the mesh) and then 
extracting 3D Harris keypoints~\citep{Rasu11icra} with
$0.15$m radius and $10^{-4}$ non-maximum suppression threshold.
Second, we match every keypoint on the CAD model with any keypoint on the \Kimera model.
 Clearly, this step produces many incorrect putative matches (outliers).
  Third, we apply a robust open-source registration technique, \TEASERpp~\citep{Yang20tro-teaser},
   to find the best alignment between the point clouds in the presence of extreme outliers. 
The output of these three steps is a 3D pose of the object (from which it is also easy to extract an axis-aligned bounding box), 
see result in~\FigFrontCover(e).

\subsection{\KimeraBuildingParser: Extracting Places, Rooms, and Structures}
\label{sec:buildingParser}

\begin{figure*}[htbp]
  \begin{center}
  \includegraphics[trim=0mm 1mm 0mm 0mm, clip, width=0.8\columnwidth]{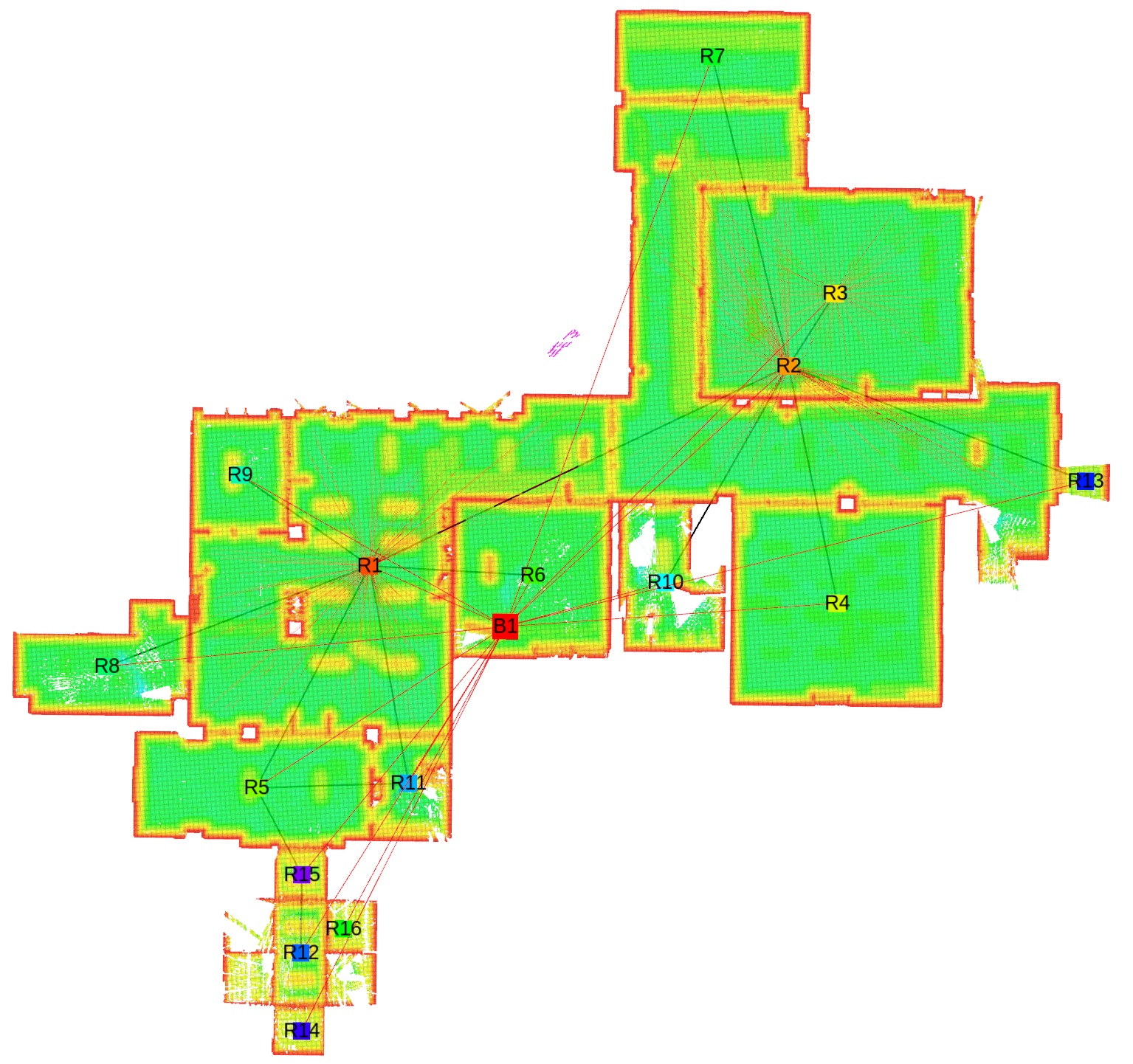}
  \includegraphics[trim=0mm 1mm 0mm 0mm, clip, width=0.8\columnwidth]{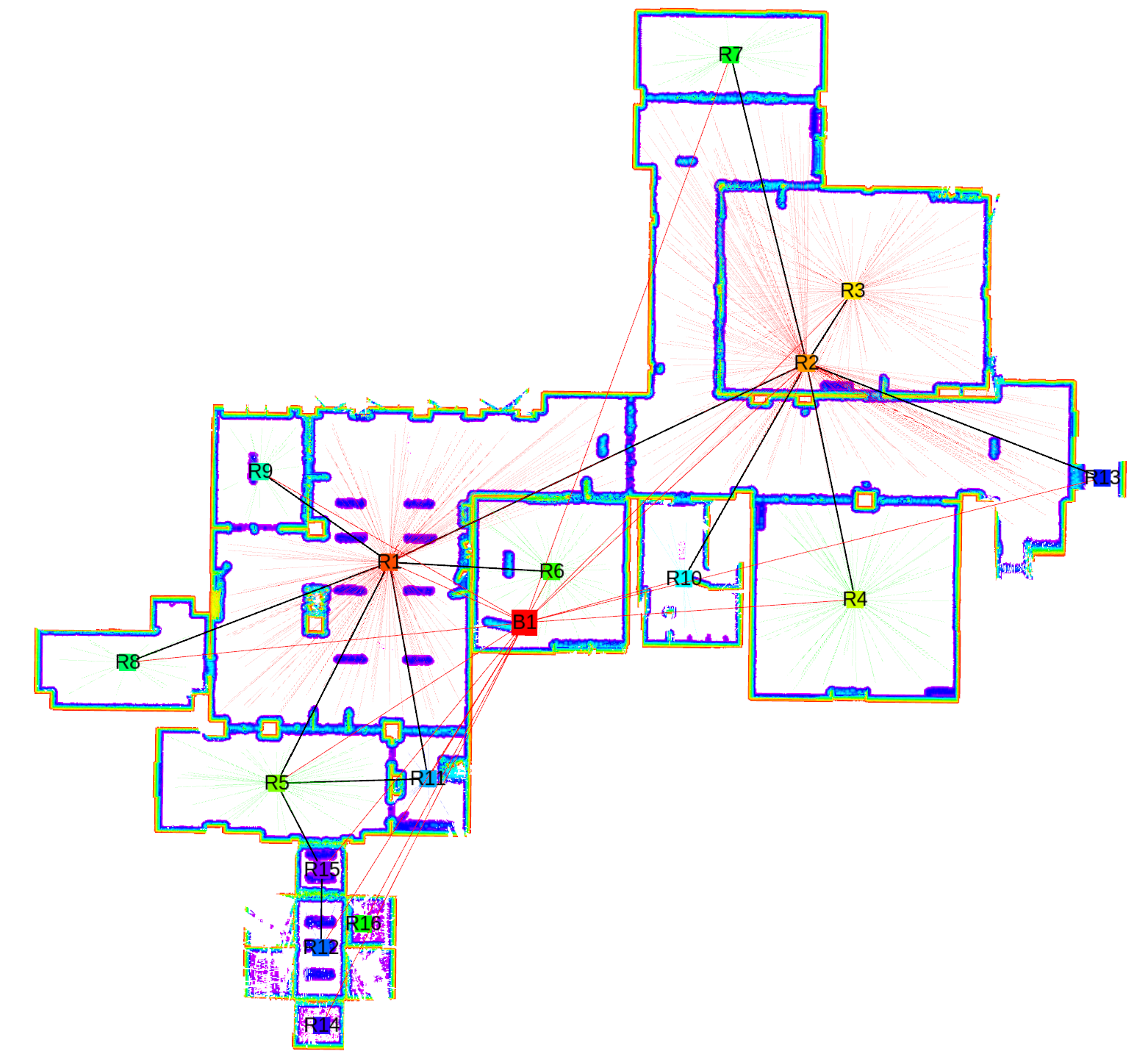}
  \caption{(Left) 2D slice of 3D ESDF. The euclidean distance is color coded from red ($0$m) to green ($0.5$m).
  (Right) Truncated ($\leq 0.10$m) 2D ESDF, revealing the room's contours.
  Overlaid in both figures are the estimated room layout (square nodes) and their connectivity (black edges).}
  \label{fig:truncated_esdf}
  \end{center}
\end{figure*}

\KimeraBuildingParser implements simple-yet-effective methods to 
parse places, structures, and rooms from \Kimera's 3D mesh.

\myParagraph{Places}
\KimeraSemantics uses Voxblox~\citep{Oleynikova2017iros-voxblox} to extract a global mesh and an ESDF.
We also obtain a topological graph from the ESDF using~\citep{Oleynikova18iros-topoMap}, where nodes %
sparsely sample  the free space, while edges represent straight-line traversability between two nodes.
We directly use this graph to extract the places and their topology (\Cref{fig:rooms}(a)).  
After creating the places, we associate each object and agent pose
to the nearest place to model a proximity relation.

\myParagraph{Structures}
\Kimera's semantic mesh already includes different labels for walls, floor, and ceiling. Hence, isolating these three 
structural elements is straightforward (\Cref{fig:structure}). For each type of structure, we then compute a centroid, assign a canonical orientation (aligned with the world frame), and compute an axis-aligned bounding box.
We further segment the walls depending on the room they belong to.
To do so, we leverage the property that the 3D mesh vertices normals are oriented: the normal of each vertex points towards the camera.
For each 3D vertex of the walls' mesh, we query the nearest nodes in the places layer that are in the normal direction, and limit the search to a conservative radius ($0.5$m).
We then use the room IDs of the retrieved places to vote for the room label of the current wall mesh vertex.
To make the approach more robust to cases such as when two rooms meet (frames of doors for example), we weight the votes of the places by $\frac{\bf{n} \cdot \bf{d}}{\|\bf{d}\|^2_2}$  where $\bf{n}$ is the normal (unit norm) at the wall vertex and $\bf{d}$ is the vector from the wall vertex to the place node.
This downweights votes of places that are not immediately in front and near the wall vertex.

\myParagraph{Rooms}
While floor plan computation is challenging in general, 
(i) the availability of a 3D \ESDF and 
(ii) the knowledge of the gravity direction given by \KimeraVIO
enable a simple-yet-effective approach to partition the environment into different rooms.
The key insight is that an horizontal 2D section of the 3D ESDF, cut below the level of the detected ceiling, is relatively unaffected by clutter in the room (\Cref{fig:truncated_esdf}).
 This 2D section gives a clear signature of the room layout: the voxels in the section have a value of $0.3$m almost everywhere (corresponding to the distance to the ceiling), except close to the walls, where the distance decreases to $0$m.
  We refer to this 2D \ESDF (cut at $0.3$m below the ceiling) as an \emph{\ESDF section}.

 To compensate for noise, we further truncate the \ESDF section to distances above $0.2$m, such that small openings between rooms (possibly resulting from error accumulation) are removed. 
 The result of this partitioning operation is a set of disconnected 2D ESDFs corresponding to each room, that we refer to as \emph{2D \ESDF rooms}. 
Then, we label all the ``Places'' (nodes in Layer 3) that fall inside a 2D \ESDF room depending on their 2D (horizontal) position.
 At this point, some places might not be labeled (those close to walls or inside door openings).
 To label these, we use majority voting over the neighborhood of each node in the topological graph of ``Places'' in Layer 3; we repeat majority voting until all places have a label. 
 Finally, we add an edge between each place (Layer 3) and its corresponding room (Layer 4), see~\Cref{fig:rooms}(b-c),
  and add an edge between two rooms (Layer 4) if there is an edge connecting two of its places (red edges in~\Cref{fig:rooms}(b-c)). 
 We also refer the reader to the second video attachment.

\subsection{Debugging Tools}
\label{sec:debugging}
 
\Kimera also provides an open-source\footnote{\url{https://github.com/MIT-SPARK/Kimera-Evaluation}} suite of evaluation tools  for debugging, visualization, and benchmarking of VIO, SLAM, and
metric-semantic reconstruction.
\Kimera includes a Continuous Integration server (Jenkins) that asserts the quality of the code (compilation, unit tests),
but also automatically evaluates \KimeraVIO, \KimeraRPGO, and \KimeraPGMO on the \Euroc's datasets using \emph{evo}~\citep{Grupp17evo}. 
Moreover, we provide Jupyter Notebooks to visualize intermediate VIO statistics (\eg quality of the feature tracks, IMU preintegration errors),
as well as to automatically assess the quality of the 3D reconstruction using Open3D~\citep{Zhou18arxiv-open3D}.


\section{Experimental Evaluation}
\label{sec:experiments}

We start by introducing the datasets that we use for evaluation in \Cref{ssec:datasets},
which feature real and simulated scenes, with and without dynamic agents,
as well as a large variety of environments (indoors and outdoors, small and large).
\Cref{sec:experiments-VIO} shows that \KimeraVIO and \KimeraRPGO attain \rebuttal{competitive} pose estimation performance on the \Euroc dataset.
\Cref{ssec:geometric_performance} demonstrates \Kimera's 3D mesh geometric accuracy on \Euroc, using the subset of scenes providing a ground-truth point cloud.
\Cref{ssec:semantic_performance} provides a detailed evaluation of \KimeraMesher and \KimeraSemantics' 3D metric-semantic reconstruction on the \UnityHumans dataset.
\Cref{ssec:experiments-MeshLoopClosure} evaluates the localization and reconstruction performance of \KimeraPGMO.
\Cref{ssec:exp-humanAndObjects} evaluates the humans and object localization errors.
\Cref{ssec:exp-roomParsing} evaluates the accuracy of the segmentation of places into rooms.
\Cref{ssec:timing_performance} highlights \Kimera's real-time performance and analyzes the runtime of each module.
Furthermore, \Cref{ssec:tx2_timing_performance} shows how \Kimera's real-time performance scales when running on embedded computers.
Finally, \Cref{ssec:real_life_experiments} qualitatively shows the performance of \Kimera on real-life datasets that we collected.

\subsection{Datasets}
\label{ssec:datasets}

\begin{figure*}[htbp]
    \centering
    \includegraphics[trim={0cm 0cm 0cm 0cm}, clip, width=\textwidth]{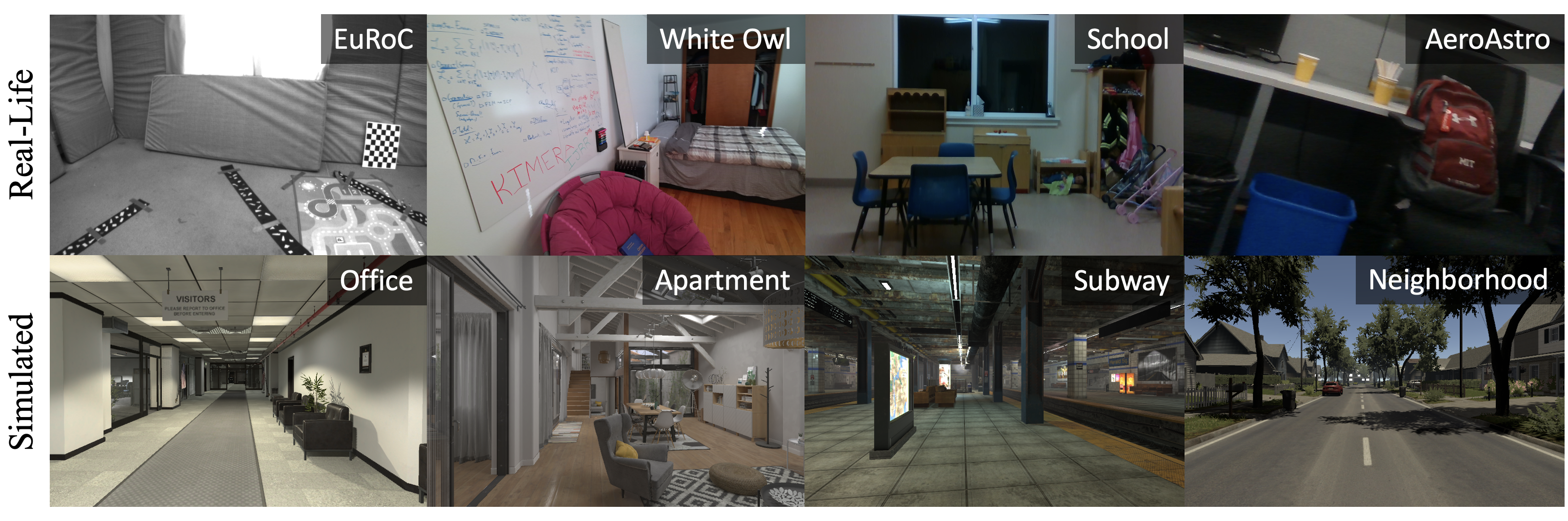}
    \caption{Overview of the datasets used for evaluation.
     We evaluate \Kimera on a variety of datasets, both real-life (top row) and simulated (bottom row), indoors and outdoors, small and large.}
    \label{fig:datasets_overview}
\end{figure*}

\Cref{fig:datasets_overview} gives an overview of the datasets used, while their characteristics are detailed below.

\subsubsection{\Euroc.}

We use the \Euroc dataset~\citep{Burri16ijrr-eurocDataset} that features a small drone flying indoors with a mounted stereo camera and an IMU.
The \Euroc dataset includes eleven datasets in total, recorded in two different \textit{static} scenarios.
The \textit{Machine Hall} scenario ({MH}) is the interior of an industrial facility.
The \textit{Vicon Room} ({V}) is similar to an office room.
Each dataset has different levels of difficulty for VIO, where the speed of the drone increases
(the larger the numeric index of the dataset the more difficult it is; e.g. MH\_01 is easier for VIO than MH\_03).
The dataset features ground-truth localization of the drone in all datasets.
Furthermore, a ground-truth pointcloud of the \textit{Vicon Room} is available.

For our experimental evaluation, we use \Euroc to analyze both the localization performance of \KimeraVIO, \KimeraRPGO, and \KimeraPGMO,
as well as the geometric reconstruction from \KimeraMesher, \KimeraSemantics, and \KimeraPGMO.
Since there are no dynamic elements in this dataset, nor semantically meaningful objects,
 we do not use it to evaluate the semantic accuracy of the mesh or the robustness to dynamic scenes.

\subsubsection{\UnityHumans and \UnityHumansTwo.}

To evaluate the accuracy of the metric-semantic reconstruction and the robustness against dynamic elements,
we use the \UnityHumans simulated dataset that we introduced in \citep{Rosinol20rss-dynamicSceneGraphs},
and further release a new \UnityHumansTwo dataset with this paper.

\UnityHumans and \UnityHumansTwo are generated using a photo-realistic Unity-based simulator provided by MIT Lincoln Laboratory,
 that provides sensor streams (in ROS) and ground truth for both the
geometry and the semantics of the scene, and has an interface similar to~\citep{SayreMcCord18icra-droneSystem,Guerra19arxiv-flightGoggles}.

\UnityHumans features a large-scale office space with multiple rooms, as well as a small (e.g.~12) and large (e.g.~60) number of humans in it,
 and it is the one reconstructed in \Cref{fig:DSG}.
Despite having different number of humans, \UnityHumans had the issue that the trajectories for each run were not the same;
 thereby coupling localization errors due to dynamic humans in the scene and the intrinsic drift of the VIO.
For this reason, and to extend the dataset for a variety of other scenes, we collected the \UnityHumansTwo dataset.
\UnityHumansTwo features indoor scenes, such as the `Apartment', the `Subway',
 and the `Office' scene, as well as the `Neighborhood' outdoor scene.
Note that the `Office' scene in \UnityHumansTwo is the same as in \UnityHumans, but the trajectories are different.
Finally, to avoid biasing the results towards a particular 2D semantic segmentation method,
 we use ground truth 2D semantic segmentations and we refer the reader to~\cite{Hu19ral-fuses} for a review of potential alternatives.

\subsubsection{\BldgThirtyOne, \School, and \WhiteOwl.}
\label{ssec:real_dataset}

Since the \UnityHumans and \UnityHumansTwo datasets are simulated,
 we further evaluate our approach on three real-life datasets that we collected.
The three datasets consist of RGB-D and IMU data recorded
using a hand-held device. The first dataset features
a collection of student cubicles in one of the MIT academic buildings
(\BldgThirtyOne), and is recorded using a custom-made sensing rig.
The second is recorded in a \School, using the same custom-made sensing rig.
The third dataset is recorded in an apartment (\WhiteOwl), using Microsoft's \AzureKinect.

\begin{figure}[htbp]
    \centering
    \begin{tabular}{cc}
    \begin{minipage}{0.44\columnwidth}
        \centering
        \includegraphics[trim={2cm 30cm 00cm 15cm}, clip, height=\columnwidth, angle=0]{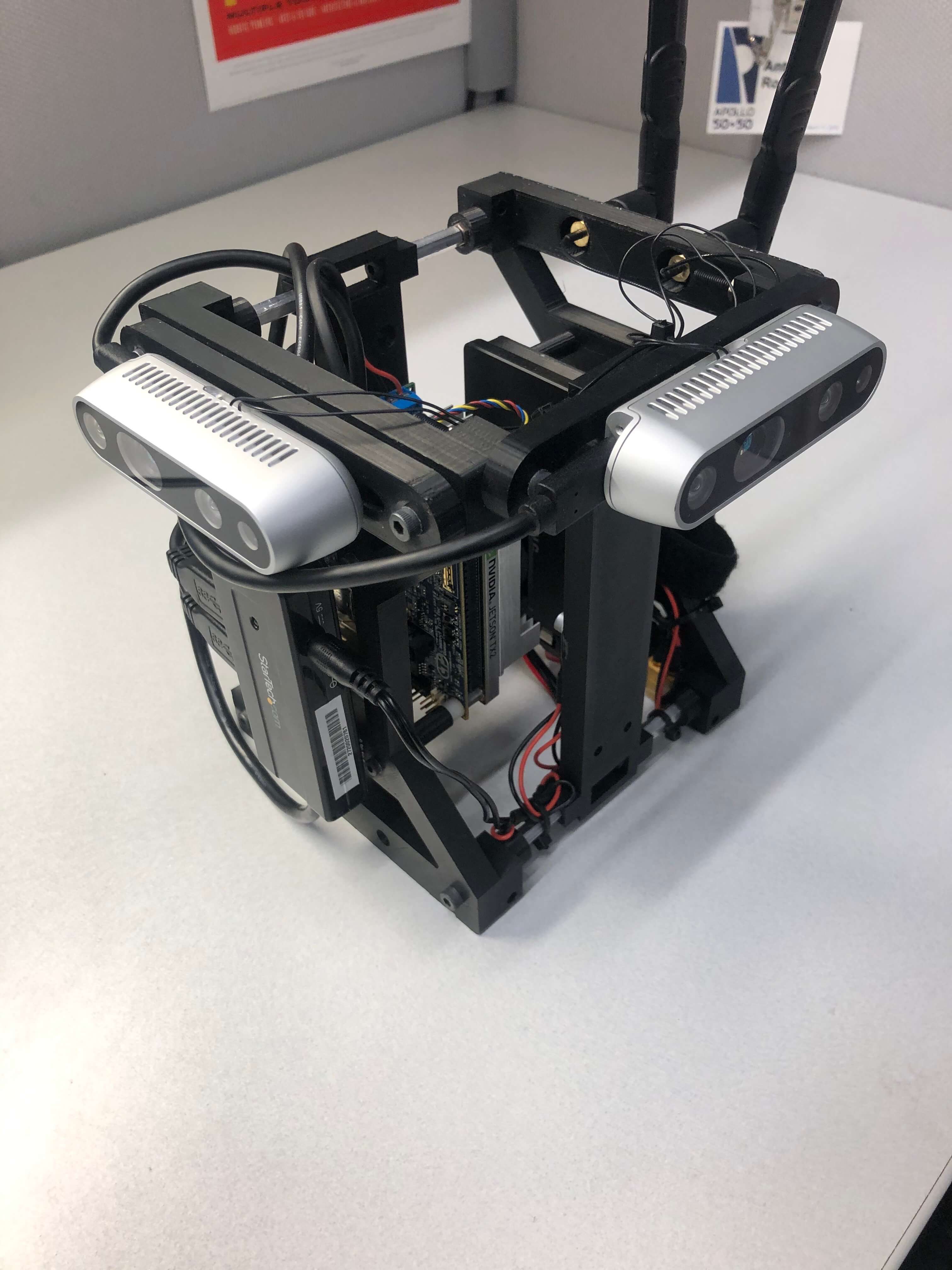} \\
        (a) perspective view
    \end{minipage} &
    \begin{minipage}{0.44\columnwidth}
        \centering
        \includegraphics[trim={4cm 28cm 30cm 42cm}, clip, width=\columnwidth, angle=90]{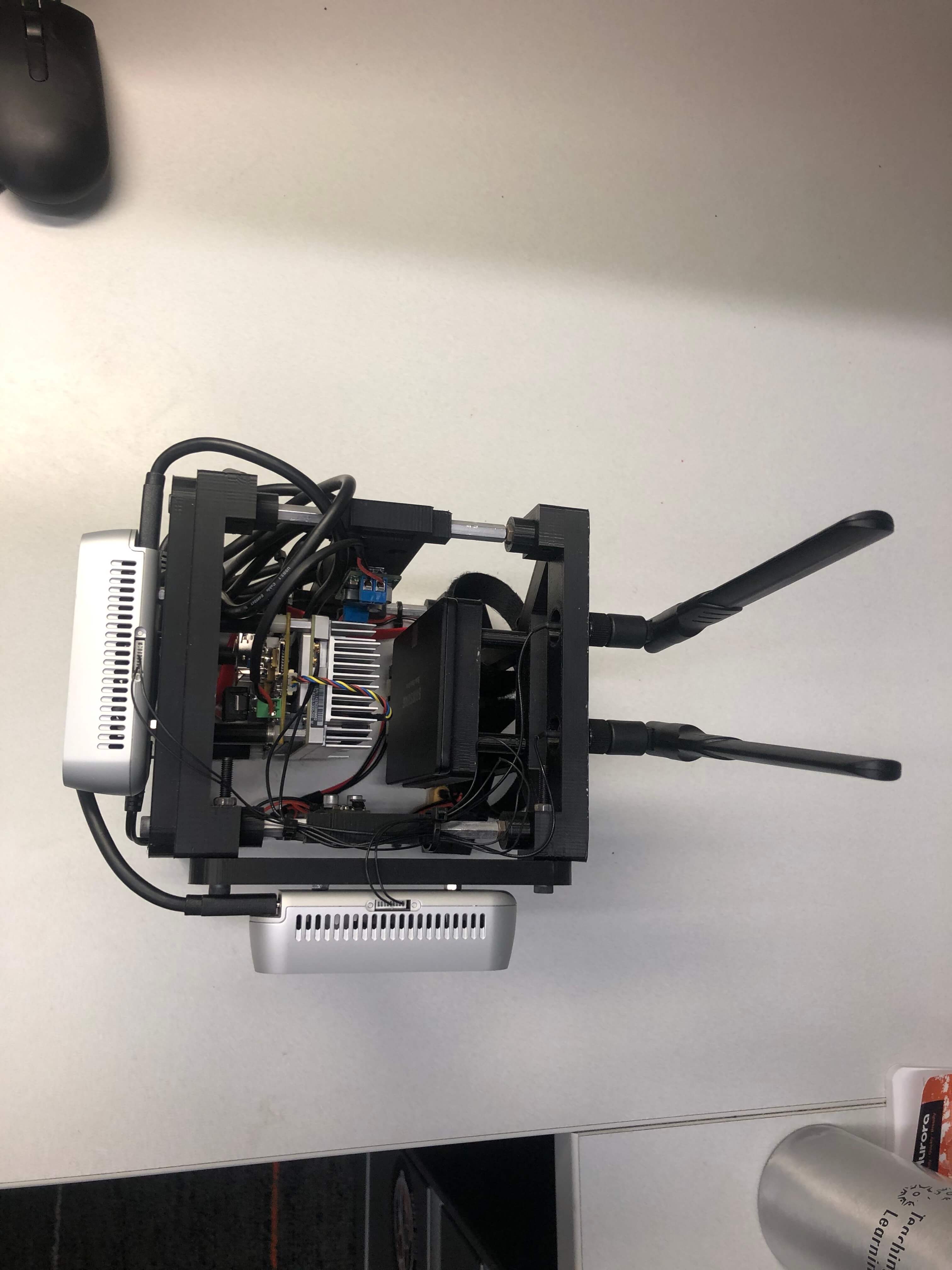} \\
        (b) top view
    \end{minipage}
    \end{tabular}
    \caption{(a)-(b) Data collection rig used for reconstruction of the \BldgThirtyOne and \School scenes, consisting of two \RealSenseLong devices and a NVIDIA Jetson TX2.}
    \label{fig:kimera_rig}
\end{figure}

To collect the \BldgThirtyOne and \School datasets, we use a custom-made sensing rig designed to mount two \RealSenseLong devices in a perpendicular configuration,
 as shown in~\Cref{fig:kimera_rig}.
While a single \RealSenseLong already provides the RGB-D and IMU data needed for \KimeraCore,
 we used two cameras with non-overlapping fields of view because the infrared pattern emitted
 by the depth camera is visible in the images.
This infrared pattern makes the images unsuitable for feature tracking in \KimeraVIO.
Therefore, we disable the infrared pattern emitter for one of the \RealSense cameras, which we use for VIO.
We enable the infrared emitter for the other camera to capture high-quality depth data.
Since the cameras do not have an overlapping field of view, the camera used for tracking is unaffected by
the infrared pattern emitted by the other camera.

Before recording the \BldgThirtyOne and \School dataset, we calibrate the IMUs, the extrinsics, and the intrinsics of both \RealSense cameras.
In particular, the IMU of each \RealSense device is calibrated using the provided script from
Intel\footnote{\url{https://github.com/IntelRealSense/librealsense/tree/master/tools/rs-imu-calibration}}, and the intrinsics and
extrinsics of all cameras are calibrated using the Kalibr toolkit \citep{Furgale13iros}.
Furthermore, the transform between the two \RealSense devices is estimated using the Kalibr extension
for IMU to IMU extrinsic calibration \citep{Rehder16icra-extending}.
Since the cameras do not share the same field of view, we cannot use camera to camera extrinsic calibration.
Both \RealSense devices use hardware synchronization.

The \BldgThirtyOne scene consists of an approximately $40$m loop around the interior of the space that
passes four cubicle groups and a kitchenette, with a standing human visible at two different times.
The \School scene consists of three rooms connected by a corridor and the trajectory is approximately $20$m long.

\begin{figure*}[htbp]
    \centering
    \hspace{-2em}
    \includegraphics[trim={0cm 2.2cm 0cm 2cm}, clip, width=0.85\textwidth]{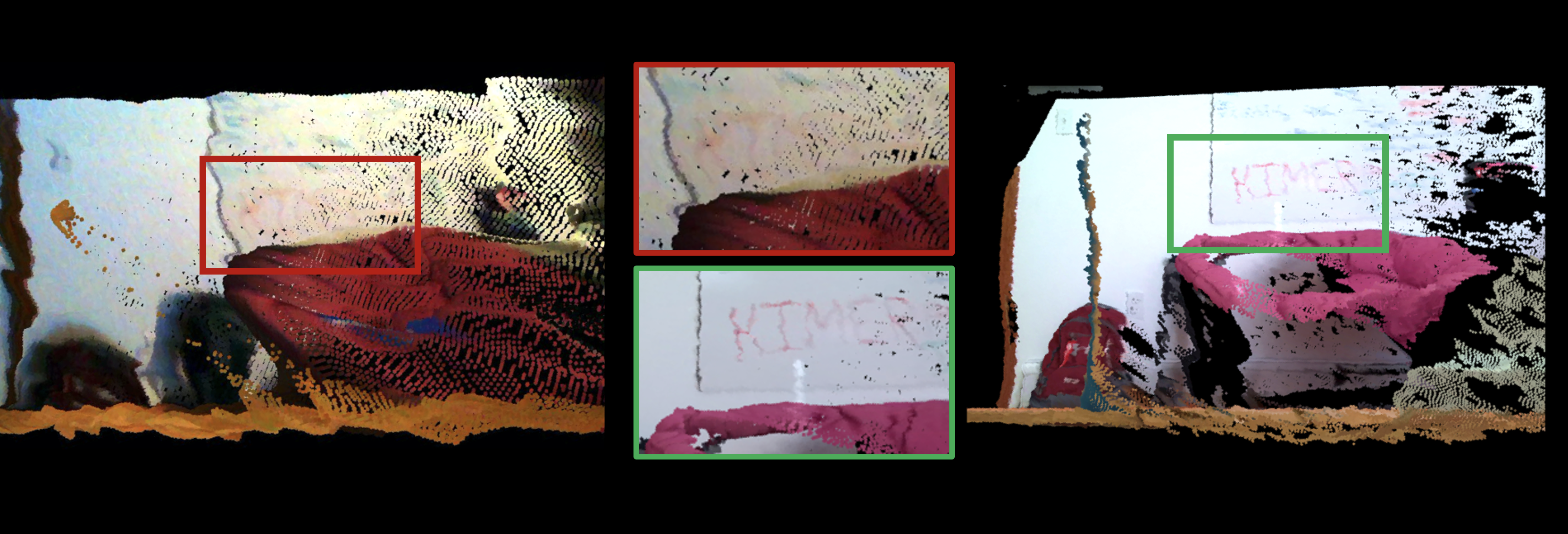}
    \caption{Side-view of the 3D point clouds generated by the \RealSenseLong (left) and \AzureKinect (right) RGB-D cameras of the same scene and from the same viewpoint.
     (Middle) Zoomed-in view shows that the \AzureKinect provides depth estimates of higher quality compared to \RealSense .
     Despite these differences, \Kimera is robust to noise, as shown in \Cref{ssec:real_life_experiments}.
    }
    \label{fig:pcls}
\end{figure*}

To collect the \WhiteOwl scene, we used the \AzureKinect, as it provides a denser, more accurate depth-map than the \RealSenseLong (see \Cref{fig:pcls} for a comparison).
We also calibrate the extrinsincs and intrinsics of the camera and IMU using Kalibr~\citep{Furgale13iros}.
The \WhiteOwl scene consists of an approximately $15$m long trajectory through three rooms: a bedroom, a
kitchen, and a living room. A single seated human is visible in the kitchen area.

Finally, we semantically segment the RGB images for both scenes using Mask-RCNN~\citep{He17iccv-maskRCNN}
with the pre-trained model weights from the COCO dataset~\citep{Abdulla17github-maskrcnn}.

\subsection{Pose Estimation}
\label{sec:experiments-VIO}

\begin{table*}[htbp]
  \centering
  \caption{RMSE [m] for the absolute translation error (ATE) of state-of-the-art VIO pipelines (reported from~\citep{Delmerico18icra}, and the results from the respective papers)
   compared to \Kimera, on the \Euroc dataset.
   In \textbf{bold} the best result for each category: fixed-lag smoothing, and PGO with loop closure. $-$ indicates \rebuttal{missing data}.}
  \label{tab:ape_accuracy_comparison_sopa}
  \setlength{\tabcolsep}{3pt}
  \begin{tabularx}{\textwidth}{*1{Y} *5{Y} || *2{Y} *3{Y}}
    \toprule
    & \multicolumn{10}{c}{RMSE ATE [m]} \\
    \cmidrule(l){2-11}
    & \multicolumn{5}{c}{Fixed-lag Smoothing} & \multicolumn{5}{c}{Loop Closure} \\
    \cmidrule(l){2-6} \cmidrule(l){7-11}
    \specialcell[b]{\Euroc \\ Seq.} & \rotatebox[origin=c]{90}{OKVIS} & \rotatebox[origin=c]{90}{MSCKF} & \rotatebox[origin=c]{90}{ROVIO} & \rotatebox[origin=c]{90}{\specialcell[b]{VINS-\\Mono}} & \rotatebox[origin=c]{90}{\specialcell[b]{\textbf{\!\Kimera-} \\ \textbf{VIO}}}
         & \rotatebox[origin=c]{90}{\specialcell[b]{\rebuttal{Basalt}}}   &
          \rotatebox[origin=c]{90}{\specialcell[b]{\rebuttal{ORB-}\\\rebuttal{SLAM3}}}
         & \rotatebox[origin=c]{90}{\specialcell[b]{VINS-\\LC}} & \rotatebox[origin=c]{90}{\specialcell[b]{\textbf{\Kimera-} \\ \textbf{RPGO}}} 
         & \rotatebox[origin=c]{90}{\specialcell[b]{\textbf{\Kimera-} \\ \textbf{PGMO}}} \\
    \midrule
    MH\_1 & 0.16 & 0.42 & 0.21           & \vinVersion{0.27}{0.15}                    & \textbf{0.11}        & 0.08             & \textbf{0.04}  & \vinVersion{0.07}{0.12}          & 0.13          & 0.09 \\
    MH\_2 & 0.22 & 0.45 & 0.25           & \vinVersion{0.12}{0.15}                    & \textbf{0.10}        & 0.06             & \textbf{0.03}  & \vinVersion{0.05}{0.12}          & 0.21          & 0.11 \\
    MH\_3 & 0.24 & 0.23 & 0.25           & \vinVersion{\textbf{0.13}}{0.22}           & \textbf{0.16}        & 0.05             & \textbf{0.04}  & \vinVersion{0.08}{0.13}          & 0.12          & 0.12 \\
    MH\_4 & 0.34 & 0.37 & 0.49           & \vinVersion{0.23}{0.32}                    & \textbf{0.16}        & 0.10             & \textbf{0.05}  & \vinVersion{0.12}{0.18}          & 0.12          & 0.16 \\
    MH\_5 & 0.47 & 0.48 & 0.52           & \vinVersion{0.35}{0.30}                    & \textbf{0.15}        & \textbf{0.08}    & \textbf{0.08}  & \vinVersion{0.09}{0.21}          & 0.15          & 0.18 \\
    V1\_1 & 0.09 & 0.34 & 0.10           & \vinVersion{0.07}{0.08}                    & \textbf{0.05}        & \textbf{0.04}    & \textbf{0.04}  & \vinVersion{0.07}{0.06}          & 0.06          & 0.05 \\
    V1\_2 & 0.20 & 0.20 & 0.10           & \vinVersion{0.10}{0.11}                    & \textbf{0.08}        & 0.02             & \textbf{0.01}  & \vinVersion{0.06}{0.08}          & 0.05          & 0.06 \\
    V1\_3 & 0.24 & 0.67 & 0.14           & \vinVersion{0.13}{0.18}                    & \textbf{0.13}        & 0.03             & \textbf{0.02}  & \vinVersion{0.11}{0.19}          & 0.11          & 0.13 \\
    V2\_1 & 0.13 & 0.10 & 0.12           & \vinVersion{0.08}{0.08}                    & \textbf{0.06}        & \textbf{0.03}    & \textbf{0.03}  & \vinVersion{0.06}{0.08}          & 0.06          & 0.05 \\
    V2\_2 & 0.16 & 0.16 & 0.14           & \vinVersion{\textbf{0.08}}{0.16}           & \textbf{0.07}        & 0.02             & \textbf{0.01}  & \vinVersion{0.06}{0.16}          & 0.06          & 0.07 \\
    V2\_3 & 0.29 & 1.13 & \textbf{0.14}  & \vinVersion{0.21}{0.27}                    & 0.21                 & $-$              & \textbf{0.02}  & \vinVersion{x}{1.39}             & 0.24          & 0.23 \\
    \bottomrule
  \end{tabularx}
\end{table*}

In this section, we evaluate the performance of \KimeraVIO and \KimeraRPGO.

\Cref{tab:ape_accuracy_comparison_sopa} compares the Root Mean Squared Error (RMSE) of the Absolute Translation Error (ATE) 
of \KimeraVIO 
against state-of-the-art open-source VIO pipelines: OKVIS~\citep{Leutenegger13rss}, MSCKF~\citep{Mourikis07icra},
 ROVIO~\citep{Blosch15iros}, VINS-Mono~\citep{Qin18tro-vinsmono}, 
 \rebuttal{Basalt~\citep{Usenko19ral-basalt}, and ORB-SLAM3~\citep{Campos21-TRO}}.
 using the independently reported values in~\citep{Delmerico18icra} and the \rebuttal{self-reported values from the respective authors}.
Note that OKVIS, MSCKF, ROVIO, and VINS-Mono use a monocular camera, while the rest use a stereo camera. 
We align the estimated and ground-truth trajectories using an \SEthree transformation before evaluating the errors. 
Using a $\mathrm{Sim}(3)$ alignment, as in~\citep{Delmerico18icra}, would result in an even smaller error for 
\Kimera: we preferred the \SEthree alignment, since it is more appropriate 
for VIO, where the scale is observable thanks to the IMU.
We group the techniques depending on whether they use fixed-lag smoothing or loop closures.
Kimera-VIO, \KimeraRPGO, \rebuttal{and \KimeraPGMO (see \cref{ssec:experiments-MeshLoopClosure})}
achieve \rebuttal{competitive} performance.
\rebuttal{We also compared \KimeraVIO with SVO-GTSAM~\citep{Forster14icra,Forster15rss-imuPreintegration} in our previous paper~\citep{Rosinol20icra-Kimera}.}

\begin{table}[htbp]
  \centering
  \caption{RMSE ATE [m] vs. loop closure threshold $\alpha$, on the V1\_01 \Euroc dataset.}
  \label{tab:ape_alpha}
  \begin{tabular}{l|ccccc} %
    \toprule
    \diagbox[width=2.3cm]{}{$\alpha$} & $10^1$ &  $10^0$ &  $10^{-1}$ &  $10^{-2}$ & $10^{-3}$ \\
    \midrule
      PGO w/o PCM          & 0.05 & 0.45 & 1.74 & 1.59 & 1.59 \\
      \textbf{\KimeraRPGO} & \textbf{0.05} & \textbf{0.05} & \textbf{0.05} & \textbf{0.05} &  \textbf{0.05} \\
    \bottomrule
  \end{tabular}
\end{table}

Furthermore, \KimeraRPGO ensures robust performance, and is less sensitive to loop closure parameter tuning.
\Cref{tab:ape_alpha} shows the \KimeraRPGO accuracy with and without 
 outlier rejection (PCM) for different values of the loop closure threshold $\alpha$ used in DBoW2.
 Small values of $\alpha$ lead to more loop closure detections, but these are less conservative (more outliers).
\Cref{tab:ape_alpha} shows that, by using PCM, \KimeraRPGO is fairly insensitive to the choice of $\alpha$.
 The results in \Cref{tab:ape_accuracy_comparison_sopa} use $\alpha=0.001$.

\subsubsection{Robustness of Pose Estimation in Dynamic Scenes.}
\label{ssec:robust_pose_estimation}

Table~\ref{tab:robustVIO} reports the absolute trajectory errors of \Kimera when using 5-point RANSAC,
2-point RANSAC, and when using 2-point RANSAC and IMU-aware feature tracking (label: \KimeraDVIO). 
Best results (lowest errors) are shown in bold.

The rows from MH\_01--V2\_03, corresponding to tests on the static \Euroc dataset, 
confirm that, in absence of dynamic agents, the proposed approach performs on-par with the state of the art, 
while the use of 2-point RANSAC already boosts performance.
The rest of rows  (\UnityHumans and \UnityHumansTwo), further show that in the presence of an increasing number of dynamic entities (third column with number of humans), 
the proposed DVIO approach remains robust.

\subsection{Geometric Reconstruction}
\label{ssec:geometric_performance}

We now show how \Kimera's accurate pose estimates and robustness against dynamic scenes improve the geometric accuracy of the reconstruction. 

We use the ground truth point cloud available in the \Euroc \texttt{V1} and \texttt{V2} datasets to assess the quality of the 3D meshes
produced by \Kimera.
We evaluate each mesh against the ground truth using the \textit{accuracy} and \textit{completeness} metrics as in~\citep[Sec. 4.3]{Rosinol18thesis}:
(i) we compute a point cloud by sampling our mesh with a uniform density of $10^3~\text{points}/\text{m}^2$, 
(ii) we register the estimated and the ground truth clouds with ICP~\citep{Besl92pami} using \emph{CloudCompare}~\citep{CloudCompare}, 
and (iii) we evaluate the average distance from ground truth point cloud to its nearest neighbor in the estimated point cloud (accuracy), and vice-versa (completeness).

\begin{figure}[htbp]
  \centering
  \includegraphics[width=0.7\columnwidth]{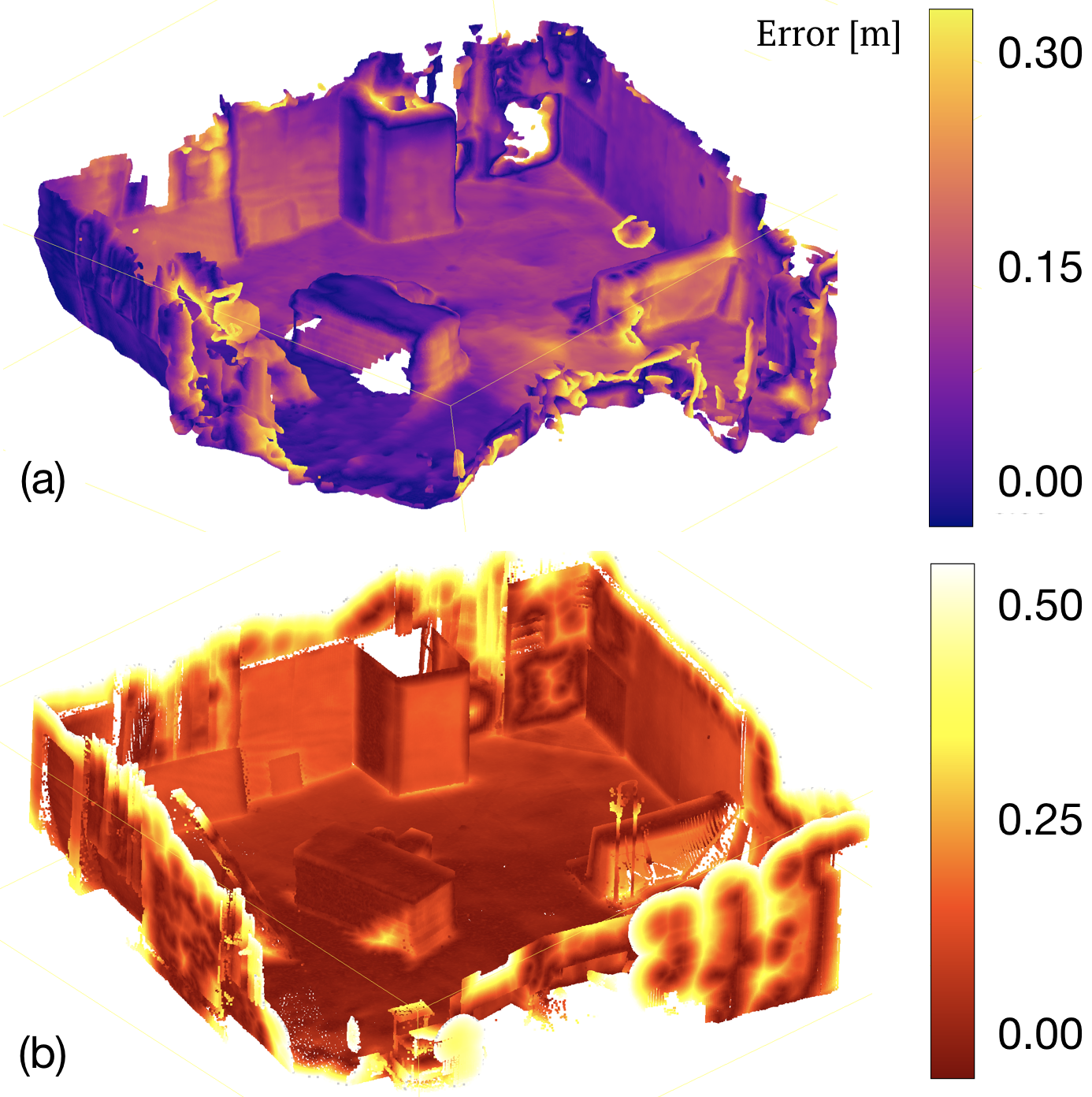}
  \caption{
    (a) \Kimera's 3D mesh color-coded by the distance to the ground-truth point cloud.
    (b) Ground-truth point cloud color-coded by the distance to the estimated  cloud. 
    \Euroc  V1\_01 dataset.}
  \label{fig:accuracy_mesh}
\end{figure}

\Cref{fig:accuracy_mesh}(a) shows the estimated cloud (corresponding to the global mesh of \KimeraSemantics on V1\_01) color-coded  by the distance to the closest point
in the ground-truth cloud (accuracy); \Cref{fig:accuracy_mesh}(b) shows the ground-truth cloud,
 color-coded with the distance to the closest-point in the estimated cloud (completeness).

Table~\ref{tab:geometric_accuracy} provides a quantitative comparison between the fast multi-frame mesh produced by \KimeraMesher and the slow mesh produced via TSDF by \KimeraSemantics. To obtain a complete mesh from \KimeraMesher we set a large VIO horizon 
(\ie we perform full smoothing).

As expected from \Cref{fig:accuracy_mesh}(a), the global mesh from \KimeraSemantics is very accurate, with an average error of $0.35-0.48$m across datasets. \KimeraMesher produces a more noisy mesh (up to $24\%$ error increase), 
but %
requires two orders of magnitude less time to compute (see Section~\ref{ssec:timing_performance}).

\begin{table}[htbp]
  \centering
  \caption{Evaluation of \Kimera multi-frame and global meshes' completeness \citep[Sec. 4.3.3]{Rosinol18thesis} with an ICP threshold of $1.0$m.}
  \label{tab:geometric_accuracy}
  \begin{tabular}{cccc}
    \toprule
      & \multicolumn{2}{c}{\specialcell[b]{RMSE [m]}} & \multirow{2}{*}{\specialcell[b]{Relative \\ Improvement [\%]}} \\
    \cmidrule(l{2pt}r{2pt}){2-3}
    \multicolumn{1}{c}{\specialcell{\Euroc \\ Seq.}} & \multicolumn{1}{c}{\specialcell[b]{Multi-Frame}} & \multicolumn{1}{c}{\specialcell[b]{Global}} &   \\
    \cmidrule(l{2pt}r{2pt}){1-1} \cmidrule(l{2pt}r{2pt}){2-2} \cmidrule(l{2pt}r{2pt}){3-3} \cmidrule(l{2pt}r{2pt}){4-4}
    V1\_01 & 0.482 & 0.364 & 24.00 \\ 
    V1\_02 & 0.374 & 0.384 & -2.00 \\
    V1\_03 & 0.451 & 0.353 & 21.00 \\
    V2\_01 & 0.465 & 0.480 & -3.00 \\
    V2\_02 & 0.491 & 0.432 & 12.00 \\
    V2\_03 & 0.530 & 0.411 & 22.00 \\
    \bottomrule
  \end{tabular}
\end{table}

\subsubsection{Robustness of Mesh Reconstruction in Dynamic Scenes.}
\label{sec:exp-dynamicScenes}

Here we show that the the enhanced robustness against dynamic objects afforded by \KimeraDVIO (and quantified in \Cref{ssec:robust_pose_estimation}), combined with \emph{dynamic masking} (\Cref{ssec:humans}), 
results in robust and accurate metric-semantic meshes in crowded dynamic environments.

\begin{table*}[htbp]
  \centering
  \caption{RMSE for the absolute translation error in meters when using 5-point, 2-point, and \KimeraDVIO poses, as well as \KimeraRPGO and \KimeraPGMO's optimized trajectory.
  We show the drift in \% for \KimeraDVIO to account for the length of the different trajectories.}
  \label{tab:robustVIO}
  \begin{tabular}{cccccc|cc}
    \toprule
     & & & \multicolumn{5}{c}{Absolute Translation Error [m] (Drift [\%])} \\
    \cmidrule(l{2pt}r{2pt}){4-8}
    & & & \multicolumn{3}{c}{\textbf{\KimeraVIO}} & \multicolumn{2}{c}{Loop Closure} \\
    \cmidrule(l{2pt}r{2pt}){4-6} \cmidrule(l{2pt}r{2pt}){7-8}
    Dataset & Scene & \specialcell{\# of \\ Humans} & \makecell{5-point} & \makecell{2-point} & \textbf{\makecell{DVIO}}
      & \specialcell[b]{\Kimera- \\ RPGO}
      & \specialcell[b]{\textbf{\Kimera-} \\ \textbf{PGMO}} \\
    \midrule
    \multirow{11}{*}{\Euroc}
                               & MH_01 & 0 & \textbf{0.09} & 0.14           & 0.11 (0.1)            & 0.13          & \textbf{0.09} \\
                               & MH_02 & 0 & 0.10          & 0.12           & \textbf{0.10} (0.1)   & 0.21          & \textbf{0.11} \\
                               & MH_03 & 0 & \textbf{0.11} & 0.17           & 0.16 (0.1)            & 0.12          & \textbf{0.12} \\
                               & MH_04 & 0 & 0.42          & 0.19           & \textbf{0.16} (0.2)   & \textbf{0.12} & 0.16          \\
                               & MH_05 & 0 & 0.21          & \textbf{0.14}  & 0.15          (0.2)   & \textbf{0.15} & 0.18          \\
                               & V1_01 & 0 & 0.07          & 0.07           & \textbf{0.05} (0.1)   & 0.06          & \textbf{0.05} \\
                               & V1_02 & 0 & 0.12          & 0.08           & \textbf{0.08} (0.1)   & \textbf{0.05} & 0.06          \\
                               & V1_03 & 0 & 0.17          & 0.13           & \textbf{0.13} (0.2)   & \textbf{0.11} & 0.13          \\
                               & V2_01 & 0 & \textbf{0.05} & 0.06           & 0.06          (0.2)   & 0.06          & \textbf{0.05} \\
                               & V2_02 & 0 & 0.08          & 0.07           & \textbf{0.07} (0.1)   & \textbf{0.06} & 0.07          \\
                               & V2_03 & 0 & 0.30          & 0.27           & \textbf{0.21} (0.3)   & 0.24          & \textbf{0.23} \\
    \cmidrule(l{1pt}r{1pt}){1-8}
    \multirow{3}{*}{\UnityHumans}  & \multirow{3}{*}{Office}   & 12 & 0.92   & 0.78    & \textbf{0.59} (0.2)                          & 0.68 &  \textbf{0.66} \\
                               &                               & 24 & 1.45   & 0.79    & \textbf{0.78} (0.4)                          & 0.78 &  \textbf{0.78} \\
                               &                               & 60 & 1.60   & 1.11    & \textbf{0.88} (0.4)                          & 0.72 &  \textbf{0.61} \\
    \cmidrule(l{1pt}r{1pt}){1-8}
    \multirow{12}{*}{\UnityHumansTwo} & \multirow{3}{*}{Office}& 0  & 0.47   & 0.46              & \textbf{{0.46}} (0.2)              & 0.46 & \textbf{0.21} \\
                               &                               & 6  & 0.50   & \textbf{{0.48}}   & 0.50 (0.2)                         & 0.49 & \textbf{0.47} \\
                               &                               & 12 & 0.50   & 0.50              & \textbf{0.45} (0.2)                & 0.45 & \textbf{0.32} \\
    \cmidrule(l{1pt}r{1pt}){2-8}
                               & \multirow{3}{*}{Neighborhood} & 0  & 3.67              & 5.77              & \textbf{3.37}  (0.8)    & 2.78  & \textbf{1.70} \\
                               &                               & 24 & $\times$          & $\times$          & \textbf{6.65} (1.6)     & 3.76  & \textbf{3.01} \\
                               &                               & 36 & $\times$          & $\times$          & \textbf{11.58} (2.7)    & 1.74 & \textbf{1.48} \\
    \cmidrule(l{1pt}r{1pt}){2-8}
                               & \multirow{3}{*}{Subway}       & 0  & 3.38              & 2.65              & \textbf{{1.79}}  (0.4)  & 1.68 & \textbf{1.47} \\
                               &                               & 24 & $\times$          & $\times$          & \textbf{2.37}    (0.5)  & 1.92 & \textbf{0.82} \\
                               &                               & 36 & $\times$          &  1.70             & \textbf{{1.14}}  (0.2)  & 0.87 & \textbf{0.68} \\
    \cmidrule(l{1pt}r{1pt}){2-8}
                               & \multirow{3}{*}{Apartment}    & 0  & 0.08               & 0.07               & \textbf{0.07} (0.1)   & \textbf{0.07} & 0.08 \\
                               &                               & 1  & 0.09               & 0.07               & \textbf{0.07} (0.1)   & 0.07 & \textbf{0.07} \\
                               &                               & 2  & 0.08               & 0.08               & \textbf{0.07} (0.1)   & 0.07 & \textbf{0.07} \\
    \bottomrule
  \end{tabular}
\end{table*}{}

\renewcommand{\myhspace}{\hspace{0mm}}
\renewcommand{\mpw}{4.2cm}
\renewcommand{\myRate}{1}

\begin{figure}[htbp]
	\begin{center}
	\begin{minipage}{\textwidth}
	\begin{tabular}{cc}%
	\myhspace \hspace{-3mm}
			\begin{minipage}{\mpw}%
			\centering%
			\includegraphics[trim=400mm 6mm 500mm 50mm, clip,width=\myRate\columnwidth]{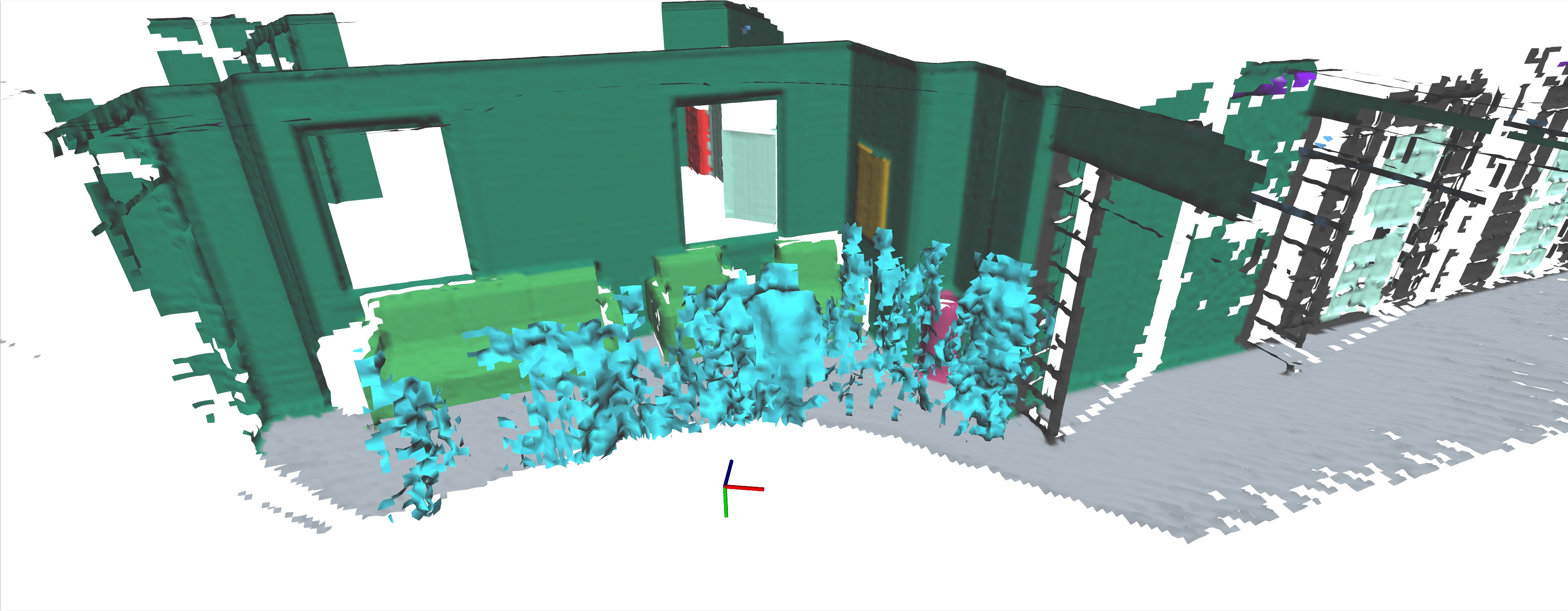} 
			\vspace{-18mm}\\
			\hspace{-1cm}(a) 
			\end{minipage}
		& \myhspace 
			\begin{minipage}{\mpw}%
			\centering%
			\includegraphics[trim=400mm 6mm 500mm 50mm, clip,width=\myRate\columnwidth]{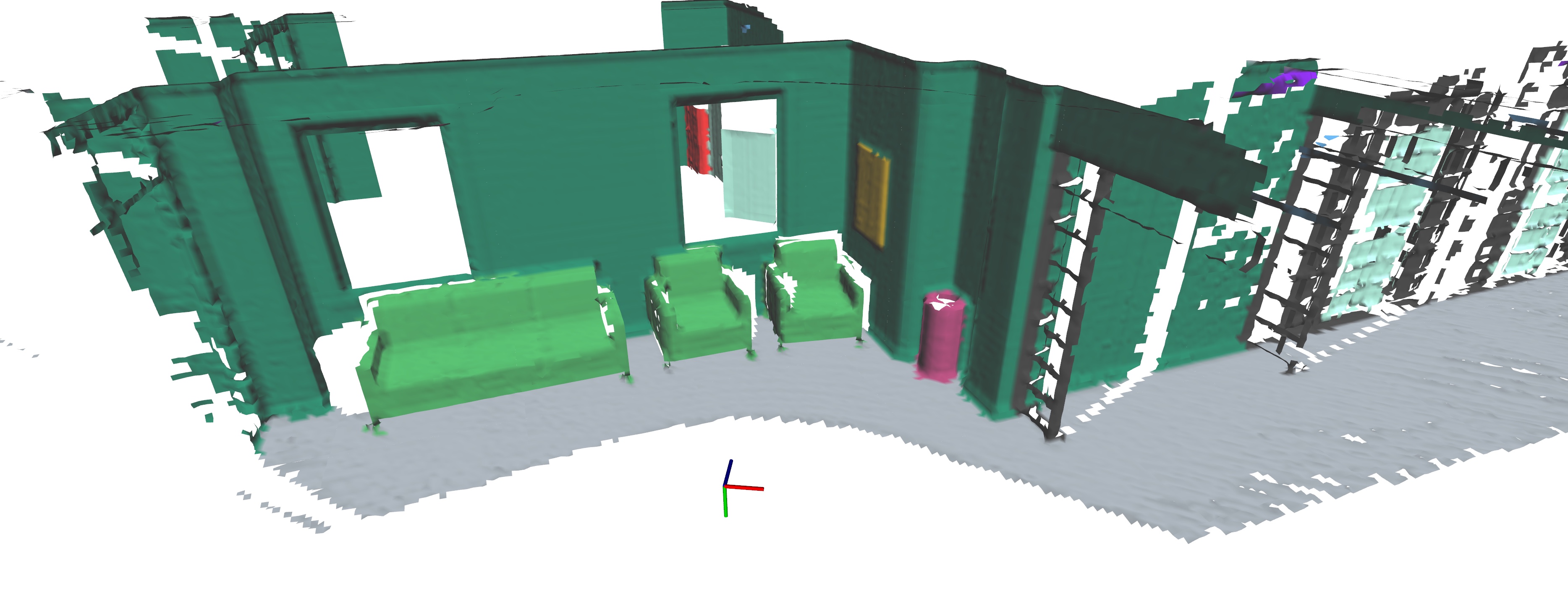} 
			\vspace{-18mm}\\
			\hspace{-1cm}(b) 
			\end{minipage}
		\end{tabular}
	\end{minipage}
	\begin{minipage}{\textwidth}
	\end{minipage}
	\caption{3D mesh reconstruction (a) without and (b) with \emph{dynamic masking}.
	 Note that the human moves from right to left, while the robot with the camera rotates back and forth when mapping this scene.
	 \label{fig:dynamicMesh}}
	\end{center}
\end{figure}

\myParagraph{Dynamic Masking}
\Cref{fig:dynamicMesh} visualizes the effect of dynamic masking on \Kimera's metric-semantic mesh reconstruction.
\Cref{fig:dynamicMesh}(a) shows that without dynamic masking a human walking in front of the camera leaves 
a ``contrail'' (in cyan) and creates artifacts in the mesh. 
\Cref{fig:dynamicMesh}(b) shows that 
dynamic masking avoids this issue and leads to clean mesh reconstructions. 
\Cref{tab:mapAccuracy} reports the RMSE mesh error (see \emph{accuracy metric} in~\citep{Rosinol19icra-mesh}) 
with and without dynamic masking. 
To assess the mesh accuracy independently from the VIO localization errors, we also report 
the mesh geometric errors when using ground-truth poses (``GT pose'' column in \Cref{tab:mapAccuracy}) compared to the mesh errors when using VIO poses
(``DVIO pose'' column).
The ``GT pose'' columns in the table show that even with perfect localization, the artifacts created by dynamic entities
(and visualized in Fig.~\ref{fig:dynamicMesh}(a)) significantly hinder the mesh accuracy,
while dynamic masking ensures highly accurate reconstructions.
The advantage of dynamic masking is preserved when VIO poses are used.
\rebuttal{It is worth mentioning that the 3D mesh errors in \Cref{tab:mapAccuracy} are small compared to the localization errors from \Cref{tab:ape_accuracy_comparison_sopa}.
This is because the floor is the most visible surface in all reconstructions,
and since our z-estimate (height) of the trajectory is very accurate,
most of the mesh errors are small.
We also use ICP to align the ground-truth 3D mesh with the estimated 3D mesh, which further reduces the mesh errors.}

\begin{table*}[htbp]
  \centering
  \caption{3D Mesh RMSE in meters with and without Dynamic Masking (DM) for \KimeraSemantics when using ground-truth pose (GT pose) and \KimeraDVIO's estimated pose (DVIO pose) in the \UnityHumans and \UnityHumansTwo datasets. The third column shows the number of humans in each scene (\# of Humans).}
  \label{tab:mapAccuracy}
  \begin{tabular}{ccccccc}
    \toprule
    & & & \multicolumn{4}{c}{\KimeraSemantics 3D mesh RMSE [m]}\\
      \cmidrule(l{2pt}r{2pt}){4-7}
     & & &
      \multicolumn{2}{c}{GT pose} &
      \multicolumn{2}{c}{DVIO pose}
      \\
      \cmidrule(l{2pt}r{2pt}){4-5}
      \cmidrule(l{2pt}r{2pt}){6-7}
    Dataset & Scene &  \specialcell{\# of \\ Humans}  & \makecell{Without DM} & With DM & \makecell{Without DM} & \makecell{With DM} \\
    \midrule
    \multirow{3}{*}{\UnityHumans}  & \multirow{3}{*}{Office} & 12  & 0.09 & \textbf{0.06} & 0.23 & \textbf{0.23} \\
                               &                         & 24  & 0.13 & \textbf{0.06} & 0.35 & \textbf{0.30} \\
                               &                         & 60  & 0.19 & \textbf{0.06} & 0.35 & \textbf{0.33} \\
    \cmidrule(l{1pt}r{1pt}){1-7}
    \multirow{12}{*}{\UnityHumansTwo} & \multirow{3}{*}{Office} & 0  & 0.03 & \textbf{0.03} & 0.16  & \textbf{0.16} \\
                               &                         & 6  & 0.03 & \textbf{0.03} & 0.21  & \textbf{0.17} \\
                               &                         & 12 & 0.03 & \textbf{0.03} & 0.18  & \textbf{0.13} \\
    \cmidrule(l{1pt}r{1pt}){2-7}
                               & \multirow{3}{*}{Neighborhood} & 0  & 0.06 & \textbf{0.06} & 0.27 & \textbf{0.27} \\
                               &                               & 24 & 0.08 & \textbf{0.06} & 0.66 & \textbf{0.61} \\
                               &                               & 36 & 0.08 & \textbf{0.06} & 0.70 & \textbf{0.65} \\
    \cmidrule(l{1pt}r{1pt}){2-7}
                               & \multirow{3}{*}{Subway}    & 0  & 0.06 & \textbf{0.06} & 0.42 & \textbf{0.42} \\
                               &                            & 24 & 0.19 & \textbf{0.06} & 0.58 & \textbf{0.53} \\
                               &                            & 36 & 0.19 & \textbf{0.06} & 0.49 & \textbf{0.43} \\
    \cmidrule(l{1pt}r{1pt}){2-7}
                               & \multirow{3}{*}{Apartment} & 0  & 0.05 & \textbf{0.05} & 0.06 & \textbf{0.06} \\
                               &                            & 1  & 0.05 & \textbf{0.05} & 0.07 & \textbf{0.07} \\
                               &                            & 2  & 0.05 & \textbf{0.05} & 0.07 & \textbf{0.07} \\
    \bottomrule
  \end{tabular}
\end{table*}{}

\subsection{Semantic Reconstruction}
\label{ssec:semantic_performance}

\begin{table}[htbp]
  \centering
  \caption{Evaluation of \KimeraSemantics \rebuttal{ using the simulated dataset from \cite{Rosinol20icra-Kimera}, which has no humans},
           using a combination of ground-truth (GT) and dense stereo (Stereo) depth maps,
           as well as ground-truth (GT) and \KimeraDVIO (DVIO) poses. }
  \label{tab:semantic_accuracy}
  \begin{tabular}{ccccc}
    \toprule
     & \multicolumn{4}{c}{\specialcell[b]{\KimeraSemantics} using:} \\
     \cmidrule(l{2pt}r{2pt}){2-5}
                         & Depth from: & GT & GT  & Stereo \\
    \multicolumn{1}{c}{} & Poses from: & GT & DVIO & DVIO \\
    \midrule
    \multirow{2}{*}{\specialcell{Semantic \\ Metrics}} 
                              & mIoU [\%] & 80.10 & 80.03 & 57.23 \\ 
                              & Acc [\%]  & 94.68 & 94.50 & 80.74 \\
    \midrule
    Geometric    & RMSE [m] & 0.079 & 0.131 & 0.215 \\
    \midrule
    Localization & ATE [m] & 0.00 & 0.04 & 0.04 \\
                 & \rebuttal{Drift [\%]} & \rebuttal{0.0} & \rebuttal{0.2} & \rebuttal{0.2} \\
    \bottomrule
  \end{tabular}
\end{table}

\KimeraSemantics builds a 3D mesh from the VIO pose estimates,
 and uses a combination of dense stereo (or RGB-D if available) and bundled raycasting.
We evaluate the impact of each of these components by running three different experiments.
\rebuttal{For these experiments, we use the simulated dataset in~\citep{Rosinol20icra-Kimera},
which has ground-truth semantics and geometry,
and allows us to determine the effects of each module on the performance.}

First, we use \KimeraSemantics with ground-truth (GT) poses and ground-truth depth maps (available in simulation) to assess the initial loss of performance due to bundled raycasting.
Second, we use \KimeraVIO's pose estimates. %
Finally, we use the full \KimeraSemantics pipeline including dense stereo.
To analyze the semantic performance, we calculate the mean Intersection over Union (mIoU)~\citep{Hackel17arxiv-semantic3d},
 and the overall portion of correctly labeled points (Acc)~\citep{Wolf15ral}.
We also report the ATE to correlate the results with the drift incurred by \KimeraVIO.
Finally, we evaluate the metric reconstruction
registering the estimated mesh with the ground truth and computing the RMSE for the points as in Section~\ref{ssec:geometric_performance}.

\Cref{tab:semantic_accuracy} summarizes our findings and shows that bundled raycasting
results in a small drop in performance both geometrically ($<\!8$cm error on the 3D mesh) as well as semantically (accuracy $>\!94\%$).
Using \KimeraVIO also results in negligible loss in performance since our VIO has a small drift ($<0.2\%$, $4$cm for a $32$m long trajectory).
Certainly, the biggest drop in performance is due to the use of dense stereo.
Dense stereo~\citep{Hirschmuller08pami} %
has difficulties resolving the depth of textureless regions such as walls, which are frequent in
 simulated scenes.

\begin{figure}[htbp]
  \centering
  \vspace{-0.5mm}
  \includegraphics[trim={0cm 2mm 0cm 0cm}, clip, width=\columnwidth]{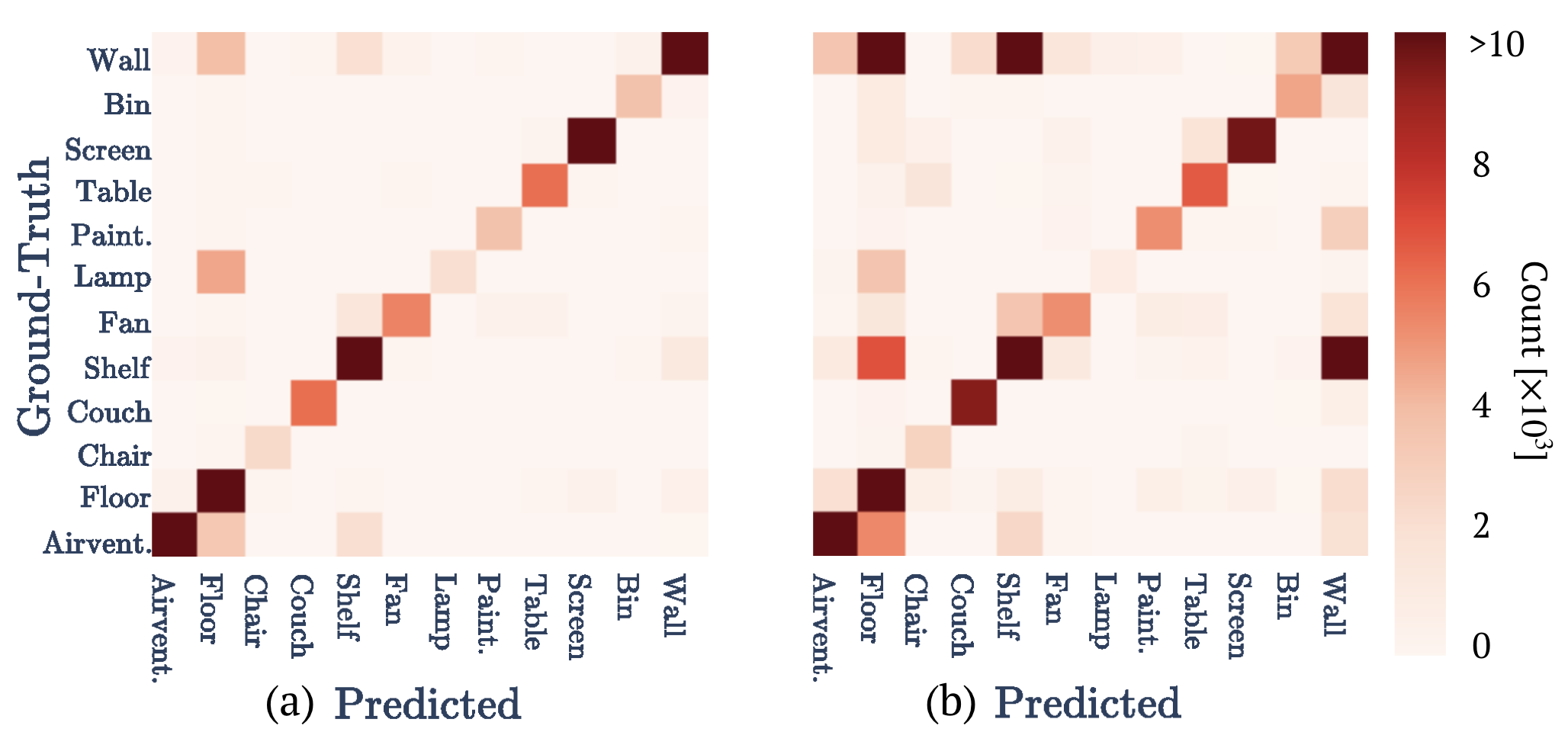}
  \caption{Confusion matrices for \KimeraSemantics using bundled raycasting and (a) ground truth stereo depth or (b) dense stereo~\citep{Hirschmuller08pami}.  Both experiments use ground-truth 2D semantics.
  Values are saturated to $10^4$ for visualization purposes. \vspace{-2mm}}
  \label{fig:confusion_matrix}
  \vspace{-1mm}
\end{figure}

\Cref{fig:confusion_matrix} shows the confusion matrix when running \KimeraSemantics with \KimeraVIO and ground-truth depth (\Cref{fig:confusion_matrix}(a)), compared with using dense stereo (\Cref{fig:confusion_matrix}(b)).
Large  values in the confusion matrix appear between \textit{Wall/Shelf} and \textit{Floor/Wall}.
This is exactly where dense stereo suffers the most; textureless walls are difficult to reconstruct and are close to shelves and floor, resulting in increased geometric and semantic errors.

\subsection{Loop Closure and Mesh Deformation}
\label{ssec:experiments-MeshLoopClosure}

\myParagraph{Localization and geometric evaluation in \Euroc}
We evaluate the performance of \KimeraPGMO in terms of the localization errors in \Cref{tab:robustVIO} (last column).
We observe that \KimeraPGMO achieves significant improvements on large-scale scenes (such as the `Neighborhood' and `Subway' scenes) where the accumulated ATE is noticeable.
Instead, in \Euroc, where \KimeraDVIO already achieves very accurate localization, \KimeraPGMO may only deliver marginal gains in localization.
Note that the localization errors of \KimeraPGMO in \Cref{tab:robustVIO} differ from the ones of \KimeraRPGO in \Cref{tab:ape_accuracy_comparison_sopa},
because \KimeraRPGO does not optimize the mesh, contrary to \KimeraPGMO.

\begin{figure}[htbp]
\centering
   \begin{minipage}{.49\textwidth}%
    \centering
    \includegraphics[width=\textwidth,trim=0mm 0mm 0mm 0mm,clip]{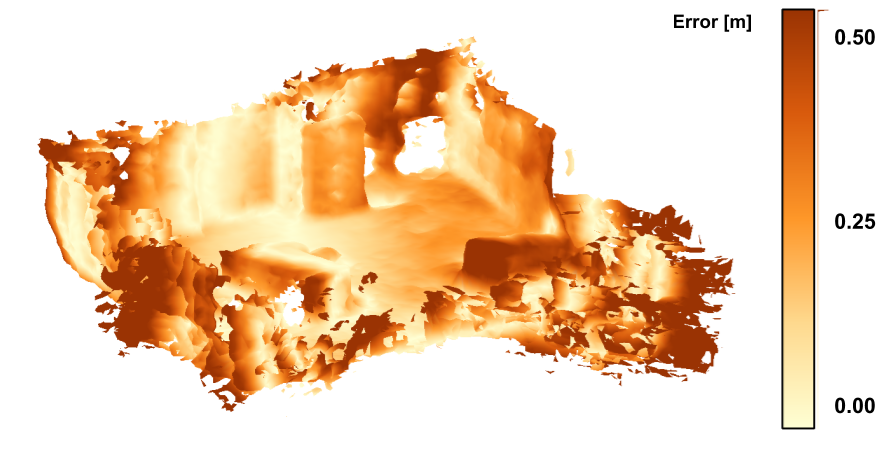} \\
    (a) Before deformation with Kimera-PGMO.
    \end{minipage}
\hspace{1mm}
    \begin{minipage}{.49\textwidth}%
    \centering
    \includegraphics[width=\textwidth,trim=0mm 0mm 0mm 0mm,clip]{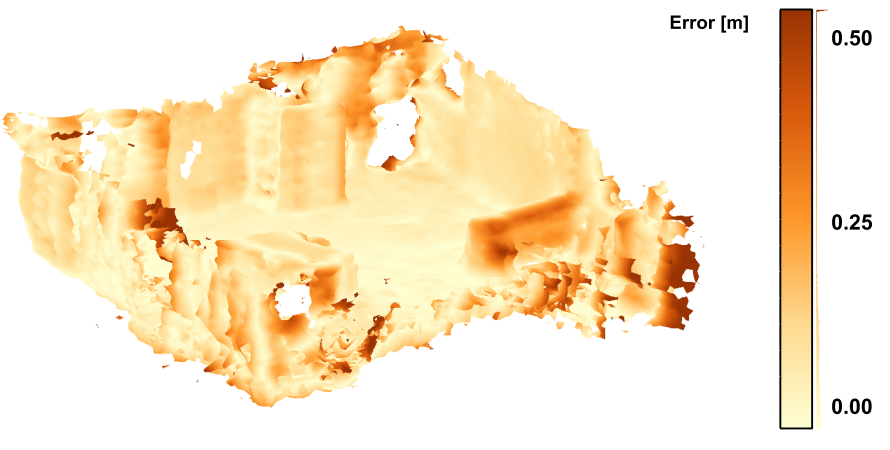} \\
    (b) After deformation with Kimera-PGMO.
    \end{minipage}
\caption{\rebuttal{Reconstruction from \Euroc V1\_01 dataset, before and after mesh deformation from \KimeraPGMO.}
   The trajectory RMSE before loop closures are applied is $15$ cm and 
   the trajectory RMSE after loop closures are optimized in \KimeraPGMO is $13$ cm.}
 \label{fig:euroc-mesh-deformation}
\end{figure}

We evaluate \KimeraPGMO in terms of the geometric errors associated to the deformed 3D mesh in the subset of the \Euroc dataset with ground-truth point clouds.
\Cref{tab:pgmoUhumans} shows that \KimeraPGMO's 3D mesh achieves better geometric accuracy than the unoptimized \KimeraSemantics 3D mesh.
\rebuttal{For a qualitative visualization of the effects of \KimeraPGMO, \Cref{fig:euroc-mesh-deformation} shows the mesh reconstruction before and after deformation.}

\myParagraph{Metric-semantic evaluation in \UnityHumans and \UnityHumansTwo}
We further evaluate the impact that deforming a mesh has on the metric-semantic reconstruction in the \UnityHumans and \UnityHumansTwo datasets.
\Cref{tab:pgmoUhumans} shows the RMSE of the reconstruction error, along with the percentage of correct semantics matches, for \KimeraPGMO compared to \KimeraDVIO.
We observe that the mesh deformation from \KimeraPGMO achieves the best geometric and semantic performance, compared to the non-optimized 3D mesh from \KimeraSemantics when using the \KimeraDVIO pose estimates.

\begin{table*}[htbp]
  \centering
  \caption{3D mesh RMSE [m] and percentage of correct semantic labels when using \KimeraSemantics and \KimeraPGMO 
  (both using dynamic masking and DVIO poses), in the \UnityHumans and \UnityHumansTwo dataset,
   and the subset of \Euroc datasets having ground-truth point clouds.
  \Euroc's ground-truth point clouds do not have semantic labels.}
  \label{tab:pgmoUhumans}
  \begin{tabular}{ccccccc}
    \toprule
     & & &
      \multicolumn{2}{c}{Geometric Reconstruction [m]} &
      \multicolumn{2}{c}{Semantic Reconstruction [\%]}
      \\
      \cmidrule(l{2pt}r{2pt}){4-5}
      \cmidrule(l{2pt}r{2pt}){6-7}
    Dataset & Scene &  \# of Humans & \KimeraSemantics & \makecell{\KimeraPGMO} &  \makecell{\KimeraSemantics} & \makecell{\KimeraPGMO} \\
    \midrule
    \multirow{6}{*}{\Euroc} & V1_01 & 0 & 0.15 & \textbf{0.13} & --  & -- \\
                            & V1_02 & 0 & 0.13 & \textbf{0.11} & --  & -- \\
                            & V1_03 & 0 & 0.22 & \textbf{0.19} & --  & -- \\
                            & V2_01 & 0 & 0.23 & \textbf{0.19} & --  & -- \\
                            & V2_02 & 0 & 0.19 & \textbf{0.17} & --  & -- \\
                            & V2_03 & 0 & 0.25 & \textbf{0.24} & --  & -- \\
    \midrule
    \multirow{3}{*}{\UnityHumans}  & \multirow{3}{*}{Office} & 12  & 0.23 & \textbf{0.20}  & 78.5 & \textbf{79.1} \\
                               &                         & 24  & 0.30 & \textbf{0.17}  & 73.2 & \textbf{81.6} \\
                               &                         & 60  & 0.33 & \textbf{0.24}  & 70.1 & \textbf{72.3} \\
    \cmidrule(l{1pt}r{1pt}){1-7}
    \multirow{12}{*}{\UnityHumansTwo} & \multirow{3}{*}{Office} & 0    & 0.16 & \textbf{0.12}  & 72.7 & \textbf{78.3} \\
                               &                         & 6    & 0.17 & \textbf{0.10}  & 71.3 & \textbf{81.5} \\
                               &                         & 12   & \textbf{0.13} & 0.15  & \textbf{76.1} & 73.8 \\
    \cmidrule(l{1pt}r{1pt}){2-7}
                               & \multirow{3}{*}{Neighborhood} & 0    & 0.27 & \textbf{0.09}  & 62.9  & \textbf{93.3} \\
                               &                               & 24   & 0.61 & \textbf{0.11}  & 51.2  & \textbf{93.7} \\
                               &                               & 36   & 0.65 & \textbf{0.43}  & 53.0  & \textbf{81.2} \\
    \cmidrule(l{1pt}r{1pt}){2-7}
                               & \multirow{3}{*}{Subway}    & 0   & 0.42  & \textbf{0.26}  & 80.1 & \textbf{89.3} \\
                               &                            & 24  & 0.53  & \textbf{0.31}  & 73.5 & \textbf{86.3} \\
                               &                            & 36  & 0.43  & \textbf{0.39}  & 81.3 & \textbf{82.8} \\
    \cmidrule(l{1pt}r{1pt}){2-7}
                               & \multirow{3}{*}{Apartment} & 0   & 0.06  & \textbf{0.06} & 65.5 & \textbf{65.9} \\
                               &                            & 1   & 0.08  & \textbf{0.08} & 61.1 & \textbf{65.1} \\
                               &                            & 2   & 0.08  & \textbf{0.08} & 61.1 & \textbf{65.3} \\
    \bottomrule
  \end{tabular}
\end{table*}

\subsection{Parsing Humans and Objects}
\label{ssec:exp-humanAndObjects}

Here we evaluate the accuracy of human tracking and object localization on the \UnityHumans datasets.

\myParagraph{Human Nodes}
\Cref{tab:humanObjectsAccuracyLong} shows the average localization error (mismatch between the pelvis
estimated position and the ground truth) for each human on the \UnityHumans datasets.
Each column adds a feature of the proposed model that improves performance.
The first column reports the error of the detections produced by~\cite{Kolotouros19cvpr-shapeRec} (label: ``Single Image'').
The second column reports the error for the case in which we filter out detections
when the human is only partially visible in the camera image, or when the bounding box of the human is too small
($\leq 30$ pixels, label: ``Single Image filtered'').
The third column reports errors with the proposed pose graph model discussed in~\Cref{ssec:humans} (label: ``Pose-Graph track'')
and includes PCM outlier rejection and pose-graph pruning. 
The fourth column reports errors when the mesh feasibility check for data association is enabled (label: ``Mesh Check''),
and the fifth reports errors when the beta-parameter data-association technique is also enabled (label: ``Beta Check'').
 The simulator's humans all have randomized beta parameters within a known range to
better approximate the distribution of real human appearance. 

The Graph-CNN approach~\citep{Kolotouros19cvpr-shapeRec} for SMPL detections
tends to produce incorrect estimates when the human is occluded.
Filtering out these detections improves the localization performance,
but occlusions due to objects in the scene still result in significant errors. 
Adding the mesh-feasibility check decreases error by making data association more effective once detections are registered.
The beta-parameter check also significantly decreases error,
signifying that data association can be effectively done using SMPL body-parameter estimation.

Only the apartment scene did not follow the trend; results are best without any of the proposed techniques. These are outlier results; 
the apartment environment had many specular reflections that could have led to false detections. 

\myParagraph{Object Nodes}
\Cref{tab:humanObjectsAccuracyLong} reports the average localization errors for objects of unknown and known shape detected in the scene. %
In both cases, we compute the localization error as the distance between the estimated and the ground truth centroid of the object 
(for the objects with known shape, we use the centroid of the fitted CAD model).
We use CAD models for objects classified as ``couch'', ``chair'', and ``car'' (which we obtain from Unity's 3D asset store).
In both cases, we can correctly localize the objects, while the availability of a CAD model further boosts accuracy.
\rebuttal{It is worth noting that most of the known objects in our datasets are visualized early in the runs,
when the localization drift is low, making the object localization almost independent from the drift.
Finally, we use ICP to align the 3D groundt-truth mesh with the estimated 3D mesh before evaluating the object localization errors.}

\begin{table*}[htbp]
  \centering
  \caption{Human and object localization errors in meters. A dash (--) indicates that the object is not present in the scene. `\#H' column indicates the number of humans in the scene. `uH1' and `uH2' stand for the \UnityHumans and \UnityHumansTwo datasets respectively.}
  \label{tab:humanObjectsAccuracyLong}
  \begin{tabular}{ccc cccccc cccc}
    \toprule
     & & & \multicolumn{9}{c}{Localization Errors [m]}\\
    \cmidrule(l{2pt}r{2pt}){4-12}
     & & & \multicolumn{5}{c}{Humans} & \multicolumn{4}{c}{Objects}  \\
    \cmidrule(l{2pt}r{2pt}){4-8} \cmidrule(l{2pt}r{2pt}){9-12} 
    Dataset & Scene &  \makecell{\#H} & \makecell{Single \\ Image} & \makecell{Single \\ Image \\ Filtered} & \makecell{Pose \\ Graph \\ Track.} & \makecell{Mesh \\ Check} & \makecell{Beta \\ Check} 
    & \makecell{Un- \\ known \\ Obj. } & \multicolumn{3}{c}{Known Obj.}  \\
    \cmidrule(l{2pt}r{2pt}){10-12} 
     & & & & & & & & & Couch & Chair & Car \\
    \midrule
    \multirow{3}{*}{uH1}       & \multirow{3}{*}{Office}                      & 12 & 2.51 & 1.82 & 1.60 & 1.57 & \textbf{1.52} & 1.31 & 0.20 & 0.20 &  --  \\
                               &                                              & 24 & 2.54 & 2.03 & 1.80 & 1.67 & \textbf{1.50} & 1.70 & 0.35 & 0.35 &  --  \\
                               &                                              & 60 & 2.03 & 1.78 & 1.65 & 1.65 & \textbf{1.63} & 1.52 & 0.38 & 0.38 &  --  \\
    \cmidrule(l{1pt}r{1pt}){1-12}
    \multirow{12}{*}{uH2}      & \multirow{3}{*}{Office}                      & 0  &  --  &  --  &  --  &  --  &  --  & 1.27 & 0.19 & 0.19 &  --  \\
                               &                                              & 6  & 1.87 & 1.21 & 0.86 & 0.82 & \textbf{0.63} & 1.05 & 0.17 & 0.18 &  --  \\
                               &                                              & 12 & 2.00 & 1.43 & 1.16 & 1.05 & \textbf{0.61} & 1.32 & 0.21 & 0.22 &  --  \\
    \cmidrule(l{1pt}r{1pt}){2-12}
                               & \multirow{3}{*}{\makecell{Neigh-\\borhood}}  & 0  &  --  &  --  &  --  &  --  &  --  & 2.89 &  --  &  --  & 2.23 \\
                               &                                              & 24 & 21.3 & 2.02 & 1.06 & 1.03 & \textbf{0.74} & 3.31 &  --  &  --  & 3.29 \\
                               &                                              & 36 & 14.0 & 2.50 & 1.44 & 1.14 & \textbf{0.55} & 3.56 &  --  &  --  & 3.47 \\
    \cmidrule(l{1pt}r{1pt}){2-12}
                               & \multirow{3}{*}{Subway}                      & 0  &  --  &  --  &  --  &  --  &  --  & 3.96 & 2.60 &  --  &  --  \\
                               &                                              & 24 & 8.34 & 6.56 & 5.53 & 5.31 & \textbf{1.92} & 3.76 & 2.70 &  --  &  --  \\
                               &                                              & 36 & 7.61 & 5.80 & 5.20 & 5.12 & \textbf{2.83} & 3.10 & 2.30 &  --  &  --  \\
    \cmidrule(l{1pt}r{1pt}){2-12}
                               & \multirow{3}{*}{\makecell{Apart-\\ment}}     & 0  &  --  &  --  &  --  &  --  &  --  & 0.48 & 0.22 & 0.21 &  --  \\
                               &                                              & 1  & \textbf{4.32} & 4.79 & 5.38 & 5.64 & 6.43 & 0.43 & 0.21 & 0.21 &  --  \\
                               &                                              & 2  & 2.83 & \textbf{2.52} & 2.66 & 2.69 & 3.79 & 0.45 & 0.21 & 0.21 &  --  \\
    \bottomrule
  \end{tabular}
\end{table*}{}

\subsection{Parsing Places and Rooms}
\label{ssec:exp-roomParsing}

We also compute the average precision and recall for the classification of places into rooms.
The ground-truth labels are obtained by manually segmenting the places.

\begin{figure*}[htbp]
  \centering
  \includegraphics[trim={0cm 0cm 0cm 0cm}, clip, width=\textwidth]{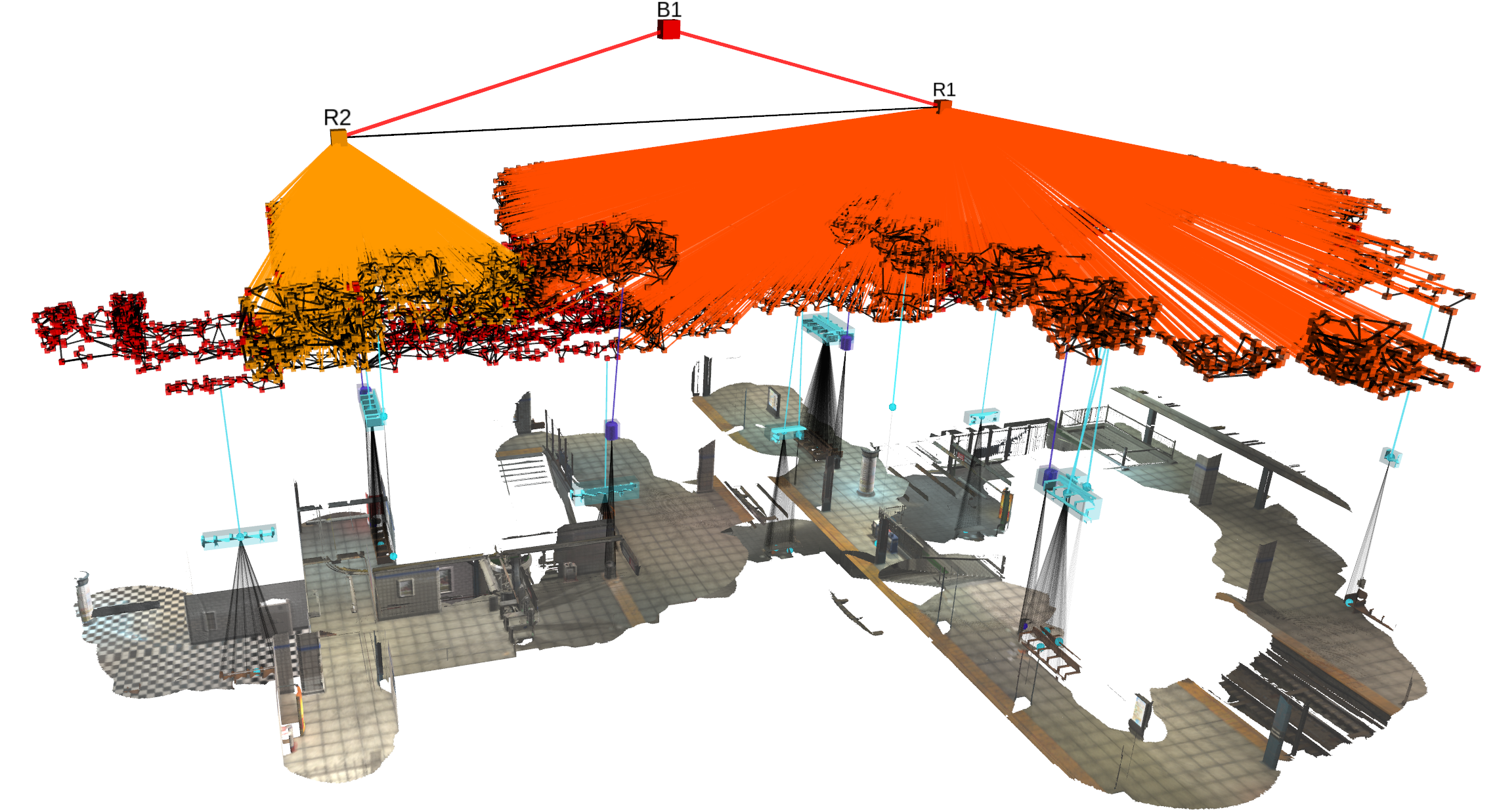}
  \caption{\DSG for the subway dataset in \UnityHumansTwo.
   The room `R2' belongs to the subway's lobby (mezzanine), while the room `R1' corresponds to the subway's platform.
   The fare collection gates physically separate these two rooms. Segmented benches are shown in cyan, as well as bins in purple.}
  \label{fig:subway_dsg}
\end{figure*}

For the `Office' (\Cref{fig:DSG}) and `Subway' (\Cref{fig:subway_dsg}) scenes in \UnityHumansTwo, we obtain an average precision of $99\%$ and $87\%$, respectively, 
and an average recall of $99\%$ and $92\%$, respectively.
Similarly, for the real-life \WhiteOwl (left \DSG in \Cref{fig:white_owl_dsg}) and \School (\Cref{fig:sunday_school_dsg}) scenes,
 we achieve a precision of $93\%$ and $91\%$, respectively, and an average recall of $94\%$ and $90\%$, respectively.
In fact, all the rooms in the `Office', `Subway', \WhiteOwl, and \School scenes are correctly detected and the places are precisely labelled.
Incorrect classifications of places typically occur near doors, where room misclassification is inconsequential.

\begin{figure*}[htbp]
  \centering
  \includegraphics[trim={0cm 0cm 0cm 0cm}, clip, width=0.9\textwidth]{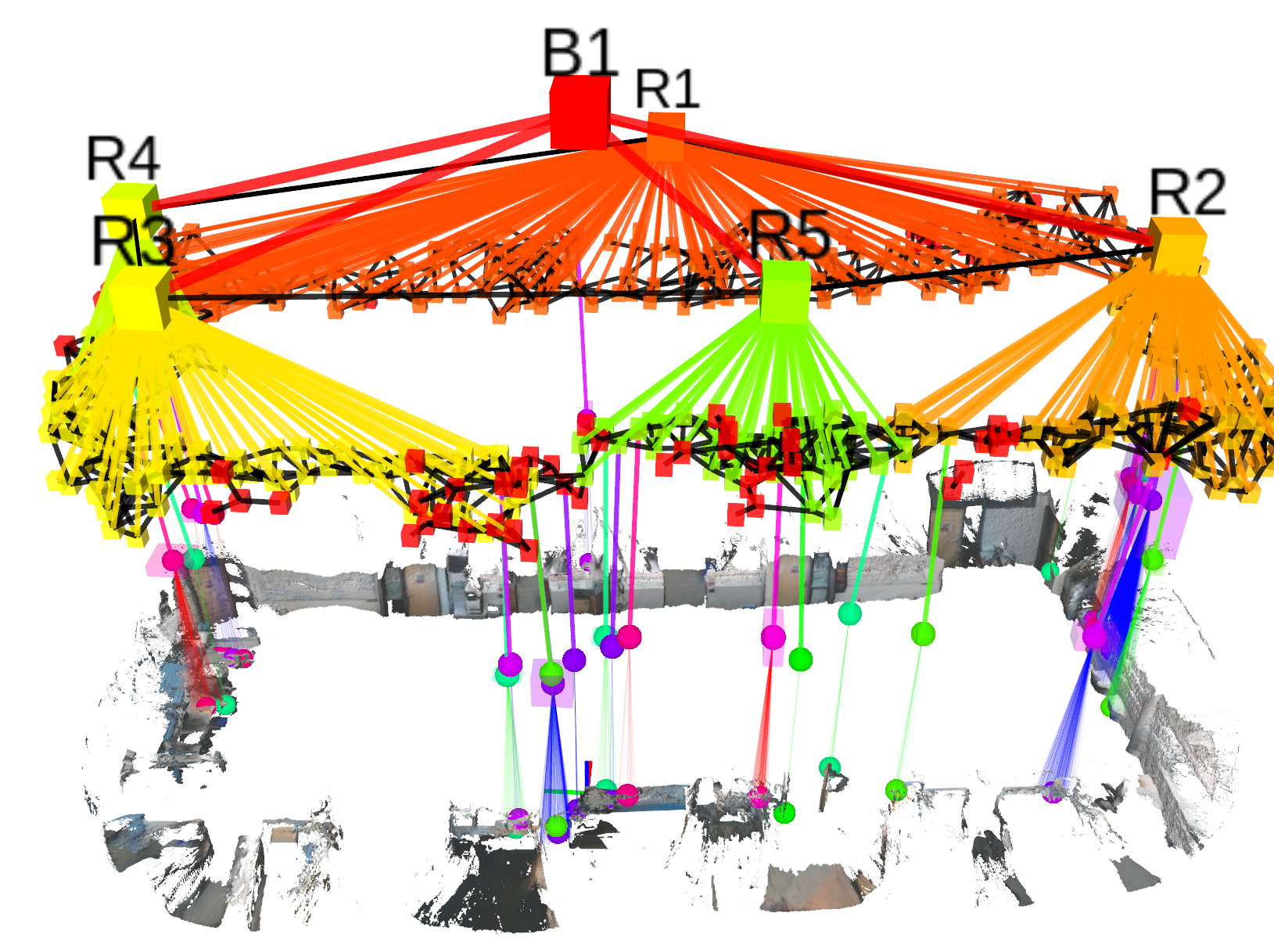}
  \caption{Real-life \DSG for the \BldgThirtyOne dataset using a custom-made sensing rig with \RealSenseLong sensors (\Cref{ssec:datasets}).
   The reconstructed 3D mesh by \Kimera is globally consistent, with the loop around the office being accurately closed.
   While the \DSG reconstructed by \Kimera is coherent, the corridor parallel to `R1' is over-segmented into three rooms (`R2', `R5', `R3'), see \Cref{ssec:real_life_experiments}.}
  \label{fig:aero_astro_dsg}
\end{figure*}

Nevertheless, our approach has difficulties dealing with either complex architectures, such as the `Apartment' in \UnityHumansTwo (right \DSG in \Cref{fig:white_owl_dsg}), and
largely incomplete scenes such as the \BldgThirtyOne dataset (\Cref{fig:aero_astro_dsg}).
In particular, for the `Apartment' scene, the presence of exposed ceiling beams distorts the 2D ESDF field in a way that the living room is instead over-segmented as three separated rooms:
`R1', `R3', and `R6' should be one room node for the \DSG on the right side of \Cref{fig:white_owl_dsg}.
The precision and recall for the place segmentation of the `Apartment' scene is of $68\%$ and $61\%$ respectively.
For the \BldgThirtyOne scene (\Cref{fig:aero_astro_dsg}), the parallel corridor to `R1' is also over-segmented into `R3', `R5', and `R2'.
In this case, the observed free-space in the corridor narrows significantly between `R2' and `R5', and between `R5' and `R3', as can be seen by the density of nodes in the topological graph (\Cref{fig:aero_astro_dsg}).
The precision and recall for the place segmentation of the \BldgThirtyOne scene is of $71\%$ and $67\%$ respectively.
Note that with a careful tuning of the parameters for room detection, we can avoid such over-segmentations (\Cref{sec:buildingParser}).
We also discuss the influence of room over-segmentation when performing path-planning queries in \Cref{sec:hierarchical-path-planning}.

\begin{figure*}[htbp]
  \centering
  \includegraphics[trim={0cm 0cm 0cm 0cm}, clip, width=\textwidth]{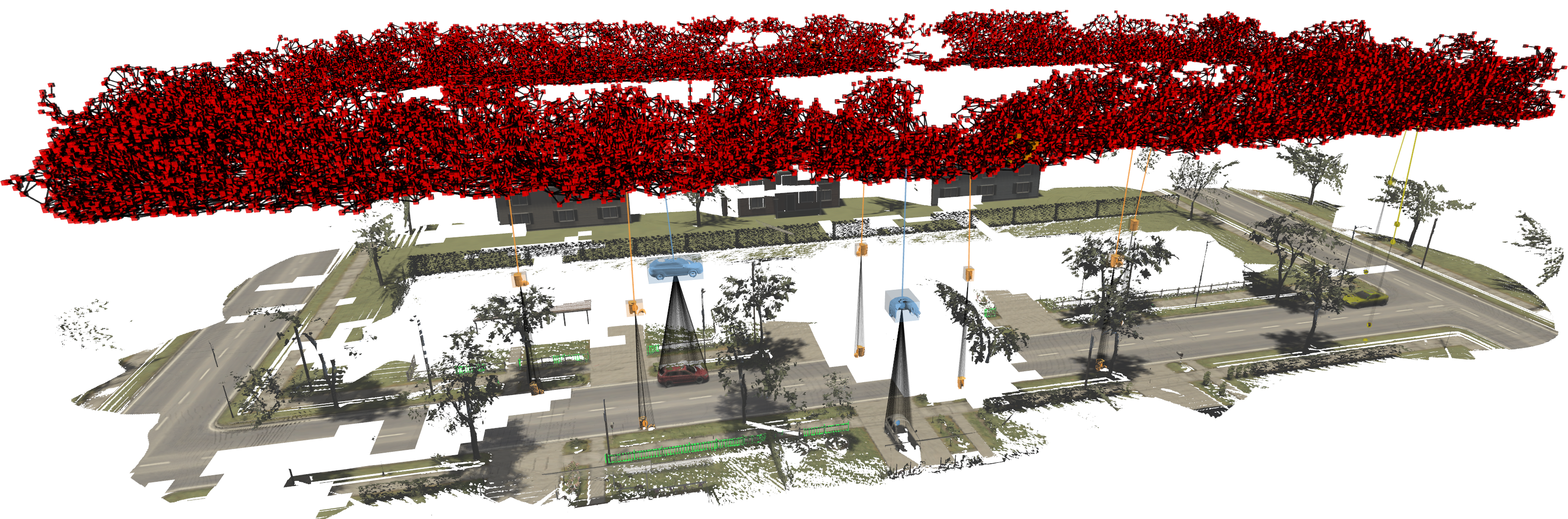}
  \caption{\DSG for the neighborhood dataset in \UnityHumansTwo.
   Despite not having multiple rooms, a \DSG can also be built from outdoor scenes.
    Segmented cars are shown in blue, as well as bins in orange.}
  \label{fig:neighborhood_dsg}
\end{figure*}

Finally, we note that \DSGs are general enough to handle outdoor scenes,
such as the `Neighborhood' scene (\Cref{fig:neighborhood_dsg}),
while in this case some of the layers (rooms, buildings) are skipped.
    
\subsection{Timing}
\label{ssec:timing_performance}

\begin{figure}[htbp]
  \centering   
  \includegraphics[trim={0cm 0cm 0cm 0cm}, clip, width=\columnwidth]{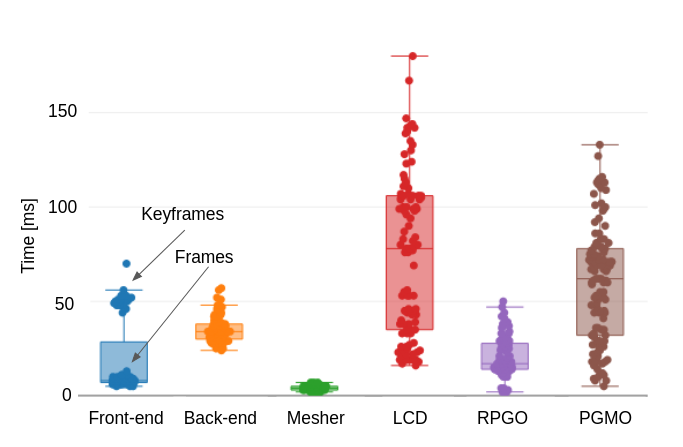} 
  \vspace{-2mm}
  \caption{Runtime breakdown for \KimeraVIO, \KimeraMesher, loop-closure detection (LCD), \KimeraRPGO, and \KimeraPGMO. 
  Note that the timing for \KimeraRPGO and \KimeraPGMO will increase with the size of the pose graph and the mesh. 
  The timing here is collected on the \Euroc V1\_01 dataset. 
  The complete run of the dataset consists of $728$ poses and $71$ loop closures, 
  and the final mesh has $324474$ vertices.} 
  \vspace{-2mm}
  \label{fig:timing}
\end{figure}

\Cref{fig:timing} reports the timing performance of \KimeraCore's modules.
The IMU front-end requires around $40\mu\text{s}$ for preintegration,
 hence it can generate state estimates at IMU rate ($>200$~Hz).
The vision front-end module shows a bi-modal distribution since, for every frame, we just perform feature tracking 
(which takes an average of $7.5$~ms), while, 
at keyframe rate, we perform feature detection, stereo matching, and geometric verification,
which, combined, take an average of $51$~ms.
\KimeraMesher is capable of generating per-frame 3D meshes in less than $7$~ms, while building the multi-frame mesh takes $15$~ms on average.
The back-end solves the factor-graph optimization in less than $60$~ms.
Loop-closure detection (LCD) took up to $180$~ms to look for loop closure candidates and perform geometric verification and compute the relative transform.
\KimeraRPGO, \KimeraPGMO and \KimeraSemantics run on slower threads since their outputs are not required for time-critical 
actions (\eg control, obstacle avoidance). 
\KimeraRPGO took up to $50$~ms in our experiments on \Euroc for a pose graph with $734$ edges and $728$ nodes,
 but in general its runtime depends on the size of the pose graph.
Similarly, \KimeraPGMO took up to $140$~ms in our experiments on \Euroc for a pose graph with $14914$ edges and 1759 nodes ($728$ pose nodes and $1031$ mesh nodes), but in general its runtime depends on the size of the pose graph, the size of the mesh, and the resolution of the simplified mesh.
Finally, \KimeraSemantics (not reported in figure for  clarity) takes an average of $\globalMeshLatency$ to update the global metric-semantic mesh at each keyframe.

\KimeraDSG's modules run sequentially once the 3D metric-semantic mesh is built by \KimeraCore.
\KimeraObjects segments the object instances from the 3D metric-semantic mesh in minutes, depending on the scale of the scene 
(from $3$~min for the \WhiteOwl scene of $\sim{100}\text{m}^2$ to $12$~min for the largest `Subway' scene of $\sim 3000\text{m}^2$), 
where the instance segmentation is done sequentially for each object class at a time (which can be parallelized).
For the known objects, \KimeraObjects fits a CAD model using \TEASERpp which runs in milliseconds 
(more details about the timing performance of \TEASERpp are provided in \citep{Yang20tro-teaser}).
\KimeraHumans first detects humans using \MeshName \citep{Kolotouros19cvpr-shapeRec} which runs for a single image in $\sim 33$~ms on a Nvidia RTX 2080 Ti GPU,
 and the tracking is performed on the CPU in milliseconds ($\sim 10$~ms).
\KimeraBuildingParser first builds an ESDF out of the TSDF using Voxblox~\citep{Oleynikova2017iros-voxblox},
 which may take several minutes depending on the scale of the scene ($\sim 10$~min for large scenes such as the `Subway').
Then, the topological map is built using~\citep{Oleynikova18iros-topoMap},
 which also takes several minutes depending on the scale of the scene ($\sim 10$~min for `Subway').
Finally, the detection of rooms, segmentation of places into rooms, and finding the connectivity between rooms,
 takes a few minutes depending as well on the size of the scene ($\sim 2$~min for the `Subway' scene).
Hence, to build a \DSG for a large scale scene such as the `Subway' scene, \KimeraDSG may take approximately $\sim 30$~min,
with the computation of the ESDF,
the topological map, and the object instance segmentation being the most time-consuming operations.

\subsection{Timing on NVIDIA TX2 Embedded Computer}
\label{ssec:tx2_timing_performance}

We also assessed the performance capabilities of Kimera on a common low-SWaP
processor for many robotic applications, the Nvidia Jetson TX2.  For all
benchmarking conducted with the TX2, the \texttt{MAXN} performance mode was
used via \texttt{nvpmodel}. We limited our analysis to \KimeraCore.
\KimeraVIO~was able to achieve real time performance when run
against \Euroc~with default settings, but we also understand that there may be
applications that are more sensitive to latency or processing time. We
therefore analyzed the impact of various parameter settings on the runtime of
\KimeraVIO.

\begin{figure}[htbp]
  \centering
  \begin{minipage}{.49\textwidth}
  \includegraphics[trim={0cm 0cm 0cm 0cm}, clip, width=1.0\columnwidth]{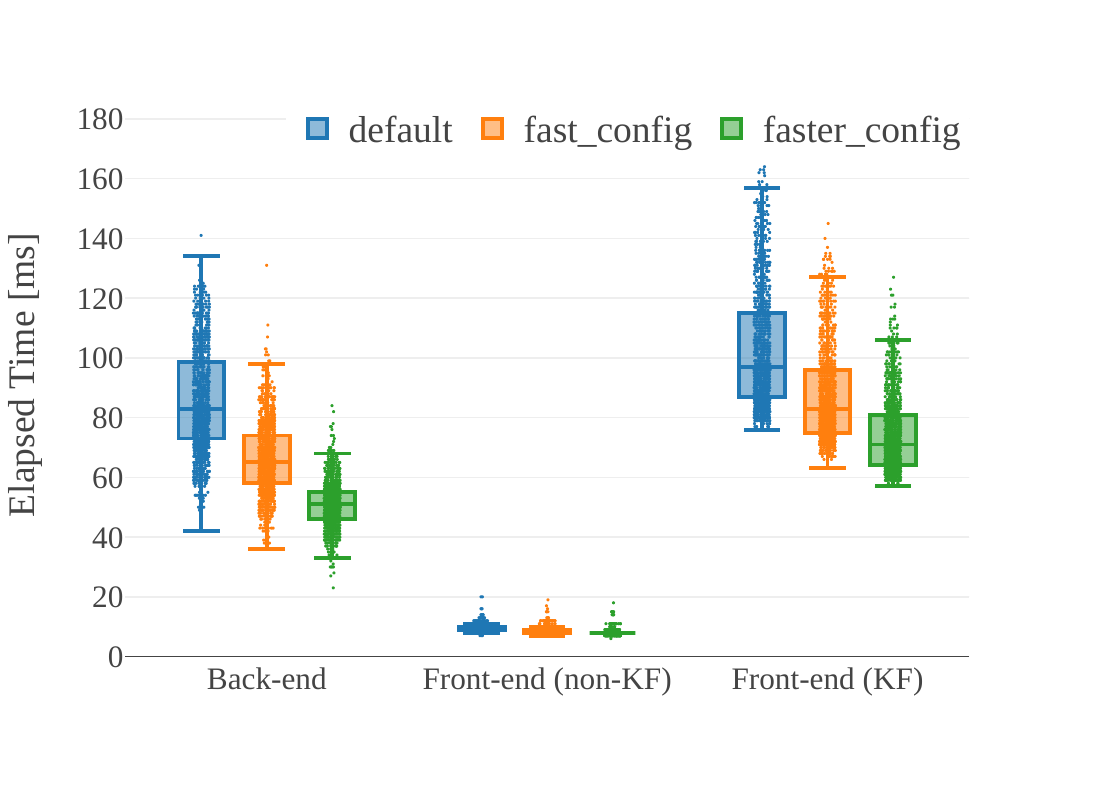}
  (a) TX2 Runtime breakdown for \KimeraVIO's back-end and front-end.
   The front-end's timing distribution is bi-modal: its runtime is longer on every keyframe (KF) than its per-frame runtime (non-KF).
  \end{minipage}
  \hspace{1mm}
  \begin{minipage}{.49\textwidth}
  \includegraphics[trim={0cm 0cm 0cm 0cm}, clip, width=1.0\columnwidth]{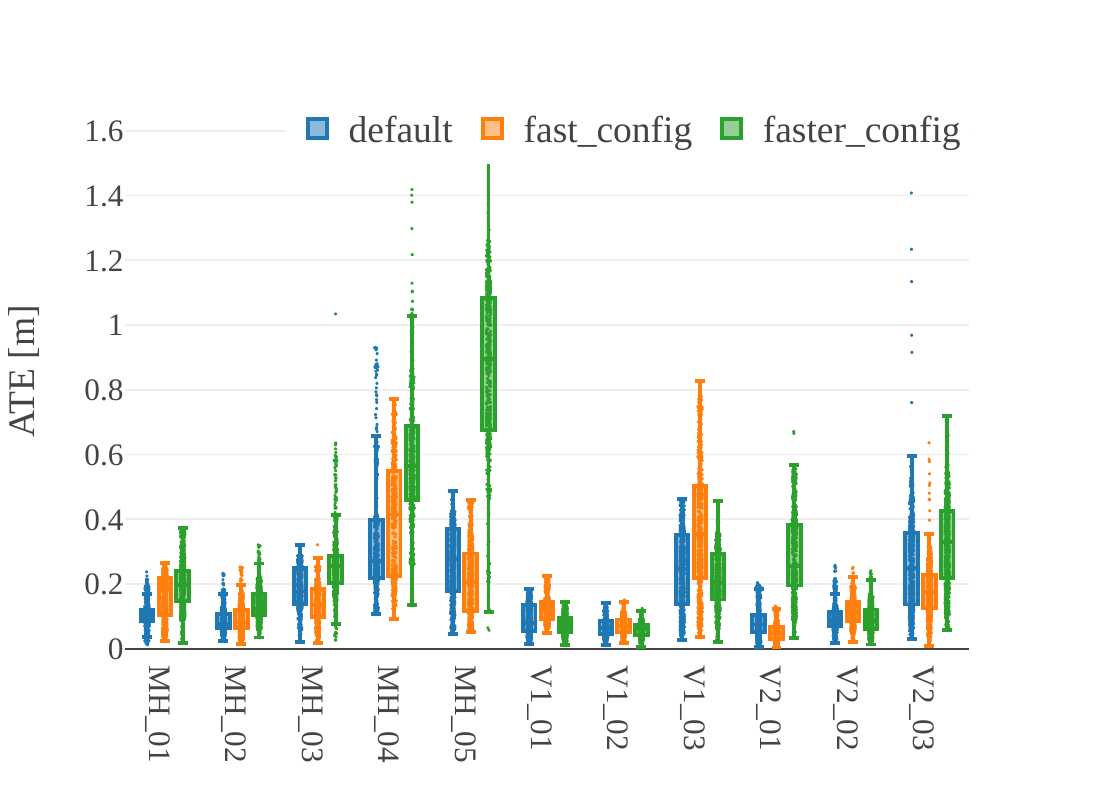}
  (b) TX2 ATE breakdown for \KimeraVIO
  \end{minipage}
    \caption{ Runtime and ATE for \KimeraVIO~for different parameter settings across
        \Euroc~evaluated on the TX2. Labels KF and non-KF denote timing for
        keyframe insertions in the frontend (involving feature detection and
        geometric verification through RANSAC) and normal frame insertions
        (involving feature tracking) respectively.  The default, \textit{fast},
        and \textit{faster} configurations extract a maximum of 300, 250, and
        200 features respectively and have a horizon size of 6, 5, and 4.5
        seconds respectively}
    \label{fig:tx2_timing_and_error}
\end{figure}

Two candidate parameters were identified: (i)~the maximum number of features
tracked (\textit{maxFeaturesPerFrame}), and (ii)~the time horizon for smoothing
(\textit{horizon}). 
Based on this analysis, we provide two additional sets of parameters in the open-source repository: a
\textit{fast} (250 features maximum and a horizon size of 5 seconds) and a
\textit{faster} (200 features maximum and a horizon size of 4.5 seconds)
parameter configuration. The effect of these settings on the processing time of
\KimeraVIO~is shown in~\Cref{fig:tx2_timing_and_error}. For all configurations,
the frontend portion of the module took 10ms on average on non-keyframes. The
\textit{fast} and \textit{faster} configurations takes about 85\% and 75\% of
the processing time for keyframes, and about 65\% and 50\% of the processing
time for the back-end respectively as compared to the default settings.  Note
that on more challenging portions of \Euroc, the accuracy of
\KimeraVIO~degrades under the \textit{fast} and \textit{faster} configurations,
which may not be desired in all instances.  This is most noticeable in the
increase in APE for the difficult machine hall sequences of \Euroc, shown
in~\Cref{fig:tx2_timing_and_error}.

In addition, we characterized the timing of \KimeraSemantics~utilizing the
\Euroc~dataset again. With the default settings for \KimeraVIO~and for
\KimeraSemantics, most notably a voxel size of $0.1$ meters, we were able to
achieve suitable performance for real-time operation. Updates to
\KimeraSemantics~took $65.8$ milliseconds on average across \Euroc. Note that
\Euroc~data does not include semantic labels, so the voxel size may still need to be
tuned to maintain real time performance in other cases.

\subsection{Real-Life Experiments}
\label{ssec:real_life_experiments}

In this section, we qualitatively evaluate \Kimera's ability to generate \DSGs from real-life datasets,
collected using either our 
custom-made sensing rig (\Cref{fig:kimera_rig}) for the \School and \BldgThirtyOne datasets,
 or Microsoft's Azure Kinect camera for the \WhiteOwl dataset, as explained in \Cref{ssec:datasets}.

\subsubsection{\School \& \BldgThirtyOne.}
\label{ssec:experiments-realSense}

\begin{figure}[htbp]
  \centering
  \includegraphics[trim={2cm 0cm 1cm 0cm}, clip, width=0.9\columnwidth]{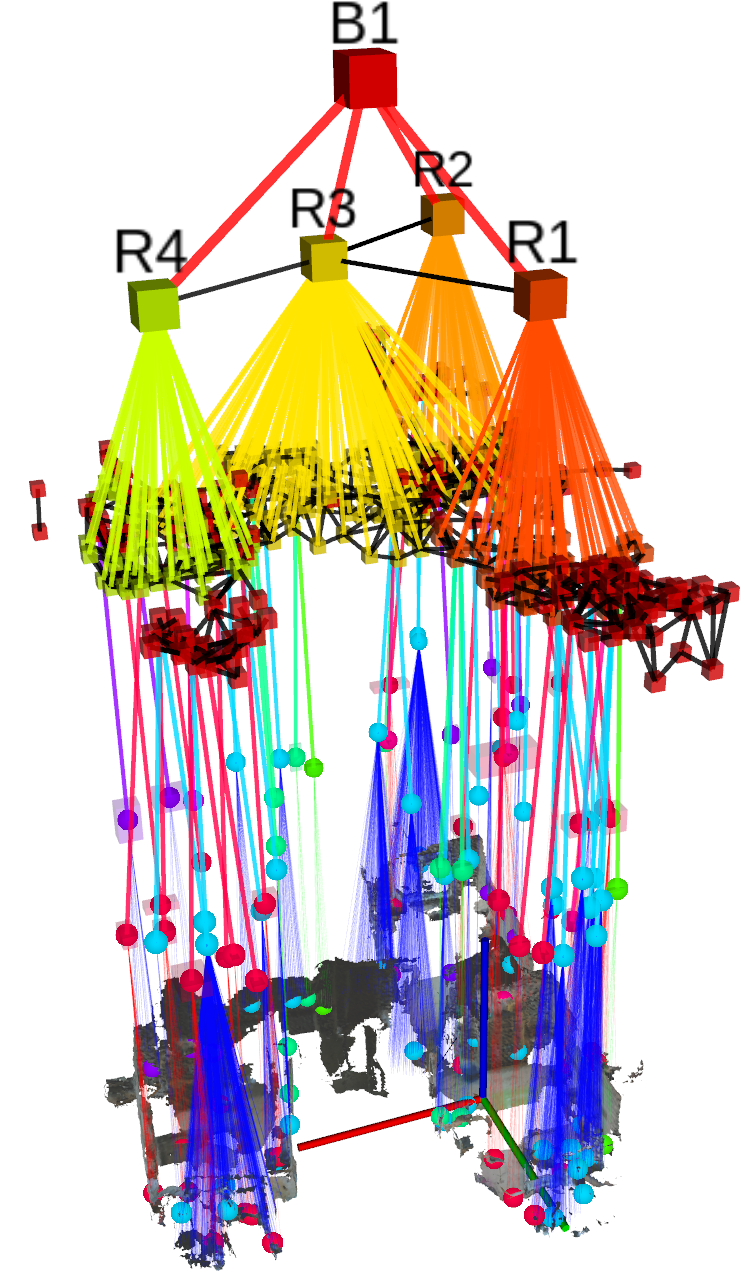}
  \includegraphics[trim={0cm 0cm 0cm 0cm}, clip, width=0.9\columnwidth]{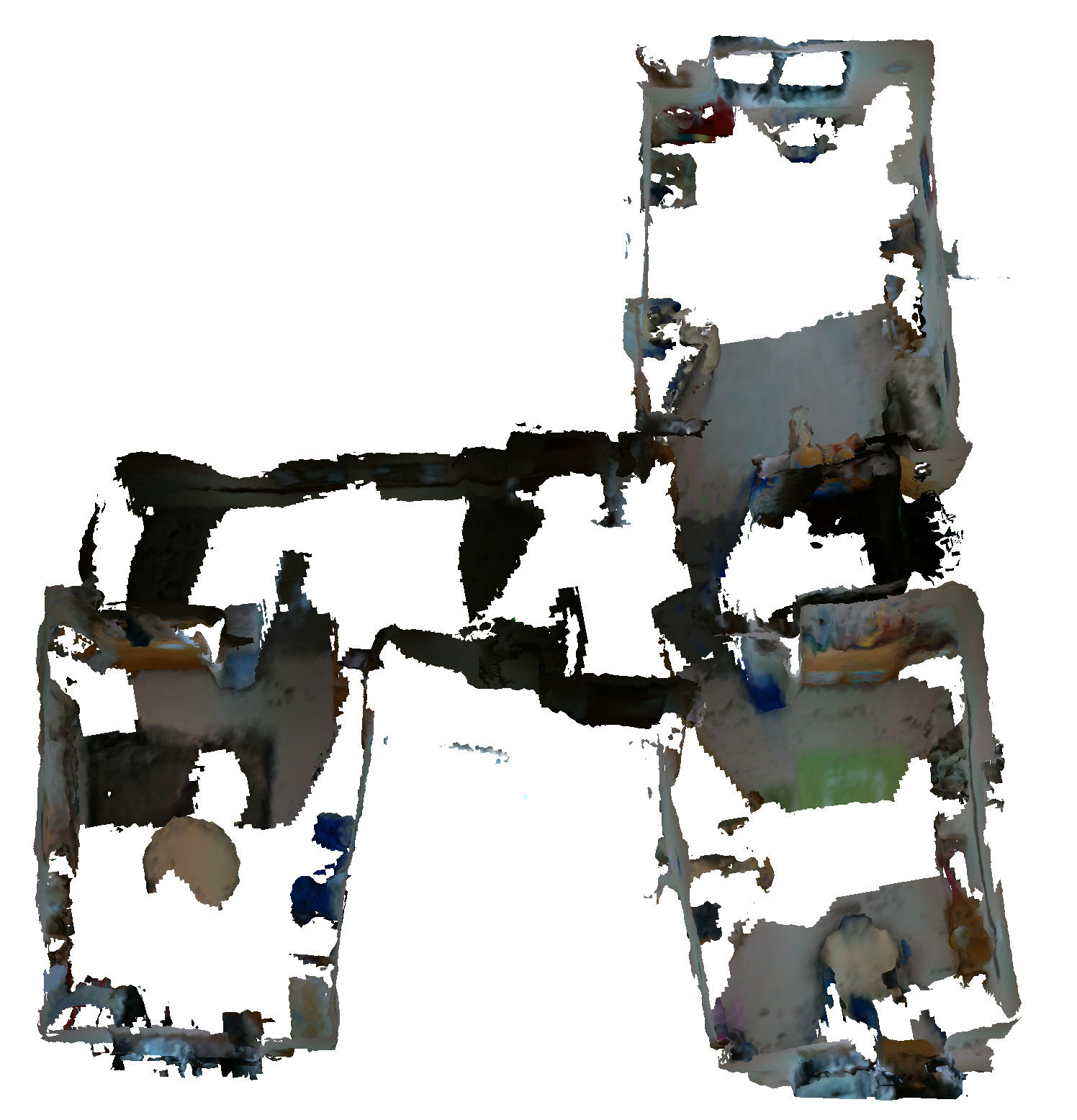}
  \caption{(Top) Real-life \DSG for the \School dataset using a custom-made sensing rig with \RealSenseLong sensors (\Cref{ssec:datasets}). (Bottom) the globally consistent 3D mesh of the scene reconstructed by \Kimera.
  Despite the noisy and incomplete reconstruction, a \DSG correctly abstracts the scene into three rooms and a corridor (`R3').}
  \label{fig:sunday_school_dsg}
\end{figure}

\Cref{fig:sunday_school_dsg} shows the reconstructed \DSG by \Kimera of the \School dataset, together with the reconstructed 3D mesh.

We notice that the reconstructed 3D mesh by \KimeraCore is globally consistent,
despite being noisy and incomplete due to \RealSenseLong's depth stream quality (see \Cref{fig:pcls}).
In particular, \KimeraPGMO is capable of leveraging loop closures to simultaneously deform the mesh and optimize the trajectory of the sensor.

Despite the noise and incomplete 3D mesh, \KimeraDSG is able to build a meaningful \DSG of the scene.
\KimeraObjects correctly detects most of the objects and approximates a conservative bounding box.
Nevertheless, some spurious detections are present.
For example, the green nodes are supposedly refrigerators in \Cref{fig:sunday_school_dsg}, but there are no refrigerators in the \School dataset.
These spurious detections can be easily removed by either fine-tuning the smallest object size valid for a detection,
or by re-training Mask-RCNN with the objects in this scene. It is also possible to stop Mask-RCNN from detecting certain object classes that are not present in the scene.

\Cref{fig:sunday_school_dsg} also shows that \Kimera correctly reconstructs the place layer, and accurately segments the places into rooms.
The room layer accurately represents the layout of the scene, with a corridor (`R3') and three rooms (`R1', `R2', `R4').
The edges between the rooms of the \DSG correctly represent the traversability between rooms.
The fact that the mesh is noisy and incomplete does not seem to negatively affect the higher levels of abstraction of the \DSG.

\Cref{fig:aero_astro_dsg} shows the reconstructed \DSG by \Kimera of the \BldgThirtyOne dataset.
Similarly to the \School dataset, \KimeraCore is able to reconstruct a consistent 3D mesh,
 which remains nonetheless noisy and incomplete since we used the same sensor.
While the objects and places layers are correctly estimated, \KimeraBuildingParser over-segments the rooms.
In particular, \KimeraBuildingParser over-segments a corridor into three rooms (`R2', `R3', `R5').
This is because, when traversing the corridor, the sensor was held too close to the wall, thereby limiting the observed free-space.
Consequently, the ESDF and the places layer narrow significantly at the locations between rooms `R2' and `R5', and between rooms `R5' and `R3'.
This narrowing is misinterpreted by \Kimera as a separation between rooms, which leads to the over-segmentation of the corridor.

\subsubsection{\WhiteOwl.}
\label{ssec:experiments-azureKinect}

\begin{figure*}[htbp]
  \centering
  \begin{minipage}{.49\textwidth}
  \includegraphics[trim={4cm 3cm 4cm 2cm}, clip, width=0.9\columnwidth]{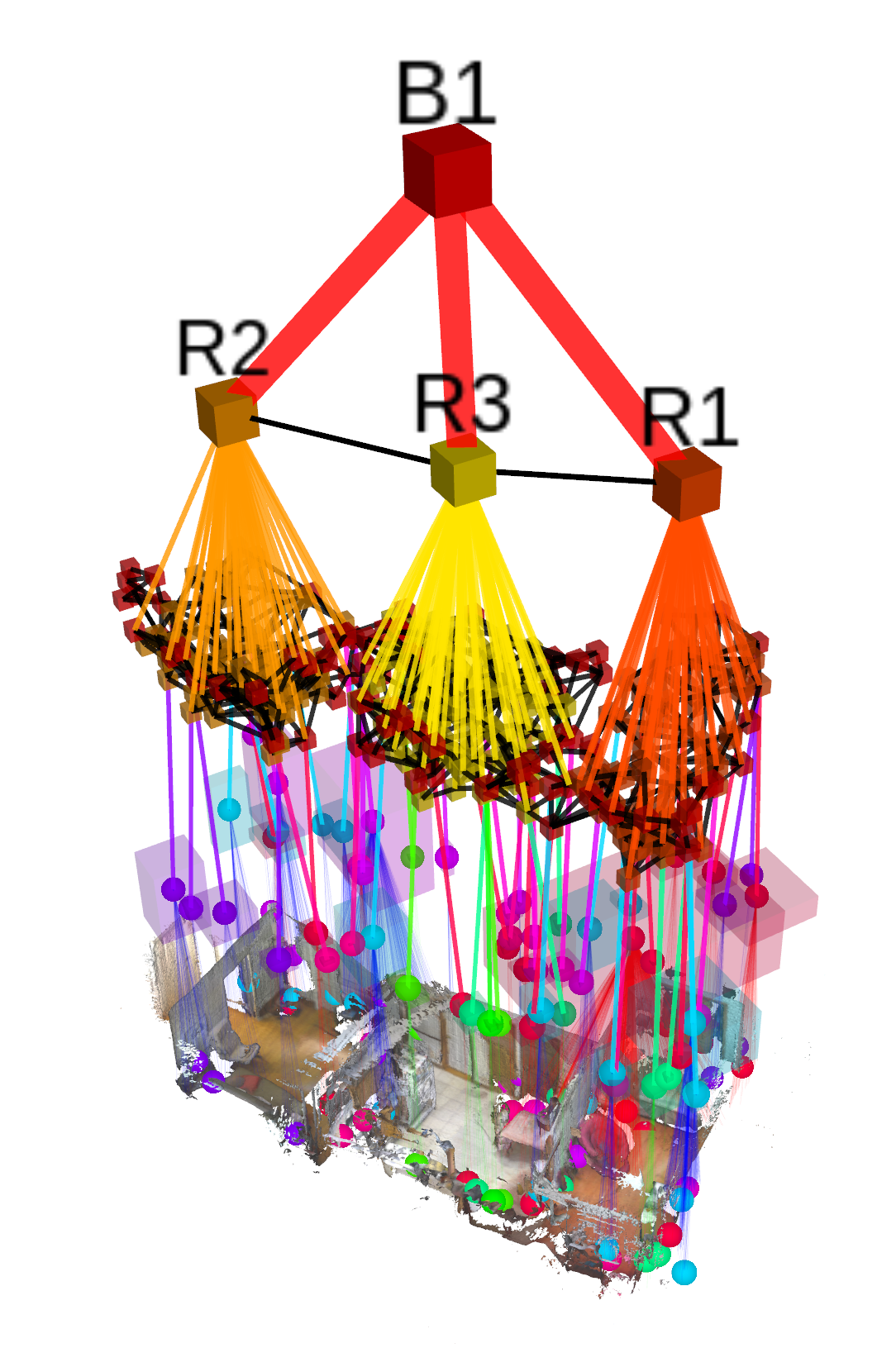}
  \includegraphics[trim={0cm 0cm 0cm 0cm}, clip, width=0.9\columnwidth]{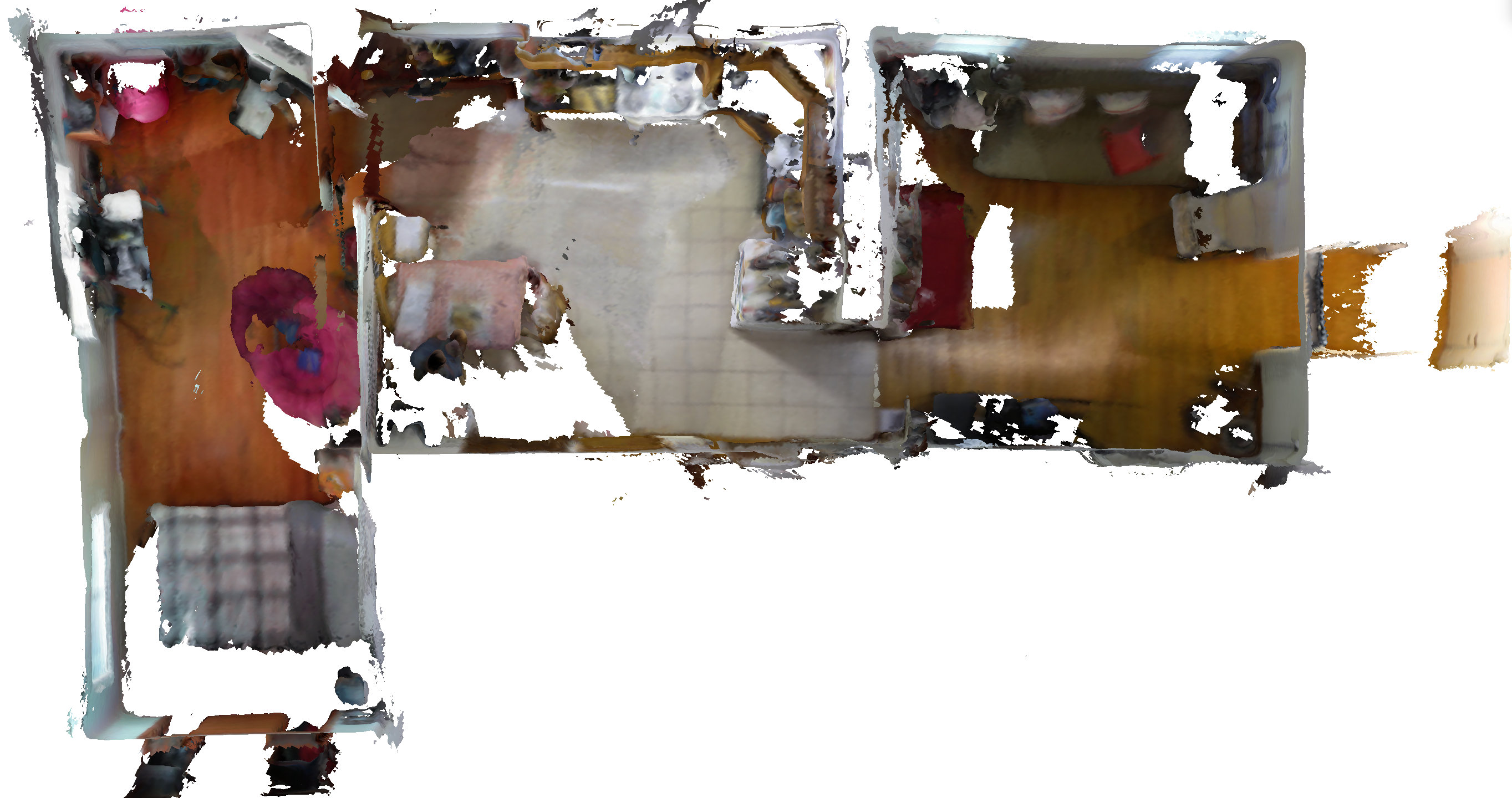}
  \end{minipage}
  \hspace{1mm}
  \begin{minipage}{.49\textwidth}
  \includegraphics[trim={0cm 0cm 0cm 0cm}, clip, width=\columnwidth]{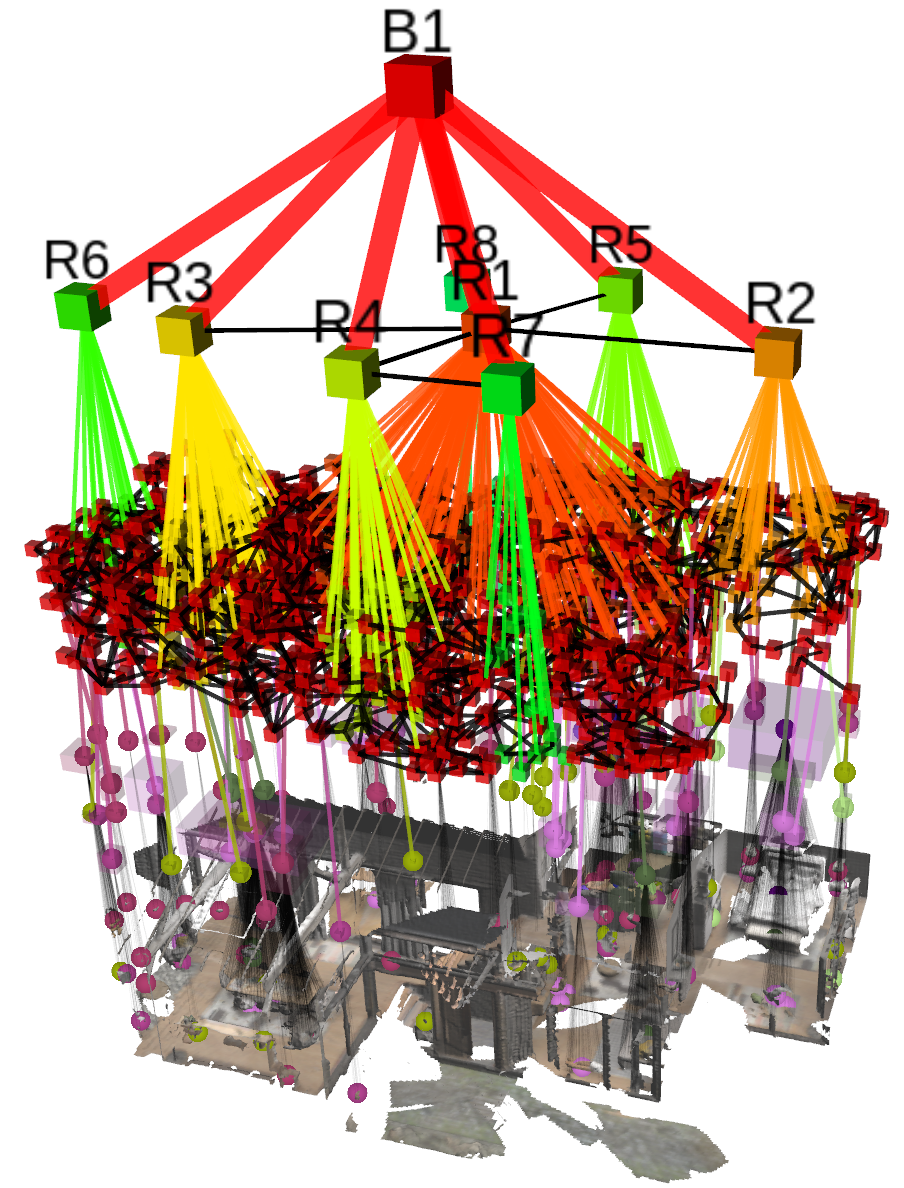}
  \includegraphics[trim={0cm 0cm 0cm 0cm}, clip, width=\columnwidth]{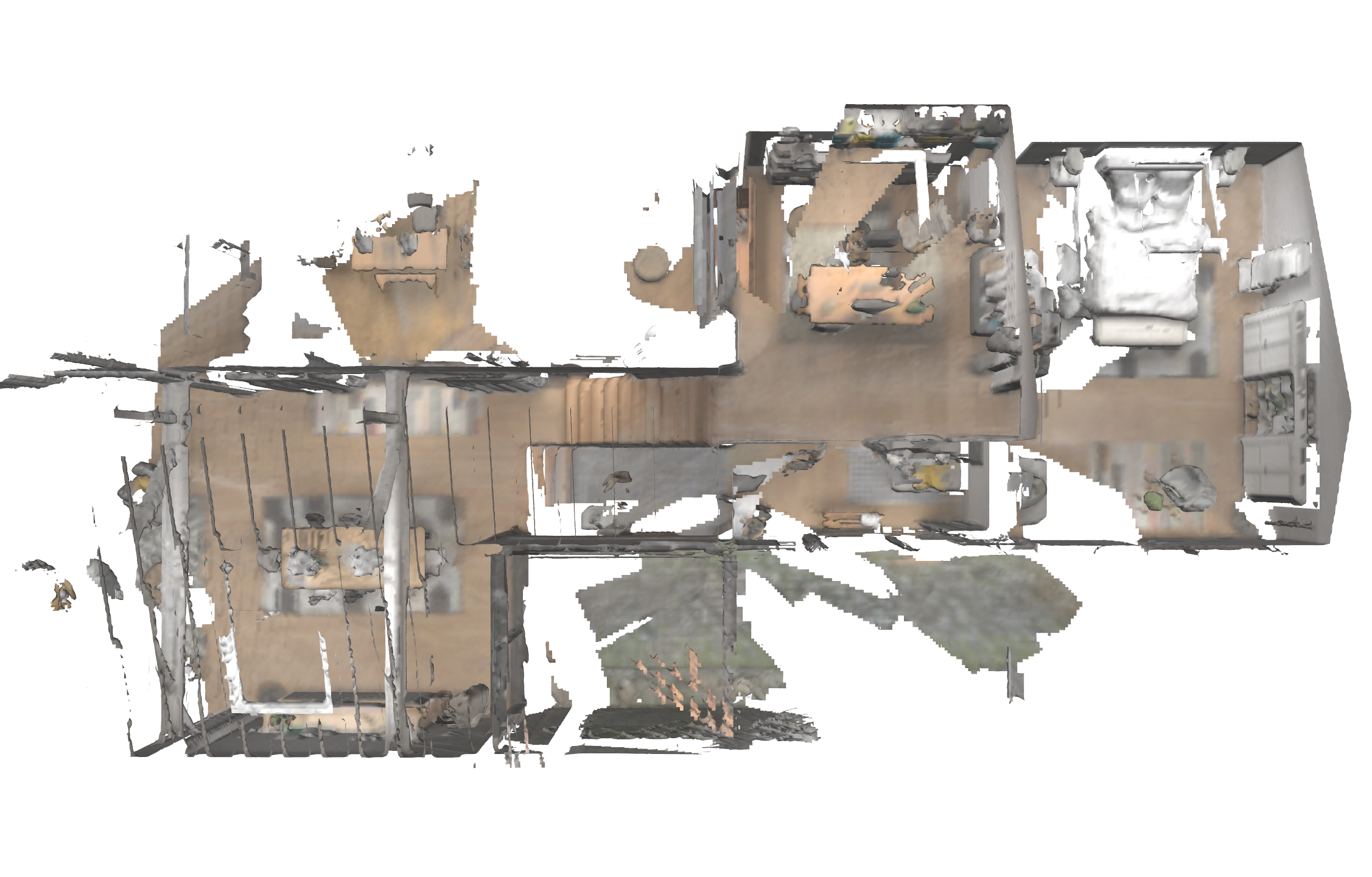}
  \end{minipage}

  \caption{
    (Left) \DSG of the \WhiteOwl apartment (top left) from collected data using Microsoft's \AzureKinect sensor, and a top-down view of the reconstructed 3D mesh (bottom left).
    (Right) \DSG of the simulated \UnityHumansTwo `Apartment' scene (top right), and a top-down view of the reconstructed 3D mesh (bottom right).
    Note that despite the data being generated from real-life (left) and simulated (right) sensors, \Kimera is capable of reconstructing a globally consistent 3D mesh and a coherent \DSG
    of comparable quality, using the same set of parameters.}
  \label{fig:white_owl_dsg}
\end{figure*}

To further assess the performance of \Kimera to build a \DSG,
 we use a high-quality depth camera, Microsoft's \AzureKinect, to reconstruct an accurate and complete 3D mesh of the scene.
\Cref{fig:white_owl_dsg} shows the \WhiteOwl scene's \DSG (left) next to the simulated `Apartment' scene's \DSG (right), both reconstructed by \Kimera.
It is remarkable that, \rebuttal{independently of the fact that one is a simulated dataset} and \rebuttal{the other is a} real-life \rebuttal{dataset},
\Kimera is capable of reconstructing a globally consistent 3D mesh as well as a coherent \DSG for both scenes, \rebuttal{while using} the same set of parameters.

\rebuttal{Furtermore, both reconstructions are qualitatively similar,
showing that our simulations are realistic, as can also be seen in \Cref{fig:datasets_overview},
and that, given a sufficiently accurate depth sensor, Kimera can achieve accurate reconstructions when using real data.
}

On the \WhiteOwl dataset, we still observe nonetheless that there are some spurious object detections (\eg green nodes). 
The places and rooms layer of the \DSG remain accurate, and correctly represent the scene and the connectivity between layers.

\section{\rebuttal{Motivating Examples}}
\label{sec:examples}

We highlight the actionable nature of a 3D Dynamic Scene Graph 
by providing examples of queries it enables.

\myParagraph{Obstacle Avoidance and Planning}  %
Agents, objects, and {rooms in our \DSG have a bounding box attribute.}
Moreover, the hierarchical nature of the \DSG ensures that bounding boxes at higher layers contain bounding boxes at lower layers 
(\eg the bounding box of a room contains the objects in that room). 
This forms a \emph{Bounding Volume Hierarchy} (BVH)~\citep{Larsson06cg-bvh}, which is extensively used for collision checking in computer graphics. BVHs provide readily available opportunities to speed up obstacle avoidance and motion planning queries
where collision checking is often used as a primitive~\citep{Karaman11ijrr-planning}.

\myParagraph{Human-Robot Interaction} 
As already explored in~\citep{Armeni19iccv-3DsceneGraphs,Kim19tc-3DsceneGraphs}, a scene graph 
 can support user-oriented tasks, such as interactive visualization and \emph{Question Answering}. 
Our Dynamic Scene Graph extends the reach of~\citep{Armeni19iccv-3DsceneGraphs,Kim19tc-3DsceneGraphs} by 
(i) allowing visualization of human trajectories and dense poses (see visualization in the video attachment), 
and (ii) enabling more complex and time-aware queries such as 
``where was this person at time $t$?'', or 
``which object did this person pick in Room A?''. 
Furthermore, \DSGs provide a framework to model plausible interactions between agents and scenes \citep{Zhang19arxiv, Hassan19iccv, Pirk17tog, Monszpart19tog-imapper}.
We believe \DSGs also complement the work on natural language grounding~\citep{Kollar17arxiv-GGG},
 where one of the main concerns is to reason over the variability of
human instructions.

\myParagraph{Long-term Autonomy} 
\DSGs provide a natural way to ``forget'' or retain information in long-term autonomy. 
By construction, higher layers in the \DSG are more compact and abstract representations of the 
scene. Hence, the robot can ``forget'' portions of the environment that are not frequently observed by simply pruning the 
corresponding branch of the \DSG. For instance, to forget a room in \FigFrontCover, we 
only need to prune the corresponding node and the connected nodes at lower layers (places, objects, etc.).
 More importantly, the robot can selectively decide which information to retain: 
 for instance, it can keep all the objects (which are typically fairly cheap to store), but can selectively forget the 
 mesh model, which can be more cumbersome to store in large environments.
 Finally, \DSGs inherit memory advantages afforded by standard scene graphs: if the robot detects $N$ instances of a known object (\eg a chair), 
 it can simply store a \emph{single} CAD model and cross-reference it in $N$ nodes of the scene graph; this simple observation 
 enables further data compression. %

\myParagraph{Prediction} 
The combination of a dense metric-semantic mesh model and a rich description of the agents allows  
performing short-term predictions of the scene dynamics and answering queries about possible future outcomes. 
For instance, one can feed the mesh model to a physics simulator and roll out potential high-level actions of the human agents.
 \optional{In general, we hope \DSG can help bridge the gap between robotics, computer vision, and graphics,  
 and enable the use of advanced graphics engines to support  decision-making tasks.}

    \section{\rebuttal{Applications}}
\label{sec:applications}

\rebuttal{In the remaining, we present two particular applications that we developed, and which we evaluate in \Cref{sec:exp-semantic_path_planning}.}

\subsection{Hierarchical Path Planning} 
\label{sec:hierarchical-path-planning}

The multiple levels of abstraction afforded by a \DSG 
have the potential to enable hierarchical and multi-resolution planning approaches~\citep{Schleich19rss-vinLevelsAbstraction,Larsson19arxiv-qSearch},
 where a robot can plan at different levels of abstraction to save computational resources.

In fact, querying paths from one point to another is computationally expensive when using volumetric maps even in small scenes.
With \DSGs, we can instead compute a path in a hierarchical fashion, which accelerates the path planning query by several orders of magnitude.

To showcase this capability, we first compute a feasible shortest path (using A$^*$) at the level of buildings.
Given the building nodes to traverse, we extract their room layer, and re-plan at the level of rooms.
Similarly, given the room nodes to traverse, we further extract the relevant graph of places, and re-plan again at this level.
Finally, we extract a smooth collision-free path using the open-source work of~\citep{Oleynikova16iros-continuous,Oleynikova18iros-topoMap}.

While querying a path at the level of the volumetric ESDF takes minutes, similar queries at higher level of abstractions finish in milliseconds (Section~\ref{sec:exp-semantic_path_planning}).
Note that, as mentioned in \Cref{sec:buildingParser}, \Kimera may over-segment a room as multiple rooms in the \DSG.
Nevertheless, since this over-segmentation is typically small (2 or 3 rooms), the runtime of hierarchical path-planning will remain in the order of milliseconds.

\subsection{Semantic Path Planning} 
\label{sec:smantic-paht-planning}

\DSGs also provide a powerful tool for high-level path-planning queries involving natural language by leveraging the semantic map and the different levels of abstractions.

For instance, the (connected) subgraph of places and objects in a \DSG can be used to 
issue the robot a high-level command (\eg object search~\citep{Joho11ras-objectSearch}), and the robot 
can directly infer the closest place in the \DSG it has to reach to complete the task, and can 
plan a feasible path to that place. 
Instead, the volumetric representation is not amenable to high-level path planning queries,
since the operator needs to provide metric coordinates to the robot. 
With \DSGs, the user can use natural language (``reach the cup near the sofa'').
In Section~\ref{sec:exp-semantic_path_planning}, we showcase examples of semantic queries we give to the robot, and that we use as input for hierarchical path-planning.

\begin{table*}[htbp]
  \centering
  \caption{Hierarchical semantic path-planning using A* at different levels of abstraction.
   Planning at the level of buildings (B), rooms (R), and then places (P), in a hierarchical fashion,
   is several orders of magnitude faster than planning at the volumetric level (ESDF).
  We `copy-paste' the 3D DSG of the office scene from \UnityHumans to form a neighborhood of offices in a grid-like pattern where each office is a building; hence creating a scenario of a much larger scale.
  \rebuttal{We also show the number of nodes and edges in the path, as well as the length of the estimated trajectories for comparison.}
  }

  \label{tab:semantic_path_planning}
  \begin{tabular}{ccccccccccccc}
    \toprule
     & & &
      \multicolumn{5}{c}{Timing [s]} &
      \multicolumn{3}{c}{Nodes [\#] / Edges [\#]} &
      \multicolumn{2}{c}{\rebuttal{Path Length [m]}}
      \\
      \cmidrule(l{2pt}r{2pt}){4-8}
      \cmidrule(l{2pt}r{2pt}){9-11}
      \cmidrule(l{2pt}r{2pt}){12-13}
    Dataset & Scene &  \makecell{\#B}  & \makecell{ESDF} & \multicolumn{4}{c}{Hierarchical Path-Planning}  & \makecell{P} & \makecell{R} & \makecell{B} & \makecell{ESDF} & \makecell{Hierar-\\chical} \\
      \cmidrule(l{2pt}r{2pt}){5-8}
            &       &                                  &                 & \makecell{P} & \makecell{R} & \makecell{B} & \makecell{Total} & & & & & \\
    \midrule
    \multirow{3}{*}{\UnityHumans}  & \multirow{3}{*}{Office} & 1  & 420.4    & 0.339  & 0.001 & 0.000     & 0.340 & 89 / 120  & 16 / 9  & 1 / 0 & 54.2 & 56.3\\
                                   &                         & 2  & 944.8    & 0.771  & 0.004 & 0.000     & 0.775 & 178 / 241 & 32 / 19 & 2 / 1 & 101.2 & 109.8 \\
                                   &                         & 6  & 21389.2  & 1.231  & 0.012 & 0.001     & 1.244 & 534 / 730 & 96 / 64 & 6 / 7 & 311.5 & 329.2 \\
    \bottomrule
  \end{tabular}
\end{table*}{}

\subsection{Path Planning Performance on \DSGs}
\label{sec:exp-semantic_path_planning}

\Cref{tab:semantic_path_planning} shows the timing performance of our semantic hierarchical path-planning implementation,
where we run A$^*$ at the level of buildings, rooms, and then places, in a hierarchical fashion as described in Section~\ref{sec:hierarchical-path-planning},
and compare its timing performance against running A$^*$ directly on the volumetric ESDF representation.
The scalability of our approach is further emphasized by taking the ``Office'' scene of the \UnityHumans dataset (see Fig.~\ref{fig:DSG}a, and floor-plans in Fig.~\ref{fig:truncated_esdf}) 
and replicating it in a grid-like pattern to increase its scale (where each new ``Office'' scene is considered a building).
As shown in \Cref{tab:semantic_path_planning}, hierarchical path-planning outperforms by several orders of magnitude the timing performance of planning at the volumetric ESDF level,
thereby making path-planning run at interactive speeds for large scale scenes.
\rebuttal{
We also report the length of the estimated trajectories when using the ESDF and when using the hierarchical path-planning approach.
We observe in~\Cref{tab:semantic_path_planning} that, despite being longer, the trajectories from our hierarchical approach are near as short as the ESDF ones.}

Furthermore, the queries given to our hierarchical path-planning module are of the type: ``get near any object $x$ in room $y$ of building $z$,''
where $x \in X,$ $X = \{\text{objects in room } y \text{ of building } z\},$ $y \in Y,$ $Y = \{\text{rooms in building } z \},$ $z \in Z = \{1,... ,N\}$, with $N$ being the number of buildings in the scene.
This is in stark contrast with metric coordinates $x,y,z \in \mathbb{R}^3$; which shows the ability to run semantically meaningful path-planning queries on \DSGs.


\section{Related Work}
\label{sec:relatedWork}

We review environment representations 
used in robotics and computer vision (Section~\ref{sec:relatedWork-representations}), 
and algorithms involved in building these representations (Section~\ref{sec:relatedWork-algorithms}). 

\subsection{World Representations}
\label{sec:relatedWork-representations}

         \myParagraph{Scene Graphs}
         Scene graphs are popular computer graphics models to describe, manipulate,
          and render complex scenes and are commonly used in game engines~\citep{Wang10book-openSceneGraph}.

          While in gaming applications, these structures are used to describe 3D environments,
          scene graphs have been mostly used in computer vision to abstract the content of 2D images.
         \cite{Krishna16arxiv-visualGenome} use a scene graph to model attributes and relations among
          objects in 2D images, relying on manually defined natural language captions.
          \cite{Xu17cvpr-sceneGraph} and \cite{Li17iccv-sceneGraphGeneration} 
          develop algorithms for 2D scene graph generation.
          2D scene graphs have been used for image retrieval~\citep{Johnson15cvpr},
          captioning~\citep{Krause17cvpr-sceneGraphDescription,Anderson16eccv-sceneGraphDescription,Johnson17cvpr-sceneGraphDescription},
          high-level understanding~\citep{Choi13cvpr-sceneParsing,Zhao13cvpr-sceneParsing,Huang18eccv-sceneParsing, Jiang18ijcv-sceneParsing},
          visual question-answering~\citep{Fukui16acl-QA,Zhu16cvpr-sceneGraphQandA},
         and action detection~\citep{Lu16eccv-visualRelations,Liang17cvpr-sceneGraphRelations,Zhang17cvpr-sceneGraphRelations}.
         \cite{Armeni19iccv-3DsceneGraphs} propose a \emph{3D scene graph} model to describe 3D static scenes,
          and describe a semi-automatic algorithm to build the scene graph.
          In parallel to~\citep{Armeni19iccv-3DsceneGraphs}, \cite{Kim19tc-3DsceneGraphs} propose a 3D scene graph model for robotics,
           which however only includes objects as nodes and misses multiple levels of abstraction afforded by~\citep{Armeni19iccv-3DsceneGraphs}
            and by our proposal.
          \veryOptional{Contrarily to~\cite{Armeni19iccv-3DsceneGraphs,Kim19tc-3DsceneGraphs},
           which propose a static representation of the environment,  we propose a dynamic and actionable representation,
            and automatically infer it from visual-inertial data.}
         
         \myParagraph{Representations and Abstractions in Robotics}
          The question of world modeling and map representations has been central in the robotics community since its
         inception~\citep{Thrun02a,Cadena16tro-SLAMsurvey}.
          The need to use hierarchical maps that capture rich spatial and semantic information was already
         recognized in seminal papers by Kuipers, Chatila, and Laumond~\citep{Kuipers00ai,Kuipers78cs,Chatila85}.
         \cite{Vasudevan06iros} propose a hierarchical representation of object constellations.
          \cite{Galindo05iros-multiHierarchicalMaps} use two parallel hierarchical representations
         (a spatial and a semantic representation) that are then \emph{anchored}
          to each other and estimated using 2D lidar data.
          \cite{Ruiz-Sarmiento17kbs-multiversalMaps} extend the framework in~\citep{Galindo05iros-multiHierarchicalMaps}
           to account for uncertain groundings between spatial and semantic elements.
           \cite{Zender08ras-spatialRepresentations} propose a single hierarchical representation that includes a 2D map,
            a navigation graph and a topological map~\citep{Ranganathan04iros,Remolina04}, which are then further abstracted into a \emph{conceptual map}.
         Note that the spatial hierarchies in~\citep{Galindo05iros-multiHierarchicalMaps} and~\citep{Zender08ras-spatialRepresentations} 
         already resemble a scene graph, with
         less articulated set of nodes and layers. A more fundamental difference is the fact that early work
         (i) did not reason over 3D models (but focused on 2D occupancy maps),
         (ii) did not tackle dynamical scenes, and
         (iii) did not include dense (e.g., pixel-wise) semantic information, which
         has been enabled in recent years by deep learning methods.
         
         \subsection{Perception Algorithms}
         \label{sec:relatedWork-algorithms}
         
         \myParagraph{SLAM and VIO in Dynamic Environments}
         This paper is also concerned with modeling and gaining robustness against dynamic elements in the scene.
         SLAM and moving object tracking (sometimes referred to as \emph{SLAMMOT}~\citep{Wang07ijrr-slammot} or as SLAM and Detection And Tracking of Moving Objects, \emph{DATMO}~\citep{Azim12ivs-datmo})
         has been extensively investigated in robotics~\citep{Wang07ijrr-slammot},
          while more recent work focuses on joint visual-inertial odometry and target pose
          estimation~\citep{Qiu19tro-vioObjectTracking,Eckenhoff19ral-vioObjectTracking,Geneva19arxiv-VIOandTargetTracking}.
         Most of the existing literature in robotics models the
         dynamic targets as a single 3D point~\citep{Chojnacki18ijmav-motionAndObjectTracking},
         or with a 3D pose and rely on lidar~\citep{Azim12ivs-datmo}, RGB-D cameras~\citep{Aldoma13icra-objectTracking},
         monocular cameras~\citep{Peiliang18eccv-motionAndObjectTracking}, and
         visual-inertial sensing~\citep{Qiu19tro-vioObjectTracking}.
         Related work also attempts to gain robustness against dynamic scenes by using an IMU~\citep{Hwangbo09iros-IMUKLT},
         masking portions of the scene corresponding to dynamic
         elements~\citep{Cui19access-SOFSLAM,Brasch18iros-dynamicSLAM,Bescos18ral-dynaSLAM},
          or jointly tracking camera and dynamic objects~\citep{Wang07ijrr-slammot,Bescos20arxiv-dynaslam}.
         To the best of our knowledge, the present paper is the first work that attempts to perform visual-inertial SLAM,
          segment dense object models, estimate the 3D poses of known objects, and
          reconstruct and track dense human SMPL meshes.
         
         \myParagraph{Metric-Semantic Scene Reconstruction}
         This line of work is concerned with estimating metric-semantic (but typically non-hierarchical) representations from sensor data.
         While early work~\citep{Bao11cvpr,Brostow08eccv} focused on offline processing,
         recent years have seen a surge of interest towards \emph{real-time} metric-semantic mapping, %
          triggered by pioneering works such as SLAM++~\citep{Salas-Moreno13cvpr}.
         \emph{Object-based approaches} compute an object map and include SLAM++~\citep{Salas-Moreno13cvpr},
          XIVO~\citep{Dong17cvpr-XVIO}, OrcVIO~\citep{Mo19tr-orcVIO}, QuadricSLAM~\citep{Nicholson18ral-quadricSLAM},
          and~\citep{Bowman17icra}.
          For most robotics applications, an object-based map
          does not provide enough resolution for navigation and obstacle avoidance.
         \emph{Dense approaches} build denser semantically annotated models in the form
         of point clouds~\citep{Behley19iccv-semanticKitti,Tateno15iros-metricSemantic,Renaud18rss-segMap,Lianos18eccv-VSO},
         meshes~\citep{Rosinol20icra-Kimera,Rosu19ijcv-semanticMesh}, 
         surfels~\citep{Whelan15rss-elasticfusion,Runz18ismar-maskfusion,Wald18ral-metricSemantic},
         or volumetric models~\citep{McCormac17icra-semanticFusion,Rosinol20icra-Kimera,Grinvald19ral-voxbloxpp,Narita19arxiv-metricSemantic}.
         Other approaches use both objects and dense models, see
         \cite{Li16iros-metricSemantic} and
           Fusion++~\citep{McCormac183dv-fusion++}.
          These approaches focus on static environments.
           Approaches that deal with moving objects, such as DynamicFusion~\citep{Newcombe15cvpr-dynamicFusion},
           Mask-fusion~\citep{Runz18ismar-maskfusion},
           Co-fusion~\citep{Runz17icra-cofusion}, and MID-Fusion~\citep{Xu19icra-midFusion}
          are currently limited to small table-top scenes and focus on objects or dense maps, rather than  scene graphs.
          Most of these works rely on GPU processing~\citep{McCormac17icra-semanticFusion,Zheng19arxiv-metricSemantic,Tateno15iros-metricSemantic,Li16iros-metricSemantic,McCormac183dv-fusion++,Runz18ismar-maskfusion,Runz17icra-cofusion,Xu19icra-midFusion}.
           Recent work investigates CPU-based approaches in combination with RGB-D sensing, \eg \cite{Wald18ral-metricSemantic},
            PanopticFusion~\citep{Narita19arxiv-metricSemantic}, and Voxblox++~\citep{Grinvald19ral-voxbloxpp}. %
            A sparser set of contributions addresses other sensing modalities, including monocular cameras
            (\eg CNN-SLAM~\citep{Tateno17cvpr-CNN-SLAM}, VSO~\citep{Lianos18eccv-VSO}, VITAMIN-E~\citep{Yokozuka19arxiv-vitamine}, XIVO~\citep{Dong17cvpr-XVIO}) and lidar~\citep{Behley19iccv-semanticKitti,Renaud18rss-segMap}.

\myParagraph{Loop Closure with Dense Representations}
This line of work is concerned with correcting a dense representation of the environment (\eg point clouds, meshes, voxels) after a loop closure %
occurs.
LSD-SLAM~\citep{Engel14eccv-lsdslam} represents the environment with a point cloud; loop closures do not have a direct effect on the 
point cloud map, but rather correct the pose graph associated to the camera keyframes and the 
semi-dense local maps attached to each keyframe are updated accordingly.
We are especially interested in the cases where the environment is represented as a mesh
and the loop closures are enforced by deforming this mesh. %
Kintinuous~\citep{Whelan13iros} accomplishes this in two optimization step: first it optimizes the pose 
graph and, by utilizing the relationship between the vertices of the mesh and the poses in the pose graph,
 it then uses the optimized pose graph as measurement constraints in order to deform the mesh with a \emph{deformation graph}~\citep{Summer07siggraph-embeddedDeformation}. 
\rebuttal{MIS-SLAM~\citep{Song18ral-misSlam} also uses the \emph{deformation graph} approach to deform the model point cloud 
using the estimate from ORB-SLAM~\citep{Mur17-TRO}.}
ElasticFusion~\citep{Whelan15rss-elasticfusion} instead deforms a dense map of surfels and 
\rebuttal{GravityFusion~\citep{Puri17iros-gravityFusion} builds on top of ElasticFusion by 
enforcing a consistent gravity direction among all the surfels.}
Voxgraph~\citep{Reijgwart20ral-voxgraph} builds a globally consistent volumetric map by applying graph optimization 
over a set of submap poses and including odometry and loop closure constraints. 
\rebuttal{Similarly, DynamicFusion~\citep{Newcombe15cvpr-dynamicFusion}, VolumeDeform~\citep{Innmann2016arxiv-volumeDeform},
and Fusion4D~\citep{Dou16acm-fusion4d} use a volumetric representation for fusion and deformation.}
\rebuttal{Our approach is the first to simultaneously optimize the pose graph and the mesh, 
and also the first to formalize this problem as a pose graph optimization.}

\myParagraph{Metric-to-Topological Scene Parsing}
This line of work focuses on %
partitioning a metric map into semantically meaningful places (\eg rooms, hallways).
\cite{Nuchter08ras-semanticMaps} encode relations among planar surfaces (\eg walls, floor, ceiling) and detect objects in the scene.
\cite{Blanco09ras-metricTopologicalSLAM,Gomez20icra} propose a hybrid metric-topological map.
\cite{Friedman07ijcai-voronoiRF} propose \emph{Voronoi Random Fields} to obtain an abstract model of a 2D
grid map.
 \cite{Rogers12icra-semanticMapping} and \cite{Lin13cvpr-holisticSceneUnderstanding}
 leverage objects to perform a joint object-and-place classification.
While \cite{Nie20-cvpr,Huang18nips,Zhao13cvpr-scene} jointly solve the problem of scene understanding and reconstruction.
\cite{Pangercic12icra-objectMaps} reason on the objects' functionality.
\cite{Pronobis12icra-MRFsemanticMapping} use a Markov Random Field to segment a 2D grid map.
\cite{Zheng18aaai-graphStructuredNet}
infer the topology of a grid map using a \emph{Graph-Structured Sum-Product Network}, while \cite{Zheng19iros-topoNet} use a neural network.
\cite{Armeni16cvpr-3DsemanticParsing} focus on a 3D mesh, and propose a method to parse a building into rooms. %
Floor plan estimation has been also investigated using single images~\citep{Hedau09cvpr-floorPlan,Schwing13iccv-boxInBox},
omnidirectional images~\citep{Lukierski17icra-floorPlan}, 2D lidar~\citep{Li20arxiv-floorPlan,Turner14grapp-floorPlan},
3D lidar~\citep{Mura14cg-floorPlan,Ochmann14grapp-floorPlan}, RGB-D~\citep{Chen18eccv-floorNet},
or from crowd-sourced mobile-phone trajectories~\citep{Alzantot12icagis-floorPlan}.
The works~\citep{Armeni16cvpr-3DsemanticParsing,Mura14cg-floorPlan,Ochmann14grapp-floorPlan} are closest to our proposal,
but contrarily to~\citep{Armeni16cvpr-3DsemanticParsing} we do not rely on a Manhattan World assumption,
and contrarily to~\citep{Mura14cg-floorPlan,Ochmann14grapp-floorPlan} we operate on a mesh model.
Recently,~\cite{Wald20cvpr-semanticSceneGraphs} propose to learn from point clouds a 3D semantic scene graph 
that focuses on representing semantically meaningful inter-instance relationships.

\myParagraph{Human Pose Estimation and Tracking}
Human pose and shape estimation from a single image is a growing research area. While
we refer the reader to~\citep{Kolotouros19cvpr-shapeRec,Koloturos19arxiv-IterativeShape,Kolotouros19iccv,Kocabas20cvpr-vibe}
for a broader review, it is worth mentioning that related work includes optimization-based approaches,
which fit a 3D mesh to 2D image keypoints~\citep{Bogo16eccv-keepItSMPL,Lassner17cvpr-humanPoseAndShape,Zanfir18cvpr-humanPoseAndShape,Koloturos19arxiv-IterativeShape,Yang20cvpr-shapeStar}, and learning-based methods,
which infer the mesh directly from pixel information~\citep{Tan17bmvc-humanPoseAndShape,Kanazawa18cvpr-humanPoseAndShape,Omran183dv-humanPoseAndShape,Pavlakos18cvpr-humanPoseAndShape,Kolotouros19cvpr-shapeRec,Koloturos19arxiv-IterativeShape}.
Human models are typically parametrized using the \emph{Skinned Multi-Person Linear Model} (SMPL)~\citep{Loper15tg-smpl}, which provides a compact pose and shape description and can be rendered as a mesh with 6890 vertices and 23 joints.
The common approach to monocular human tracking is to predict joints probabilities in 2D image space and which are optimized to 3D joints based on multiple time-series 
observations and motion priors~\citep{Andriluka10cvpr, Andriluka08cvpr-peopleTrack, Arnab19cvpr-exploiting, Bridgeman19cvpr-multi, Elhayek12cvpr-spatio, Zhou18PAMI-monocap, Wang20arxiv-combining}.
\cite{Taylor10cvpr-dynamical} combines a learned motion model with particle filtering to predict 3D human poses.
In this work, we aim to not only estimate the 3D pose of the human, but also the full SMPL shape without maintaining the persistent image history required by many of the 
approaches above. In human tracking literature, only \cite{Arnab19cvpr-exploiting} fully reconstruct the SMPL shape of the human; however, they reconstruct the 
shape after performing data association over multiple timesteps. In contrast, we use the method of~\cite{Kolotouros19cvpr-shapeRec} to 
directly get the full 3D pose of the human at each timestep, simplifying pose estimation, and allowing us to do data association based on the SMPL body shape. 


\section{Conclusion} 
\label{sec:conclusion}

We introduced \emph{3D Dynamic Scene Graphs}, a unified 
representation for actionable spatial perception.
Moreover, we presented \Kimera, the first \emph{\SPESlong} that builds a \DSG from visual-inertial data in a fully automatic fashion. 
 We showcased \Kimera in photo-realistic simulations and real data, and discussed applications enabled by the proposed \DSG representation, 
 including semantic and hierarchical path planning.

This paper opens several research avenues. 
First of all, it would be desirable to develop spatial perception engines that run incrementally and in real time.
Currently, while the creation of the metric-semantic reconstruction happens in real-time, the rest of the scene graph 
is built at the end of the run and requires few minutes to parse the entire scene. 
Second, it would be interesting to design engines that estimate a \DSG from heterogeneous sensors and 
from sensor data collected by multiple robots.
Finally, a natural direction is to enrich a \DSG with other physical attributes, including 
material type and affordances for objects, and trying to learn attributes and relations from data.


\section*{Acknowledgments} 
We are thankful to Dan Griffith, Ben Smith, Arjun Majumdar, and Zac Ravichandran for open-sourcing the TESSE simulator, and to 
 Winter Guerra and Varun Murali for the discussions about Unity.

\section*{Disclaimer}

DISTRIBUTION STATEMENT A. Approved for public release. Distribution is unlimited.

This material is based upon work supported by the Under Secretary of Defense for Research and Engineering under Air Force Contract No. FA8702-15-D-0001. Any opinions, findings, conclusions or recommendations expressed in this material are those of the author(s) and do not necessarily reflect the views of the Under Secretary of Defense for Research and Engineering.


\bibliographystyle{SageH} 
\bibliography{../references/refs.bib,../references/myRefs.bib}
\end{document}